\documentclass{article}

\usepackage[main, final]{neurips_2026}

\usepackage[utf8]{inputenc} 
\usepackage[T1]{fontenc}    
\usepackage{hyperref}       
\usepackage{url}            
\usepackage{booktabs}       
\usepackage{amsfonts}       
\usepackage{nicefrac}       
\usepackage{microtype}      
\usepackage{xcolor}         

\usepackage{times}
\usepackage{latexsym}
\usepackage{graphicx}
\usepackage{amsmath}
\usepackage{amsthm}
\usepackage{bbm}
\usepackage{caption}
\usepackage{multirow}
\usepackage{subfigure}
\usepackage{subcaption}
\usepackage{pgfplots}
\usepackage{xspace}
\usepackage{enumitem}
\usepackage{wrapfig}
\usepackage{bbding}
\usepackage{colortbl}
\usepackage{amsthm}
\usepackage{sidecap}

\usepackage{algorithm}
\usepackage{algpseudocode}

\usepackage{adjustbox}
\usepackage{amssymb}
\usepackage[table]{xcolor}

\usepackage[most]{tcolorbox}
\usepackage{caption}

\usepackage{wrapfig}

\newcommand{\methodname}{\textsc{AutoMMemo }}

\definecolor{poscolor}{RGB}{0,120,80}
\definecolor{negcolor}{RGB}{180,50,50}
\definecolor{zerocolor}{RGB}{90,90,90}

\newtcolorbox{promptbox}[2][]{
    enhanced,
    breakable,
    width=\linewidth,
    colback=gray!8,
    colframe=gray!8,
    coltitle=black,
    colbacktitle=gray!8,
    boxsep=0pt,
    left=10pt,
    right=10pt,
    top=2pt,
    bottom=2pt,
    fontupper=\linespread{0.9}\selectfont,
    fonttitle=\bfseries,
    title=#2,
    #1
}

\newtcolorbox{promptboxsm}[2][]{
	width=\linewidth,
	colback = gray!8, 
	colframe = gray!8, 
	coltitle = black,
	colbacktitle = gray!8,            
	boxsep=0pt,left=10pt,right=10pt,top=0pt,bottom=0pt,
	fontupper=\scriptsize\linespread{0.9}\selectfont,
	fonttitle=\bfseries,
	title=#2,#1
}

\newtcolorbox{examplebox}[1][]{
  width=\linewidth,
  colback = orange!6,
  colframe = orange!30,
  coltitle = black,
  boxrule = 0.4pt,
  boxsep=0pt,left=10pt,right=10pt,top=6pt,bottom=6pt,
  fontupper=\linespread{0.95}\selectfont,
  #1
}

\definecolor{blue2}{HTML}{398cd6}
\newcommand{\fstring}[1]{\textcolor{blue2}{\detokenize{{#1}}}}

\newcounter{prompt}
\newcommand{\promptcaption}[2]{%
  \refstepcounter{prompt}%
  \noindent\textbf{Prompt~\theprompt: #1}%
  \label{#2}\par\vspace{3pt}
}

\title{Learning to Learn from Multimodal Experience}

\author{%
  Xingyu Sui\textsuperscript{1}, Weixiang Zhao\textsuperscript{1}\thanks{\ \ Corresponding author} , Yongxin Tang\textsuperscript{1} \\ \textbf{Yanyan Zhao\textsuperscript{1}}, 
  \textbf{Yang Wu\textsuperscript{2}}, \textbf{Dandan Tu\textsuperscript{2}}, \textbf{Bing Qin\textsuperscript{1}} \\
  \textsuperscript{1}Harbin Institute of Technology, \textsuperscript{2}Huawei Technologies Co., Ltd\\
  \texttt{\{xysui, wxzhao, yyzhao\}@ir.hit.edu.cn} \\
}

\begin{document}

\maketitle

\begin{abstract}
Multimodal agents increasingly interact with real-world-like environments and generate rich trajectories spanning perception, reasoning, and action. A key research focus is how to transform these raw multimodal trajectories into reusable experience that can improve future task performance. Existing experience-driven learning methods are mostly developed in textual settings and typically rely on manually designed memory schemas, making them less effective for multimodal experience that requires composing heterogeneous signals across modalities, temporal scales, and abstraction levels.
We propose \methodname, a general framework that enables agents to automatically design memory mechanisms for learning from multimodal experience. \methodname represents each memory mechanism as an executable \emph{memo program}, which defines how past episodes are stored, organized, and retrieved. It discovers effective memo programs through an iterative process of update-then-retrieve evaluation, reflection-guided mutation, and budget-aware tree search.
Experiments across GUI/Web navigation and multimodal visual reasoning benchmarks show that \methodname consistently improves agent performance over no-memory agents and fixed memory baselines, while maintaining reasonable token and interaction costs. Transfer and search-process analyses further demonstrate that the learned memory designs capture reusable experience structures and progressively refine retrieval relevance. These results highlight adaptive memory design as a key optimization target for multimodal experience-driven learning. Code is available at \url{https://github.com/XingYuSSS/AutoMMemo}.
\end{abstract}
\section{Introduction}

Recent advances in multimodal large language models (MLLMs) have enabled agents to actively explore and interact within environments that more closely resemble the real world \citep{yao2023react,durante2024agent,li2025perception}. Such interactions naturally generate large volumes of trajectories, capturing rich signals across perception, reasoning, and action. A central research focus, therefore, is how to transform these raw interaction trajectories into reusable experience that can continuously improve agent performance \citep{shinn2023reflexion,silver2025welcome}.
In this context, experience-driven learning via memory mechanisms has emerged as a promising paradigm, enabling agents to accumulate, organize, and reuse experience across tasks \citep{gao2025survey,fang2025comprehensive}. As a result, the effectiveness of an agent is no longer solely determined by its underlying model, but increasingly by how it represents and utilizes its experience \citep{dou2025evalearn}.

However, existing experience-driven learning paradigms have been predominantly developed in textual settings \citep{zhao2024expel,zhang2025g,xia2026skillrl}, leaving their extension to multimodal environments largely underexplored. This creates a fundamental mismatch: in real-world scenarios, experience is inherently multimodal, arising from interactions that involve diverse signals such as vision, language, and action dynamics \citep{driess2023palm,sarch2024vlm}. These signals differ in modality, temporal scope, and semantic abstraction, forming a complex mixture of structured and unstructured information. Thus, designing an effective memory system for experience learning requires making a series of tightly coupled decisions, such as what to store, how to represent it, how to organize it, and when to retrieve it. For multimodal experience, these decisions become particularly challenging, as useful information may be distributed across visual observations, language-based reasoning, and action trajectories at different temporal scales and abstraction levels \citep{hu2025memory}.

Despite this complexity, existing approaches largely rely on manually designed memory schemas, with fixed and predefined formats for experience representation and retrieval. While effective in controlled settings, such designs inherently depend on human priors and lack the flexibility to adapt across tasks, modalities, and environments. As a result, they often lead to sub-optimal experience utilization: inappropriate experience representations may either discard critical signals, such as fine-grained visual cues \citep{bo2025agentic,jiang2026xskill}, or introduce excessive noise \citep{li2025echotrail,zhu2026hybrid}, ultimately hindering learning efficiency and generalization.

These limitations point to a deeper issue: the optimal way to structure and utilize multimodal experience is inherently task-dependent and evolves over time as both the environment and the agent itself change. Therefore, memory design in multimodal agents should not be treated as a one-shot engineering decision, but rather as a continually evolving process. Building on this insight, we argue that effective multimodal experience-driven learning requires not only accumulating or retrieving experience, but learning how to structure and utilize it adaptively. In contrast to existing approaches that operate on fixed memory designs, this shifts the focus toward a higher-level objective: enabling agents to adapt their own memory strategies for improved learning. This view is consistent with human learning, where individuals continuously refine how they organize, abstract, and recall experience based on context and goals \citep{cohen1993memory,eichenbaum2017memory}.

To this end, we propose \methodname, a general framework that enables agents to automatically design memory for learning from multimodal experience. Rather than relying on fixed memory schemas, \methodname represents each memory mechanism as an executable \emph{memo program}, which specifies how past multimodal episodes are stored, organized, and retrieved for future tasks. This abstraction provides a unified memory interface while keeping the internal design space open-ended, allowing agents to explore diverse strategies for structuring and utilizing experience.
\methodname discovers effective memo programs through iterative evaluation, reflection, mutation, and selection. Each candidate is assessed under an update-then-retrieve protocol that measures whether its constructed memory improves performance on held-out tasks. The resulting trajectories and retrieved memories are then used by a meta agent to diagnose weaknesses and generate improved descendants. To make this search efficient under limited evaluation budgets, \methodname uses a budget-aware tree search strategy that balances re-evaluating promising programs with exploring new memory designs.

Empirically, our experiments show that adaptively learned memory mechanisms consistently improve agent performance across both GUI/Web navigation and multimodal visual reasoning benchmarks. Compared with no-memory agents and manually fixed memory designs, the memo programs discovered by \methodname lead to stronger task performance while maintaining reasonable interaction and token costs. Further transfer analyses show that effective memory strategies are not merely overfitted to a single benchmark or model, but can provide measurable gains across different environments and execution agents. In addition, our search-process analysis reveals that \methodname progressively refines memory designs by retaining useful experience patterns, removing noisy or redundant information, and improving retrieval relevance over iterations. Together, these results demonstrate that memory design itself is a key object of optimization for multimodal agents, and that learning how to structure and retrieve experience can substantially improve experience-driven learning beyond fixed memory schemas.

\section{Related Works}

\noindent{\textbf{Learning from Experience with Memory.}}
Memory-augmented experience learning has emerged as a central paradigm for enabling agents to improve from past interactions \citep{gao2025survey,fang2025comprehensive}. Existing approaches primarily focus on extracting and representing experience from interaction trajectories, and largely rely on manually designed memory schemas.

In textual settings, experience is typically represented in predefined forms, such as {insights} \citep{shinn2023reflexion,zhao2024expel,fu2024autoguide,zhang2025g}, {skills} \citep{wang2024voyager,chen2024automanual,xia2026skillrl}, or structured {workflows} \citep{wang2025agent}. These representations enable experience reuse but are fixed by design and require human specification.

Recent work extends this paradigm to multimodal settings by incorporating additional sensory signals \citep{sarch2024vlm,chhikara2025mem0,yuan2026ted,chen2026polarmem}. For example, \citet{bo2025agentic,jiang2026xskill} augment experience extraction with visual information, while \citet{liu2025memverse} models cross-modal dependencies via graph-structured memory. Despite these extensions, they still rely on predefined memory formats and manually designed representation structures.

\noindent{\textbf{Automatic Memory Design.}}
To overcome the limitations of manually designed memory schemas, recent works explore automatic memory design, where agents learn to optimize their memory.

In textual settings, existing approaches can be broadly categorized based on their design spaces. The first line of works operate within constrained design spaces, where memory is defined using predefined components. These components may correspond to a single form of experience \citep{zhou2026memento,zhang2026evoskills,huang2026bilevel} or combinations of multiple predefined memory structures \citep{zhang2025memevolve}, over which optimization or selection is performed. The second line of works explore more open-ended design spaces, where memory is defined and optimized in the form of executable programs or policies \citep{xiong2026learning,pan2026m}.

Despite recent progress, existing approaches remain largely confined to purely textual environments and struggle to generalize to multimodal experience. A concurrent study extends automatic memory design to multimodal settings \citep{liu2026omni}; however, it is mainly tailored to question-answering tasks, leaving the broader challenge of learning generalizable and flexible memory mechanisms for multimodal experience largely unaddressed.

\begin{figure*}[t]
    \centering
    \includegraphics[width=\columnwidth]{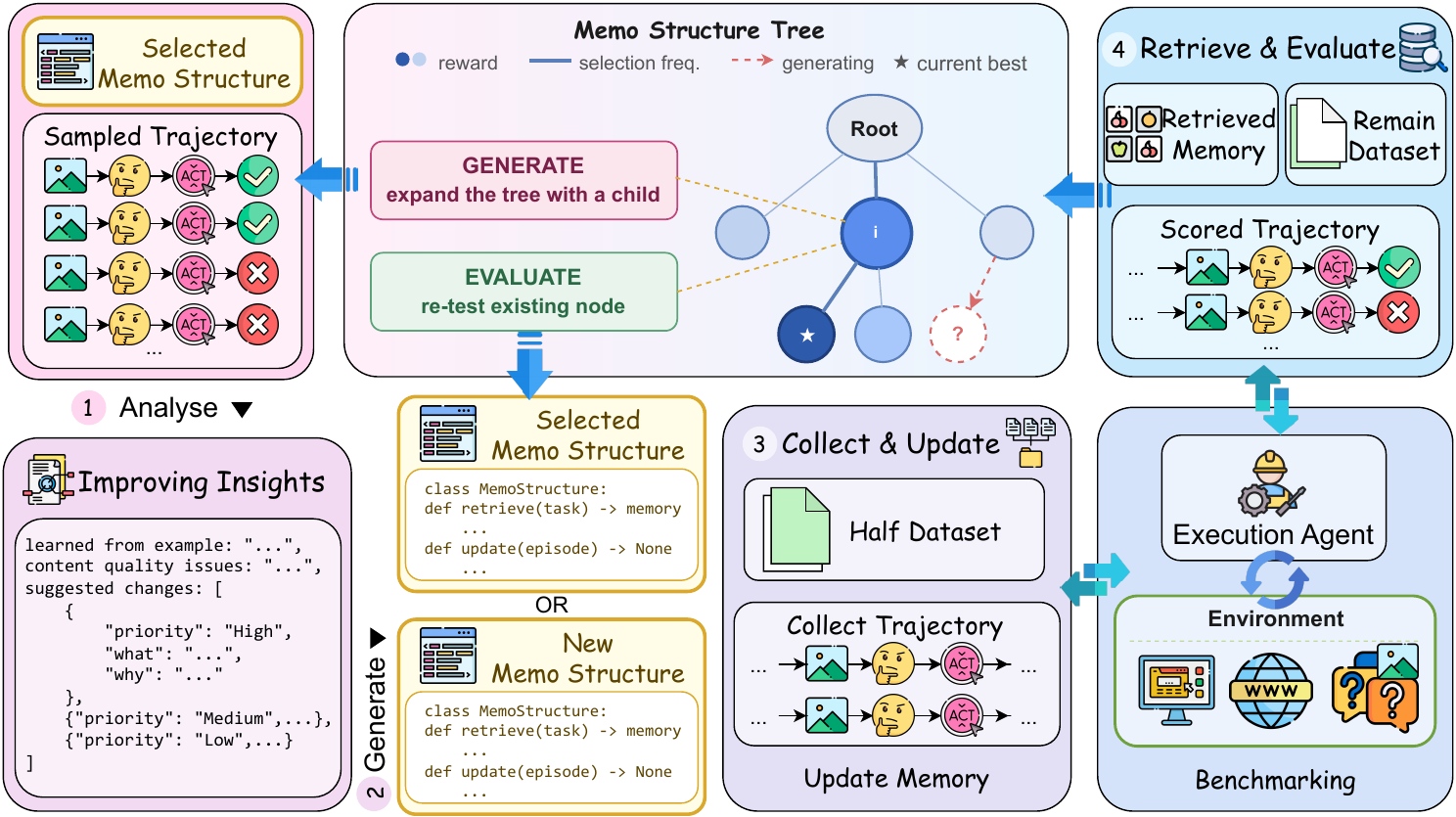}
    \caption{
Overview of \methodname. 
It searches over executable memo programs that specify how multimodal experience is stored and retrieved. 
Each candidate is evaluated with an update-then-retrieve protocol on held-out tasks, and the resulting trajectories and scores are used to derive reflection-based improvement insights for program mutation. 
A budget-aware tree search maintains a memo-structure tree and decides whether to re-evaluate an existing node or generate a new descendant.
}
    \label{fig:main}
\end{figure*}

\section{Method}
\label{sec:method}

We propose \methodname, a general framework for learning adaptive memory mechanisms from multimodal experience.
As illustrated in Figure~\ref{fig:main}, \methodname consists of four stages. First, we define a unified memo-program interface that keeps evaluation comparable while leaving the internal memory mechanism open-ended (Section~\ref{sec:memo-program}). Second, we evaluate each candidate using an update-then-retrieve protocol that measures whether the program can convert past episodes into useful context for held-out tasks (Section~\ref{sec:update-retrieve}). Third, we use execution evidence to diagnose memory failures and mutate programs into improved descendants (Section~\ref{sec:mutation}). Finally, we allocate the limited evaluation budget with a tree-search strategy that balances re-testing existing programs and exploring new memory designs (Section~\ref{sec:tree-search}).

\subsection{Executable Memo Programs}
\label{sec:memo-program}

A memo program \(m\) is an executable Python program with two operations, \(m.\mathrm{update}(e)\) and \(m.\mathrm{retrieve}(e)\), where \(e\) denotes an episode recorder. The update operation incorporates past experience into an internal memory state, while the retrieve operation returns a memory payload for a new task. This interface fixes the external contract of memory, but leaves the internal mechanism open-ended.
The episode recorder provides a unified representation of multimodal experience. It may include task instructions, visual observations, environment states, action histories, model messages, task-specific metadata, and final outcomes. The retrieved memory payload may contain textual summaries, visual references, structured metadata, or multimodal embeddings. Textual fields are injected into the agent prompt, while visual references can be provided as additional visual inputs.

Under this abstraction, a memo program can implement different memory strategies: what to store, how to summarize trajectories, how to index textual and visual observations, how to associate states with actions and outcomes, and how to rank retrieved memories. The program can use a controlled set of general primitives, such as embedding models, multimodal model calls, vector stores, graph utilities, and lightweight databases. During search, the execution agent, environment, and evaluator are fixed; only the memo program is optimized. Thus, each candidate is judged by downstream reward under the same evaluation protocol.

\subsection{Update-then-Retrieve Evaluation}
\label{sec:update-retrieve}

Given a candidate memo program \(m\), \methodname evaluates whether it can transform past multimodal experience into useful context for future tasks. Let \(\mathcal{D}_u\) be a set of update episodes and \(\mathcal{D}_r\) be a disjoint set of retrieve-time tasks. The update episodes are collected once using the fixed execution agent and shared across all candidates.

For each update episode \(\tau \in \mathcal{D}_u\), we obtain an episode recorder \(e_\tau\) and call \(m.\mathrm{update}(e_\tau)\) to build the program's memory state. After all update episodes are processed, this memory state is fixed. For each held-out task \(x \in \mathcal{D}_r\), the program receives a partial recorder \(e_x\) containing the task input and initial observations, and returns a memory payload \(p_x=m.\mathrm{retrieve}(e_x)\). The payload is provided to the fixed execution agent, which completes the task and receives a scalar reward \(R_x \in [0,1]\).

The validation score of \(m\) is the average reward over retrieve-time tasks:
\begin{equation}
    s(m)
    =
    \frac{1}{|\mathcal{D}_r|}
    \sum_{x \in \mathcal{D}_r} R_x .
    \label{eq:validation-score}
\end{equation}
Because the update episodes, execution agent, environment, and evaluator are shared across candidates, differences in \(s(m)\) reflect the quality of the memo program.

To ensure comparability, all candidates follow the same retrieval budget, including limits on payload length and the number of returned visual references. This prevents trivial gains from excessive context injection and encourages concise, task-relevant multimodal memory. The retrieved payloads and resulting trajectories are recorded as evidence for subsequent mutation.

\subsection{Reflection-Guided Program Mutation}
\label{sec:mutation}

After evaluating a memo program, \methodname uses execution evidence to generate improved descendants. For a candidate \(m_i\), the meta agent receives its source code, validation score, retrieved payloads, and representative retrieve-time trajectories. This evidence allows the meta agent to diagnose how the current memory mechanism affects downstream behavior.

The diagnosis identifies whether retrieved memories are helpful, missing, irrelevant, redundant, or misleading. For multimodal tasks, it also examines whether visual memories help ground observations, recognize relevant states, select actions, or avoid previous failures. The diagnosis may reveal structural issues such as weak text-vision alignment, unstable retrieval keys, or excessive truncation.

Conditioned on this diagnosis, a code generation model mutates the parent program into a child memo program. The mutation may revise trajectory summarization, textual or visual indexing, state-action association, memory ranking, or payload formatting. The child must preserve the same \(m.\mathrm{update}\) and \(m.\mathrm{retrieve}\) interface, so it can be evaluated under the same protocol.

Before full evaluation, the child undergoes a quick examination that checks imports, interface implementation, execution on sample episodes, payload schema validity, and retrieval-budget compliance. If the child fails, a repair model attempts to fix it for a limited number of attempts. Only valid programs are inserted into the generation tree for full evaluation. We provide the full prompts used by the meta agent, including diagnosis, mutation, quick examination, and repair prompts, in Appendix~\ref{app:prompts}.

\subsection{Budget-Aware Tree Search}
\label{sec:tree-search}

Evaluating memo programs is expensive, so \methodname must decide whether to re-evaluate existing programs or generate new ones. We maintain a generation tree \(\mathcal{T}\), where each node \(i\) is a memo program \(m_i\), and each edge \(i \rightarrow j\) indicates that \(m_j\) is generated by mutating \(m_i\).

At each round, \methodname considers two actions. The first action, \(\textsc{Evaluate}(i)\), re-evaluates an existing program to reduce uncertainty. Its score is
\begin{equation}
    \mathrm{UCB}^{\mathrm{eval}}_i
    =
    \hat{\mu}_i
    +
    c_e \sqrt{\frac{\log N}{n_i}},
    \label{eq:eval-ucb}
\end{equation}
where \(\hat{\mu}_i\) is the empirical score of node \(i\), \(n_i\) is its number of full evaluations, \(N\) is the total number of evaluations consumed, and \(c_e\) controls exploration.

The second action, \(\textsc{Generate}(i)\), expands node \(i\) by reflection-guided mutation. A good parent should both perform well and have potential to produce better descendants. For each evaluated child \(j\) of \(i\), we measure its positive improvement by
\begin{equation}
    \Delta_{i \rightarrow j}
    =
    \max\left(0,\hat{\mu}_j-\hat{\mu}^{\mathrm{snap}}_i\right),
    \label{eq:positive-improvement}
\end{equation}
where \(\hat{\mu}^{\mathrm{snap}}_i\) is the parent's score when \(j\) is generated. This avoids changing the improvement estimate when the parent is later re-evaluated.

For node \(i\), let \(K_i\) be the number of evaluated children and \(S_i\) be their cumulative positive improvement. We estimate its local improvement potential as
\begin{equation}
    \hat{\Delta}_i
    =
    \frac{
        \beta \rho \max(0,\hat{\mu}_i-\hat{\mu}_0) + S_i
    }{
        \beta + K_i
    },
    \label{eq:local-improvement}
\end{equation}
where \(m_0\) is the no-memory root, \(\rho\) controls the strength of a root-relative prior, and \(\beta\) is a pseudo-count. The generation action is scored by
\begin{equation}
    \mathrm{UCB}^{\mathrm{gen}}_i
    =
    \hat{\mu}_i
    +
    \hat{\Delta}_i
    +
    c_g \sqrt{\frac{\log N}{\beta + K_i}},
    \label{eq:gen-ucb}
\end{equation}
where \(c_g\) controls exploration over parent nodes.

To avoid collapsing into a single lineage, \methodname uses a minimum-width constraint. The root is always expandable, while a non-root node becomes expandable only after its parent has generated at least \(B\) children:
\begin{equation}
    \mathcal{E}
    =
    \{0\}
    \cup
    \left\{
    i \in \mathcal{T}\setminus\{0\}
    :
    |\mathrm{Children}(\mathrm{Parent}(i))| \ge B
    \right\}.
    \label{eq:eligible-set}
\end{equation}
At each round, \methodname selects the highest-scoring action among all evaluation actions and eligible generation actions. It then either re-evaluates an existing node or generates, repairs, evaluates, and inserts a new child.

After the search budget is exhausted, \methodname selects a robust final program using a lower confidence bound:
\begin{equation}
    m^*
    =
    \arg\max_{i \in \mathcal{T}}
    \left(
    \hat{\mu}_i
    -
    c_e \sqrt{\frac{\log N}{n_i}}
    \right).
    \label{eq:final-selection}
\end{equation}
This favors programs that are both high-performing and sufficiently verified, yielding a robust memory strategy for learning from multimodal experience.
\section{Experiments}
\label{sec:experiments}

\subsection{Experimental Setup}

\noindent{\textbf{Benchmarks.}}
We evaluate \methodname on four benchmarks covering GUI/Web navigation and multimodal visual reasoning: WebVoyager~\citep{he2024webvoyager}, Mind2Web~\citep{deng2023mind2web}, AgentVista~\citep{su2026agentvista}, and MMSearch-Plus~\citep{tao2025mmsearch}.
Detailed per-benchmark statistics and dataset split details are provided in Appendix~\ref{app:dataset_details}.

\noindent{\textbf{Models.}}
We instantiate the execution agent with both open-source and closed-source backbones, including Qwen3-VL-32B~\citep{bai2025qwen3}, GPT-5.4-nano~\citep{singh2025openai}, and Qwen3.5-Plus~\citep{bai2025qwen3}. 
Unless otherwise specified, the meta agent is GPT-5~\citep{singh2025openai}. 
We use GPT-5.4-mini~\citep{singh2025openai} as the LLM judge for all benchmarks. 
The two GUI/Web navigation benchmarks share the same evaluation prompt, while the two visual reasoning/search benchmarks share another prompt tailored to answer correctness and evidence consistency. 

\noindent{\textbf{Baselines.}}
We compare with three groups of memory baselines. 
Text-based baselines include Trajectory Retrieval~\citep{park2023generative,xu2023retrieval}, ReasoningBank~\citep{ouyang2025reasoningbank}, and G-Memory~\citep{zhang2025g}. 
Multimodal baselines include XSkill~\citep{jiang2026xskill} and M$^2$~\citep{yan2026m}. 
We also compare with ALMA~\citep{xiong2026learning}, which we adapt as a text-based automatic memory-design baseline under our evaluation protocol. 
For a fair comparison under the same sequential search budget, both ALMA and \methodname are allocated 20 search steps; for ALMA, we sample one candidate design per step to match our sequential design-search protocol. 
Detailed baseline implementations and adaptations are provided in Appendix~\ref{app:baseline_details}.

\noindent{\textbf{Evaluation protocol.}}
All methods start from an empty memory for each benchmark. 
Memory design/search and memory construction are performed independently on the training split of each benchmark, without mixing episodes across benchmarks. 
The resulting memory system is evaluated on the held-out test split of the same benchmark without further updates, resulting in an offline evaluation protocol.
Each method is evaluated three times on each benchmark, and we report the mean score. 
Task failures are counted as zero, while infrastructure-level invalid runs, such as browser crashes, page loading failures, or API errors, are excluded and rerun when possible. 
We report success rate for GUI/Web navigation benchmarks and judge-based accuracy for visual reasoning benchmarks. 
Following the evaluation protocols of WebVoyager~\citep{he2024webvoyager} and AgentVista~\citep{su2026agentvista}, we use benchmark-family-specific LLM-as-judge prompts for GUI/Web navigation and visual reasoning tasks, respectively. 
The evaluator receives the task instruction, trajectory, final answer or state, and a bounded number of screenshots, and outputs a binary correctness label. 
Full prompts are provided in Appendix~\ref{app:prompts}.
AVG. GUI and AVG. VR are computed as simple macro-averages over the corresponding benchmarks.

\noindent{\textbf{Implementation details.}}
The search budget is set to 20 steps for both \methodname and ALMA. 
During judging, the evaluator can access the task, the agent trajectory, the final answer or state, and up to five screenshots. 
During execution, we allow up to eight images in the agent context. 
Retrieved memories are capped at 50k characters and two images. 
Unless otherwise specified, all methods use the same execution budget, retrieval budget, and evaluation protocol. 
Additional hyperparameters, concurrency settings, model identifiers, system details, and all prompt templates are provided in Appendix~\ref{app:implementation_details} and Appendix~\ref{app:prompts}.

\begin{table*}[t]
\centering
\caption{
Main results on GUI-agent and visual-reasoning benchmarks across different execution models. 
Numbers in parentheses indicate absolute performance changes over the corresponding NoMemory baseline, and bold denotes the best result within each execution-model block.
}
\label{tab:main_result}
\begin{adjustbox}{width=\columnwidth}
\begin{tabular}{lcccccc}
\toprule
\multirow{2.5}{*}{\textbf{Method}}
& \multicolumn{2}{c}{\textbf{GUI Agent}}
& \multirow{2.5}{*}{\textbf{AVG. GUI}}
& \multicolumn{2}{c}{\textbf{Visual Reasoning}}
& \multirow{2.5}{*}{\textbf{AVG. VR}} \\

\cmidrule(lr){2-3} \cmidrule(lr){5-6}

& WebVoyager & Mind2Web &  & AgentVista & MMSearch-Plus \\ 
\midrule

\multicolumn{7}{c}{\textit{Qwen3-VL-32B}} \\
\midrule

NoMemory & \makebox[2.8em][r]{36.32}\makebox[2.2em][l]{\hspace{0.2em}} & \makebox[2.8em][r]{22.31}\makebox[2.2em][l]{\hspace{0.2em}} & \makebox[2.8em][r]{29.31}\makebox[2.2em][l]{\hspace{0.2em}} & \makebox[2.8em][r]{10.86}\makebox[2.2em][l]{\hspace{0.2em}} & \makebox[2.8em][r]{5.77}\makebox[2.2em][l]{\hspace{0.2em}} & \makebox[2.8em][r]{8.31}\makebox[2.2em][l]{\hspace{0.2em}} \\
\rowcolor{gray!8} \multicolumn{7}{c}{\textbf{Text-based}} \\
ReasoningBank & \makebox[2.8em][r]{45.36}\makebox[2.2em][l]{\hspace{0.2em}\scriptsize\textcolor{poscolor}{(+9.04)}} & \makebox[2.8em][r]{19.24}\makebox[2.2em][l]{\hspace{0.2em}\scriptsize\textcolor{negcolor}{(-3.07)}} & \makebox[2.8em][r]{32.30}\makebox[2.2em][l]{\hspace{0.2em}\scriptsize\textcolor{poscolor}{(+2.98)}} & \makebox[2.8em][r]{9.05}\makebox[2.2em][l]{\hspace{0.2em}\scriptsize\textcolor{negcolor}{(-1.81)}} & \makebox[2.8em][r]{7.09}\makebox[2.2em][l]{\hspace{0.2em}\scriptsize\textcolor{poscolor}{(+1.32)}} & \makebox[2.8em][r]{8.07}\makebox[2.2em][l]{\hspace{0.2em}\scriptsize\textcolor{negcolor}{(-0.24)}} \\
TrajectoryRetrieval & \makebox[2.8em][r]{43.90}\makebox[2.2em][l]{\hspace{0.2em}\scriptsize\textcolor{poscolor}{(+7.58)}} & \makebox[2.8em][r]{15.40}\makebox[2.2em][l]{\hspace{0.2em}\scriptsize\textcolor{negcolor}{(-6.91)}} & \makebox[2.8em][r]{29.65}\makebox[2.2em][l]{\hspace{0.2em}\scriptsize\textcolor{poscolor}{(+0.34)}} & \makebox[2.8em][r]{10.24}\makebox[2.2em][l]{\hspace{0.2em}\scriptsize\textcolor{negcolor}{(-0.62)}} & \makebox[2.8em][r]{7.11}\makebox[2.2em][l]{\hspace{0.2em}\scriptsize\textcolor{poscolor}{(+1.34)}} & \makebox[2.8em][r]{8.68}\makebox[2.2em][l]{\hspace{0.2em}\scriptsize\textcolor{poscolor}{(+0.36)}} \\
G-Memory & \makebox[2.8em][r]{37.35}\makebox[2.2em][l]{\hspace{0.2em}\scriptsize\textcolor{poscolor}{(+1.03)}} & \makebox[2.8em][r]{19.20}\makebox[2.2em][l]{\hspace{0.2em}\scriptsize\textcolor{negcolor}{(-3.11)}} & \makebox[2.8em][r]{28.27}\makebox[2.2em][l]{\hspace{0.2em}\scriptsize\textcolor{negcolor}{(-1.04)}} & \makebox[2.8em][r]{9.72}\makebox[2.2em][l]{\hspace{0.2em}\scriptsize\textcolor{negcolor}{(-1.14)}} & \makebox[2.8em][r]{6.92}\makebox[2.2em][l]{\hspace{0.2em}\scriptsize\textcolor{poscolor}{(+1.15)}} & \makebox[2.8em][r]{8.32}\makebox[2.2em][l]{\hspace{0.2em}\scriptsize\textcolor{poscolor}{(+0.01)}} \\
\rowcolor{gray!8} \multicolumn{7}{c}{\textbf{Multimodal-based}} \\
XSkill & \makebox[2.8em][r]{41.04}\makebox[2.2em][l]{\hspace{0.2em}\scriptsize\textcolor{poscolor}{(+4.72)}} & \makebox[2.8em][r]{20.81}\makebox[2.2em][l]{\hspace{0.2em}\scriptsize\textcolor{negcolor}{(-1.50)}} & \makebox[2.8em][r]{30.92}\makebox[2.2em][l]{\hspace{0.2em}\scriptsize\textcolor{poscolor}{(+1.61)}} & \makebox[2.8em][r]{11.43}\makebox[2.2em][l]{\hspace{0.2em}\scriptsize\textcolor{poscolor}{(+0.57)}} & \makebox[2.8em][r]{8.40}\makebox[2.2em][l]{\hspace{0.2em}\scriptsize\textcolor{poscolor}{(+2.63)}} & \makebox[2.8em][r]{9.91}\makebox[2.2em][l]{\hspace{0.2em}\scriptsize\textcolor{poscolor}{(+1.60)}} \\
M$^2$ & \makebox[2.8em][r]{43.59}\makebox[2.2em][l]{\hspace{0.2em}\scriptsize\textcolor{poscolor}{(+7.27)}} & \makebox[2.8em][r]{29.63}\makebox[2.2em][l]{\hspace{0.2em}\scriptsize\textcolor{poscolor}{(+7.32)}} & \makebox[2.8em][r]{36.61}\makebox[2.2em][l]{\hspace{0.2em}\scriptsize\textcolor{poscolor}{(+7.30)}} & \makebox[2.8em][r]{10.47}\makebox[2.2em][l]{\hspace{0.2em}\scriptsize\textcolor{negcolor}{(-0.39)}} & \makebox[2.8em][r]{4.72}\makebox[2.2em][l]{\hspace{0.2em}\scriptsize\textcolor{negcolor}{(-1.05)}} & \makebox[2.8em][r]{7.60}\makebox[2.2em][l]{\hspace{0.2em}\scriptsize\textcolor{negcolor}{(-0.72)}} \\
\rowcolor{gray!8} \multicolumn{7}{c}{\textbf{Automatic Design}} \\
ALMA & \makebox[2.8em][r]{45.73}\makebox[2.2em][l]{\hspace{0.2em}\scriptsize\textcolor{poscolor}{(+9.41)}} & \makebox[2.8em][r]{24.29}\makebox[2.2em][l]{\hspace{0.2em}\scriptsize\textcolor{poscolor}{(+1.98)}} & \makebox[2.8em][r]{35.01}\makebox[2.2em][l]{\hspace{0.2em}\scriptsize\textcolor{poscolor}{(+5.70)}} & \makebox[2.8em][r]{11.12}\makebox[2.2em][l]{\hspace{0.2em}\scriptsize\textcolor{poscolor}{(+0.26)}} & \makebox[2.8em][r]{7.38}\makebox[2.2em][l]{\hspace{0.2em}\scriptsize\textcolor{poscolor}{(+1.61)}} & \makebox[2.8em][r]{9.25}\makebox[2.2em][l]{\hspace{0.2em}\scriptsize\textcolor{poscolor}{(+0.94)}} \\
\textbf{\methodname} & \makebox[2.8em][r]{\textbf{51.84}}\makebox[2.2em][l]{\hspace{0.2em}\scriptsize\textcolor{poscolor}{(+15.52)}} & \makebox[2.8em][r]{\textbf{45.25}}\makebox[2.2em][l]{\hspace{0.2em}\scriptsize\textcolor{poscolor}{(+22.94)}} & \makebox[2.8em][r]{48.55}\makebox[2.2em][l]{\hspace{0.2em}\scriptsize\textcolor{poscolor}{(+19.23)}} & \makebox[2.8em][r]{\textbf{14.36}}\makebox[2.2em][l]{\hspace{0.2em}\scriptsize\textcolor{poscolor}{(+3.50)}} & \makebox[2.8em][r]{\textbf{11.46}}\makebox[2.2em][l]{\hspace{0.2em}\scriptsize\textcolor{poscolor}{(+5.69)}} & \makebox[2.8em][r]{\textbf{12.91}}\makebox[2.2em][l]{\hspace{0.2em}\scriptsize\textcolor{poscolor}{(+4.60)}} \\

\midrule
\multicolumn{7}{c}{\textit{GPT-5.4-nano}} \\
\midrule

NoMemory & \makebox[2.8em][r]{37.89}\makebox[2.2em][l]{\hspace{0.2em}} & \makebox[2.8em][r]{12.63}\makebox[2.2em][l]{\hspace{0.2em}} & \makebox[2.8em][r]{25.26}\makebox[2.2em][l]{\hspace{0.2em}} & \makebox[2.8em][r]{8.37}\makebox[2.2em][l]{\hspace{0.2em}} & \makebox[2.8em][r]{2.89}\makebox[2.2em][l]{\hspace{0.2em}} & \makebox[2.8em][r]{5.63}\makebox[2.2em][l]{\hspace{0.2em}} \\
\rowcolor{gray!8} \multicolumn{7}{c}{\textbf{Text-based}} \\
ReasoningBank & \makebox[2.8em][r]{35.51}\makebox[2.2em][l]{\hspace{0.2em}\scriptsize\textcolor{negcolor}{(-2.38)}} & \makebox[2.8em][r]{9.77}\makebox[2.2em][l]{\hspace{0.2em}\scriptsize\textcolor{negcolor}{(-2.86)}} & \makebox[2.8em][r]{22.64}\makebox[2.2em][l]{\hspace{0.2em}\scriptsize\textcolor{negcolor}{(-2.62)}} & \makebox[2.8em][r]{7.84}\makebox[2.2em][l]{\hspace{0.2em}\scriptsize\textcolor{negcolor}{(-0.53)}} & \makebox[2.8em][r]{3.49}\makebox[2.2em][l]{\hspace{0.2em}\scriptsize\textcolor{poscolor}{(+0.60)}} & \makebox[2.8em][r]{5.67}\makebox[2.2em][l]{\hspace{0.2em}\scriptsize\textcolor{poscolor}{(+0.04)}} \\
TrajectoryRetrieval & \makebox[2.8em][r]{34.82}\makebox[2.2em][l]{\hspace{0.2em}\scriptsize\textcolor{negcolor}{(-3.07)}} & \makebox[2.8em][r]{9.76}\makebox[2.2em][l]{\hspace{0.2em}\scriptsize\textcolor{negcolor}{(-2.87)}} & \makebox[2.8em][r]{22.29}\makebox[2.2em][l]{\hspace{0.2em}\scriptsize\textcolor{negcolor}{(-2.97)}} & \makebox[2.8em][r]{9.41}\makebox[2.2em][l]{\hspace{0.2em}\scriptsize\textcolor{poscolor}{(+1.04)}} & \makebox[2.8em][r]{3.68}\makebox[2.2em][l]{\hspace{0.2em}\scriptsize\textcolor{poscolor}{(+0.79)}} & \makebox[2.8em][r]{6.54}\makebox[2.2em][l]{\hspace{0.2em}\scriptsize\textcolor{poscolor}{(+0.92)}} \\
G-Memory & \makebox[2.8em][r]{36.50}\makebox[2.2em][l]{\hspace{0.2em}\scriptsize\textcolor{negcolor}{(-1.39)}} & \makebox[2.8em][r]{15.17}\makebox[2.2em][l]{\hspace{0.2em}\scriptsize\textcolor{poscolor}{(+2.54)}} & \makebox[2.8em][r]{25.84}\makebox[2.2em][l]{\hspace{0.2em}\scriptsize\textcolor{poscolor}{(+0.57)}} & \makebox[2.8em][r]{7.62}\makebox[2.2em][l]{\hspace{0.2em}\scriptsize\textcolor{negcolor}{(-0.75)}} & \makebox[2.8em][r]{1.84}\makebox[2.2em][l]{\hspace{0.2em}\scriptsize\textcolor{negcolor}{(-1.05)}} & \makebox[2.8em][r]{4.73}\makebox[2.2em][l]{\hspace{0.2em}\scriptsize\textcolor{negcolor}{(-0.90)}} \\
\rowcolor{gray!8} \multicolumn{7}{c}{\textbf{Multimodal-based}} \\
XSkill & \makebox[2.8em][r]{24.89}\makebox[2.2em][l]{\hspace{0.2em}\scriptsize\textcolor{negcolor}{(-13.00)}} & \makebox[2.8em][r]{6.95}\makebox[2.2em][l]{\hspace{0.2em}\scriptsize\textcolor{negcolor}{(-5.68)}} & \makebox[2.8em][r]{15.92}\makebox[2.2em][l]{\hspace{0.2em}\scriptsize\textcolor{negcolor}{(-9.34)}} & \makebox[2.8em][r]{5.92}\makebox[2.2em][l]{\hspace{0.2em}\scriptsize\textcolor{negcolor}{(-2.45)}} & \makebox[2.8em][r]{1.57}\makebox[2.2em][l]{\hspace{0.2em}\scriptsize\textcolor{negcolor}{(-1.32)}} & \makebox[2.8em][r]{3.75}\makebox[2.2em][l]{\hspace{0.2em}\scriptsize\textcolor{negcolor}{(-1.88)}} \\
M$^2$ & \makebox[2.8em][r]{38.33}\makebox[2.2em][l]{\hspace{0.2em}\scriptsize\textcolor{poscolor}{(+0.44)}} & \makebox[2.8em][r]{14.84}\makebox[2.2em][l]{\hspace{0.2em}\scriptsize\textcolor{poscolor}{(+2.21)}} & \makebox[2.8em][r]{26.59}\makebox[2.2em][l]{\hspace{0.2em}\scriptsize\textcolor{poscolor}{(+1.32)}} & \makebox[2.8em][r]{9.47}\makebox[2.2em][l]{\hspace{0.2em}\scriptsize\textcolor{poscolor}{(+1.10)}} & \makebox[2.8em][r]{4.72}\makebox[2.2em][l]{\hspace{0.2em}\scriptsize\textcolor{poscolor}{(+1.83)}} & \makebox[2.8em][r]{7.10}\makebox[2.2em][l]{\hspace{0.2em}\scriptsize\textcolor{poscolor}{(+1.47)}} \\
\rowcolor{gray!8} \multicolumn{7}{c}{\textbf{Automatic Design}} \\
ALMA & \makebox[2.8em][r]{37.47}\makebox[2.2em][l]{\hspace{0.2em}\scriptsize\textcolor{negcolor}{(-0.42)}} & \makebox[2.8em][r]{15.61}\makebox[2.2em][l]{\hspace{0.2em}\scriptsize\textcolor{poscolor}{(+2.98)}} & \makebox[2.8em][r]{26.54}\makebox[2.2em][l]{\hspace{0.2em}\scriptsize\textcolor{poscolor}{(+1.28)}} & \makebox[2.8em][r]{9.13}\makebox[2.2em][l]{\hspace{0.2em}\scriptsize\textcolor{poscolor}{(+0.76)}} & \makebox[2.8em][r]{3.76}\makebox[2.2em][l]{\hspace{0.2em}\scriptsize\textcolor{poscolor}{(+0.87)}} & \makebox[2.8em][r]{6.45}\makebox[2.2em][l]{\hspace{0.2em}\scriptsize\textcolor{poscolor}{(+0.82)}} \\
\textbf{\methodname} & \makebox[2.8em][r]{\textbf{42.30}}\makebox[2.2em][l]{\hspace{0.2em}\scriptsize\textcolor{poscolor}{(+4.41)}} & \makebox[2.8em][r]{\textbf{18.65}}\makebox[2.2em][l]{\hspace{0.2em}\scriptsize\textcolor{poscolor}{(+6.02)}} & \makebox[2.8em][r]{\textbf{30.47}}\makebox[2.2em][l]{\hspace{0.2em}\scriptsize\textcolor{poscolor}{(+5.21)}} & \makebox[2.8em][r]{\textbf{10.95}}\makebox[2.2em][l]{\hspace{0.2em}\scriptsize\textcolor{poscolor}{(+2.58)}} & \makebox[2.8em][r]{\textbf{5.40}}\makebox[2.2em][l]{\hspace{0.2em}\scriptsize\textcolor{poscolor}{(+2.51)}} & \makebox[2.8em][r]{\textbf{8.18}}\makebox[2.2em][l]{\hspace{0.2em}\scriptsize\textcolor{poscolor}{(+2.55)}} \\
\bottomrule
\end{tabular}
\end{adjustbox}
\end{table*}

\subsection{Main Results}

Table~\ref{tab:main_result} compares \methodname with text-based, multimodal-based, and automatic memory design baselines across GUI-agent and visual-reasoning benchmarks. 
We report the main results on Qwen3-VL-32B and GPT-5.4-nano in the main text, and provide the results on Qwen3.5-Plus in Appendix~\ref{app:qwen35_results}.

\noindent{\textbf{\methodname consistently improves multimodal agents.}}
Across both execution agents, \methodname achieves the best average performance on GUI navigation and visual reasoning tasks. 
Compared with the NoMemory baseline, \methodname improves AVG. GUI by 19.23 and 5.21 points on Qwen3-VL-32B and GPT-5.4-nano, respectively, and improves AVG. VR by 4.60 and 2.55 points. 
These results show that automatically discovered memory mechanisms can effectively transform past multimodal trajectories into useful task context.

\noindent{\textbf{Automatic memory design is more robust than fixed memory schemas.}}
Existing memory methods often improve certain benchmarks but degrade others, indicating that manually designed memory formats are sensitive to task type and execution model. 
For example, several text-based and multimodal-based baselines improve individual GUI tasks but fail to consistently benefit visual reasoning. 
In contrast, \methodname achieves stable gains across both task categories, demonstrating the advantage of adapting the memory mechanism itself rather than relying on fixed experience representations or retrieval rules.

\begin{figure}[t]
    \centering
    \includegraphics[width=\linewidth]{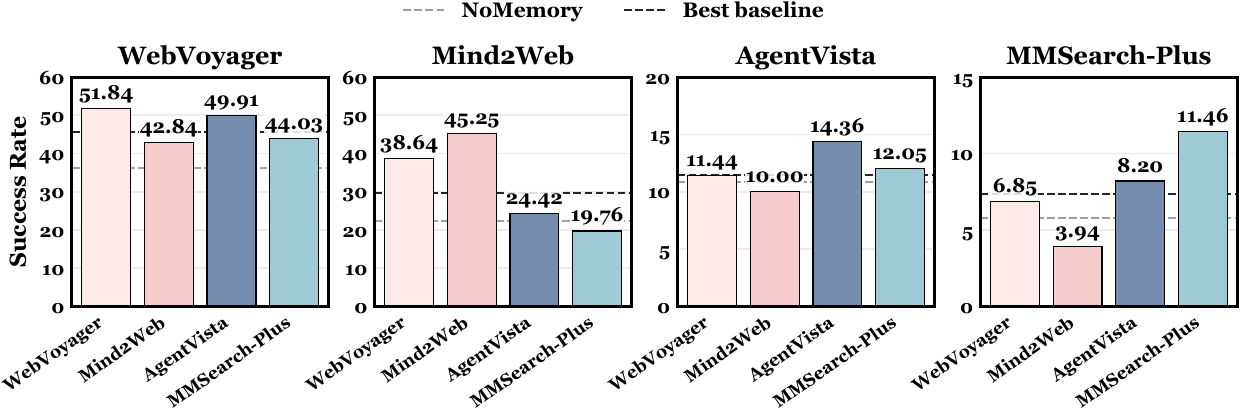}
\caption{
Cross-task transferability of the searched memory under Qwen3-VL-32B as the execution agent. 
Each subplot corresponds to a target benchmark, and each bar reports the performance obtained by applying the memory design searched on a source benchmark to that target benchmark.
}
    \label{fig:cross_task}
    \vspace{-0.3cm}
\end{figure}

\subsection{Transfer Analysis}

\begin{wrapfigure}{r}{0.5\textwidth}
\vspace{1.5mm}
    \centering
    \includegraphics[width=\linewidth]{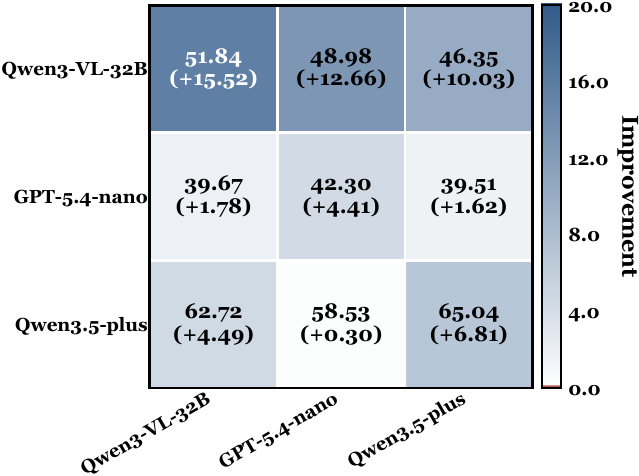}
    \vspace{-2mm}
\caption{
Cross-model memory transfer on the WebVoyager benchmark. 
Rows indicate evaluation models and columns indicate the models used to search memory designs.
}
    \label{fig:cross_model}
    \vspace{-10mm}
\end{wrapfigure}

We further examine whether the searched memory transfer across benchmarks and execution models. 
As shown in Figures~\ref{fig:cross_task} and~\ref{fig:cross_model}, the learned memory mechanisms exhibit clear transferability, but the strongest performance is typically achieved when the memory is searched under the target setting.

\noindent{\textbf{Cross-benchmark transfer.}}
Figure~\ref{fig:cross_task} evaluates memory designs searched on one benchmark and applied to other benchmarks, using Qwen3-VL-32B as the execution agent. 
The results show that transferred memory designs often remain competitive and can outperform both NoMemory and the best fixed baseline. 
However, the best results generally appear on the diagonal, where the search and evaluation benchmarks match. 
This indicates that the discovered memory mechanisms capture reusable multimodal experience structures, but still need to adapt to benchmark-specific requirements, such as GUI interaction patterns or visual-evidence organization.

\noindent{\textbf{Cross-model transfer.}}
Figure~\ref{fig:cross_model} evaluates transfer across execution models on WebVoyager. Each cell reports task performance, with the improvement over the corresponding NoMemory baseline in parentheses. 
Off-diagonal results evaluate whether a searched memory design can generalize across execution models.
All off-diagonal entries improve over the corresponding NoMemory baselines, showing that searched memory designs are not tied to a single backbone. 
Nevertheless, diagonal entries remain the strongest for all models, suggesting that different execution agents benefit from different memory behaviors due to variations in visual grounding, instruction following, and retrieval usage. 
Together, these results show that \methodname learns memory designs with both reusable and target-adaptive components.

\subsection{Cost Analysis}

Table~\ref{tab:cost_analysis} analyzes the inference-time cost of different memory settings on Qwen3-VL-32B, including task performance, token consumption, and interaction steps. 
The results show that the gains of \methodname do not come from using more context or longer trajectories.

\noindent{\textbf{\methodname improves the performance--cost trade-off.}}
\methodname achieves the best performance on all four benchmarks while keeping token usage and interaction steps among the lowest. 
For example, on WebVoyager, it improves success rate from 36.32 to 51.84, while reducing token consumption from 307.0K to 98.0K and interaction steps from 11.31 to 9.66. 
Similar trends appear on Mind2Web, AgentVista, and MMSearch-Plus, indicating that \methodname improves execution efficiency by retrieving more task-relevant memory rather than injecting more context.

\noindent{\textbf{Fixed memory designs often introduce inefficient context.}}
Prior memory methods do not consistently improve the performance--cost trade-off. 
Several baselines consume more tokens than NoMemory but yield limited gains or even performance drops, suggesting that fixed memory schemas may introduce noisy or redundant context. 
Conversely, lower cost alone is also insufficient: M$^2$ often uses very few tokens or steps, but its performance remains below \methodname. 
Overall, \methodname achieves a stronger Pareto trade-off between accuracy and inference efficiency.

\noindent{\textbf{Search-time overhead.}}
The above analysis focuses on inference-time cost. 
We report offline search-time overhead in Appendix~\ref{app:search_cost}; under the same 20-step sequential search budget as ALMA, \methodname uses fewer total search-time tokens and less wall-clock time, showing that its gains are not obtained from a larger offline design budget.

\begin{table*}[t]
\centering
\caption{
Cost and efficiency analysis under Qwen3-VL-32B. 
Perf. denotes the final task performance, Tok. reports the average number of tokens per episode in thousands, and \#Steps reports the average number of agent interaction steps.
Bold and underline indicate the best and second-best results within each benchmark and metric, respectively.
}
\label{tab:cost_analysis}
\resizebox{\textwidth}{!}{%
\begin{tabular}{lcccccccccccc}
\toprule
\multirow{2}{*}{\textbf{Memory Setting}} & \multicolumn{3}{c}{\textbf{WebVoyager}} & \multicolumn{3}{c}{\textbf{Mind2Web}} & \multicolumn{3}{c}{\textbf{AgentVista}} & \multicolumn{3}{c}{\textbf{MMSearch-Plus}} \\
\cmidrule(lr){2-4} \cmidrule(lr){5-7} \cmidrule(lr){8-10} \cmidrule(lr){11-13}
 & Perf. $\uparrow$ & Tok. $\downarrow$ & \#Steps $\downarrow$ & Perf. $\uparrow$ & Tok. $\downarrow$ & \#Steps $\downarrow$ & Perf. $\uparrow$ & Tok. $\downarrow$ & \#Steps $\downarrow$ & Perf. $\uparrow$ & Tok. $\downarrow$ & \#Steps $\downarrow$ \\
\midrule
NoMemory & 36.32 & 307.0 & 11.31 & 22.31 & 112.4 & 12.92 & 10.86 & 26.2 & 4.62 & 5.77 & 30.4 & 4.49 \\
ReasoningBank & 45.36 & \underline{126.4} & 10.45 & 19.24 & 157.2 & 13.39 & 9.05 & 36.4 & 4.61 & 7.09 & 44.1 & 4.57 \\
TrajectoryRetrieval & 43.90 & 145.0 & \underline{10.13} & 15.40 & 179.3 & 13.58 & 10.24 & 48.5 & 4.72 & 7.11 & 54.7 & 4.57 \\
G-Memory & 37.35 & 180.2 & 10.91 & 19.20 & 209.3 & 13.13 & 9.72 & 43.9 & 3.87 & 6.92 & 75.2 & 5.00 \\
XSkill & 41.04 & 150.7 & 10.60 & 20.81 & 168.2 & 13.17 & \underline{11.43} & 60.5 & 4.65 & \underline{8.40} & 68.2 & 4.87 \\
M2 & 43.59 & 148.7 & 10.54 & \underline{29.63} & \textbf{88.4} & \underline{12.23} & 10.47 & \textbf{8.1} & \textbf{2.25} & 4.72 & \textbf{12.8} & \textbf{2.50} \\
ALMA & \underline{45.73} & 137.4 & 10.23 & 24.29 & 143.3 & \textbf{12.21} & 11.12 & 41.1 & 3.91 & 7.38 & 38.2 & \underline{3.28} \\
\textbf{\methodname} & \textbf{51.84} & \textbf{98.0} & \textbf{9.66} & \textbf{45.25} & \underline{94.9} & 12.72 & \textbf{14.36} & \underline{22.6} & \underline{3.26} & \textbf{11.46} & \underline{13.4} & 3.53 \\
\bottomrule
\end{tabular}%
}
\end{table*}

\begin{figure}[t]
    \centering
    \includegraphics[width=\linewidth]{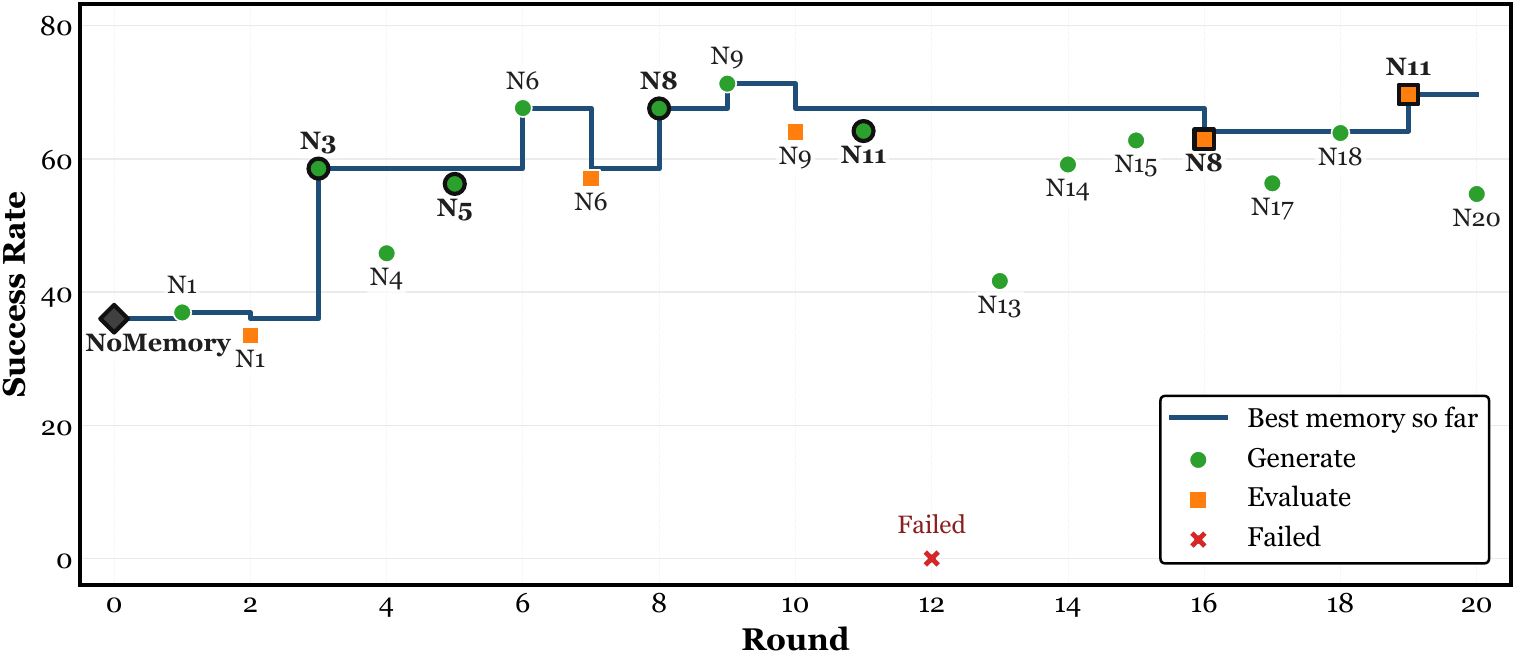}
\caption{
Memory-design search trajectory for the \textsc{WebVoyager} search run under \textsc{Qwen3-VL-32B}. 
Each point represents a memory design evaluated or re-evaluated at a search round, and the node labels correspond to the search tree in Appendix~\ref{app:memory_evolution}. 
The dashed curve tracks the best validation performance found so far. 
Failed rounds denote candidates rejected by quick examination before full evaluation. 
The final memory program selected by the LCB rule is N11.
}
    \label{fig:memory_evolution}
    \vspace{-0.3cm}
\end{figure}

\subsection{Search Process and Memory Evolution}

We further examine how \methodname improves memory designs during search. 
Figure~\ref{fig:memory_evolution} shows a representative search trajectory on \textsc{WebVoyager} with \textsc{Qwen3-VL-32B}, where node labels correspond to the search tree in Appendix~\ref{app:memory_evolution}; Appendix~\ref{app:learned_memo_programs} visualizes the final selected memo programs across benchmarks. 
Starting from NoMemory, \methodname quickly discovers stronger memory programs, and the best validation performance continues to improve as search proceeds. 
The final program selected by the LCB rule is N11.

The trajectory also shows that search is not monotonic: some generated candidates underperform the current best program, and one candidate fails the quick examination before full evaluation. 
This indicates that \methodname explores diverse memory designs rather than merely applying greedy local refinements. 
Through evaluation feedback and reflection-guided mutation, the search gradually identifies more task-conditioned memory mechanisms that organize relevant experience, filter noisy context, and provide useful guidance to the execution agent.

Additional ablation, human-evaluation, and memory-operation latency results are provided in Appendices~\ref{app:additional_experiments} and~\ref{app:additional_cost}.

\section{Conclusion}
\label{sec:conclusion}

In this paper, we studied how multimodal agents can better learn from their own experience by treating memory design as an adaptive optimization problem rather than a fixed human-designed component. We introduced \methodname, which represents memory mechanisms as executable memo programs and searches for effective update and retrieval strategies through evaluation, reflection, mutation, and budget-aware tree search.
Experiments on GUI/web navigation and multimodal visual reasoning benchmarks show that \methodname consistently outperforms no-memory agents, manually designed memory baselines, and text-only automatic memory-design methods. Further analyses demonstrate that the learned memory designs improve not only task performance, but also inference efficiency, often reducing token usage and interaction steps. Transfer and search-process results further indicate that \methodname discovers reusable yet task-adaptive memory structures.
Overall, our findings highlight memory design as a key optimization target for multimodal agents. Beyond improving foundation models or enlarging experience buffers, effective agent learning requires adaptive mechanisms for deciding how experience should be represented, organized, and reused.

\begin{ack}
We thank the anonymous reviewers for their comments and suggestions. This work was supported by the National Natural Science Foundation of China (NSFC) via grant 62441614 and 62576125.
\end{ack}


\bibliography{natbib}
\bibliographystyle{unsrtnat}

\appendix

\section{Limitations}
\label{app:limitations}

Although \methodname demonstrates consistent gains across GUI/web navigation and multimodal visual reasoning benchmarks, several limitations remain.

First, our experiments are conducted on a finite set of benchmark environments. 
While these tasks cover diverse multimodal interaction patterns, they do not fully represent more open-ended deployment scenarios, such as long-horizon embodied interaction, multi-user environments, or continuously changing task distributions. 
Evaluating whether the searched memo programs remain effective in such non-stationary settings is an important direction for future work.

Second, our current protocol optimizes memory design in an offline setting. 
For each benchmark, memory construction and search are performed on the training split, and the resulting memory system is then evaluated on held-out tasks without further adaptation. 
This controlled setup enables fair comparison across candidate memory designs, but it does not fully capture continual deployment settings where new experiences arrive over time. 
Extending \methodname to support stable online memory evolution remains an open challenge.

Third, although the final learned memory systems can reduce inference-time token usage and interaction steps, the search process itself still requires multiple evaluations of candidate memo programs. 
This offline cost may become significant when environment interaction is expensive or when much larger execution agents are used. 
Future work could improve search efficiency through surrogate evaluation, partial trajectory reuse, adaptive early stopping, or stronger candidate filtering.

Finally, multimodal memory introduces privacy and safety considerations. 
Stored trajectories may contain screenshots, user-provided content, web pages, or other sensitive information. 
Moreover, because memo programs are executable, practical deployment should include sandboxing, permission control, resource limits, and mechanisms for inspecting, filtering, or deleting stored memories.

\section{Impact Statement}
\label{app:impact_statement}

This work aims to improve how multimodal agents store, organize, and retrieve past experience. 
Its potential positive impact is to make agents more reliable and efficient by reducing their dependence on manually engineered memory schemas. 
Adaptive memory mechanisms may help agents reuse prior experience, avoid repeated mistakes, and reduce unnecessary context consumption in applications such as web navigation assistance, visual search, accessibility tools, and interactive tutoring.

At the same time, stronger memory-augmented agents introduce potential risks. 
If multimodal trajectories are stored without sufficient safeguards, they may preserve sensitive user information, private visual content, or misleading historical context. 
Retrieved memories could also bias future behavior if they are stale, irrelevant, or improperly filtered. 
In addition, more persistent agent memory may increase the capability of autonomous systems in web environments, which could be misused for harmful automation.

Responsible deployment therefore requires privacy-preserving storage, user control over memory retention and deletion, careful filtering of sensitive content, and transparent logging of memory usage. 
Because \methodname searches over executable memo programs, deployment should also require sandboxed execution, restricted permissions, and validation before a searched memory mechanism is used outside controlled benchmark settings.

\section{Additional Algorithm Details}
\label{app:algorithm}

This appendix provides the full training procedure of \methodname. 
The main text introduces the memo-program abstraction, the update-then-retrieve evaluation protocol, reflection-guided mutation, and the budget-aware tree search rule. 
Here, we give the complete pseudocode and further implementation details. 
In our experiments, both the update batch \(\mathcal{D}_u\) and the retrieve batch \(\mathcal{D}_r\) used during memory-design search are drawn from the training split; the held-out test split is used only for final offline evaluation.

\subsection{Full Training Procedure}

Algorithm~\ref{alg:method} summarizes the overall search procedure. 
The search tree stores candidate memo programs, where each node corresponds to one executable memo program and each edge corresponds to a reflection-guided mutation from a parent program. 
At each search round, \methodname either re-evaluates an existing node to reduce score uncertainty or generates a new child from an expandable parent. 
Failed generations that do not pass quick examination are discarded and do not consume full-evaluation budget.

\begin{algorithm}[t]
\caption{\methodname}
\label{alg:method}
\begin{algorithmic}[1]
\Require Update batch \(\mathcal{D}_u\), retrieve batch \(\mathcal{D}_r\), root memo program \(m_0\), search budget \(T\), constants \(c_e,c_g,\rho,\beta,B\), repair budget \(L\)
\State Initialize search tree \(\mathcal{T}\leftarrow\{0\}\)
\State Set \(\mathrm{Parent}(0)\leftarrow\varnothing\), \(\mathrm{Children}(0)\leftarrow\varnothing\)
\State Set \(K_0\leftarrow 0\), \(S_0\leftarrow 0\)
\State \(r_0\leftarrow\textsc{FullEval}(m_0,\mathcal{D}_u,\mathcal{D}_r)\)
\State Set \(\hat{\mu}_0\leftarrow r_0\), \(n_0\leftarrow 1\), \(N\leftarrow 1\)
\For{\(t=1,\ldots,T\)}
    \State Compute expandable set \(\mathcal{E}\) using Eq.~\eqref{eq:expandable-set-app}
    \State Initialize action set \(\mathcal{A}_t\leftarrow\varnothing\)
    \For{each node \(i\in\mathcal{T}\)}
        \State Compute \(\mathrm{UCB}^{\mathrm{eval}}_i\) using Eq.~\eqref{eq:ucb-eval-app}
        \State Add \((\textsc{Evaluate},i,\mathrm{UCB}^{\mathrm{eval}}_i)\) to \(\mathcal{A}_t\)
    \EndFor
    \For{each node \(i\in\mathcal{E}\)}
        \State Compute \(\hat{\Delta}_i\) using Eq.~\eqref{eq:delta-hat-app}
        \State Compute \(\mathrm{UCB}^{\mathrm{gen}}_i\) using Eq.~\eqref{eq:ucb-gen-app}
        \State Add \((\textsc{Generate},i,\mathrm{UCB}^{\mathrm{gen}}_i)\) to \(\mathcal{A}_t\)
    \EndFor
    \State Select \((a_t,i_t,\cdot)\leftarrow\arg\max_{(a,i,u)\in\mathcal{A}_t}u\)
    \If{\(a_t=\textsc{Evaluate}\)}
        \State \(r\leftarrow\textsc{FullEval}(m_{i_t},\mathcal{D}_u,\mathcal{D}_r)\)
        \State \(\hat{\mu}_{i_t}\leftarrow(n_{i_t}\hat{\mu}_{i_t}+r)/(n_{i_t}+1)\)
        \State \(n_{i_t}\leftarrow n_{i_t}+1\), \(N\leftarrow N+1\)
    \Else
        \State \(\hat{\mu}^{\mathrm{snap}}_{i_t}\leftarrow\hat{\mu}_{i_t}\)
        \State \(f_{i_t}\leftarrow\textsc{Reflect}(m_{i_t})\)
        \State \(m_j\leftarrow\textsc{MutateAndRepair}(m_{i_t},f_{i_t},L)\)
        \If{\(m_j\) passes quick examination}
            \State \(r_j\leftarrow\textsc{FullEval}(m_j,\mathcal{D}_u,\mathcal{D}_r)\)
            \State Add a new node \(j\) to \(\mathcal{T}\)
            \State \(\mathrm{Parent}(j)\leftarrow i_t\)
            \State \(\mathrm{Children}(i_t)\leftarrow\mathrm{Children}(i_t)\cup\{j\}\)
            \State Initialize \(\mathrm{Children}(j)\leftarrow\varnothing\)
            \State Set \(\hat{\mu}_j\leftarrow r_j\), \(n_j\leftarrow 1\), \(K_j\leftarrow 0\), \(S_j\leftarrow 0\)
            \State \(\Delta_{i_t\rightarrow j}\leftarrow\max(0,\hat{\mu}_j-\hat{\mu}^{\mathrm{snap}}_{i_t})\)
            \State \(S_{i_t}\leftarrow S_{i_t}+\Delta_{i_t\rightarrow j}\)
            \State \(K_{i_t}\leftarrow K_{i_t}+1\), \(N\leftarrow N+1\)
        \EndIf
    \EndIf
\EndFor
\State \Return \(m^*=\arg\max_{i\in\mathcal{T}}\mathrm{LCB}^{\mathrm{eval}}_i\)
\end{algorithmic}
\end{algorithm}

\subsection{Full Evaluation}

The procedure \(\textsc{FullEval}(m,\mathcal{D}_u,\mathcal{D}_r)\) evaluates a memo program under the update-then-retrieve protocol. 
Before evaluation, the internal memory state of \(m\) is reset. 
The program then processes the update batch by calling \(m.\mathrm{update}(e_\tau)\) for each update episode recorder \(e_\tau\in\mathcal{D}_u\). 
After all update episodes have been processed, the memory state is frozen.

For each retrieve-time task \(x\in\mathcal{D}_r\), the memo program receives a partial recorder \(e_x\) containing the task input and initial observations. 
It returns a memory payload \(p_x=m.\mathrm{retrieve}(e_x)\). 
The payload is truncated according to the fixed retrieval budget and then injected into the execution agent. 
The evaluator returns a scalar reward \(R_x\in[0,1]\). 
The full-evaluation score is
\begin{equation}
    \textsc{FullEval}(m,\mathcal{D}_u,\mathcal{D}_r)
    =
    \frac{1}{|\mathcal{D}_r|}
    \sum_{x\in\mathcal{D}_r} R_x .
    \label{eq:fulleval-app}
\end{equation}

During full evaluation, \methodname records the retrieved payload, retrieved images, execution trajectory, screenshots, intermediate observations, errors, and final outcome. 
These records are used by the reflection step to diagnose how the current memory mechanism affects downstream agent behavior.

\subsection{Reflection, Mutation, and Repair}

The reflection step analyzes an evaluated memo program using both its source code and its execution evidence. 
Given a node \(i\), \(\textsc{Reflect}(m_i)\) receives the memo program, its empirical score, retrieved memory payloads, representative task trajectories, screenshots, retrieved images, and error cases. 
It outputs a structured feedback object \(f_i\). 
This feedback identifies failure modes such as missing memories, irrelevant retrievals, redundant payloads, misleading examples, weak text-image alignment, unstable retrieval keys, insufficient visual summarization, or excessive truncation.

The mutation step then produces a child program conditioned on the parent program and the feedback:
\begin{equation}
    m_j \sim \textsc{Mutate}(m_i,f_i).
    \label{eq:mutate-app}
\end{equation}
The generated child may change the memory schema, trajectory summarization rule, text-image embedding strategy, indexing structure, retrieval ranking function, metadata filtering rule, or payload formatting strategy. 
However, it must preserve the same update and retrieve interface as all other memo programs.

Before full evaluation, the child program is checked by quick examination. 
If the program fails, a repair model is invoked with the quick-examination report and the failed test case. 
The repair process is repeated for at most \(L\) attempts. 
A child that still fails after repair is discarded and is not added to the search tree.

\begin{algorithm}[t]
\caption{\(\textsc{MutateAndRepair}\)}
\label{alg:mutate-repair}
\begin{algorithmic}[1]
\Require Parent memo program \(m_i\), reflection feedback \(f_i\), repair budget \(L\)
\State \(m\leftarrow\textsc{Mutate}(m_i,f_i)\)
\For{\(\ell=0,\ldots,L\)}
    \State \(q\leftarrow\textsc{QuickExam}(m)\)
    \If{\(q=\textsc{Pass}\)}
        \State \Return \(m\)
    \EndIf
    \If{\(\ell<L\)}
        \State \(m\leftarrow\textsc{Repair}(m,q)\)
    \EndIf
\EndFor
\State \Return invalid program
\end{algorithmic}
\end{algorithm}

\subsection{Quick Examination}

Quick examination is a lightweight validation step that prevents invalid programs from consuming full-evaluation budget. 
It checks whether the generated program:
\begin{itemize}
    \item imports successfully;
    \item implements the required update and retrieve interface;
    \item runs on sample update episodes without errors;
    \item runs on sample retrieve-time inputs without errors;
    \item returns a schema-valid memory payload;
    \item respects the maximum payload length and maximum number of retrieved images.
\end{itemize}

Only programs that pass quick examination are eligible for full evaluation and insertion into the generation tree.

\subsection{Generation Statistics}

For each node \(i\), \methodname maintains \(K_i\), the number of evaluated children generated from \(i\), and \(S_i\), the cumulative positive improvement obtained by these children. 
When a generated child \(j\) receives score \(\hat{\mu}_j\), its positive improvement over the parent snapshot is
\begin{equation}
    \Delta_{i\rightarrow j}
    =
    \max\left(0,\hat{\mu}_j-\hat{\mu}^{\mathrm{snap}}_i\right),
    \label{eq:delta-child-app}
\end{equation}
where \(\hat{\mu}^{\mathrm{snap}}_i\) denotes the empirical score of the parent at the time the child is generated. 
The snapshot is used so that a child improvement remains fixed even if the parent is later re-evaluated.

For rarely expanded nodes, we use a root-relative prior:
\begin{equation}
    \Delta_i^{\mathrm{prior}}
    =
    \rho\max\left(0,\hat{\mu}_i-\hat{\mu}_0\right),
    \label{eq:delta-prior-app}
\end{equation}
where \(m_0\) is the no-memory root and \(\rho\) controls the prior strength. 
The estimated local improvement potential of node \(i\) is
\begin{equation}
    \hat{\Delta}_i
    =
    \frac{
        \beta\Delta_i^{\mathrm{prior}}+S_i
    }{
        \beta+K_i
    }.
    \label{eq:delta-hat-app}
\end{equation}
Here, \(\beta\) is a pseudo-count. 
When \(K_i\) is small, the estimate is influenced by the root-relative prior; as more children are generated, it becomes dominated by observed improvements.

\subsection{Action Scores and Expansion Constraint}

For an existing node \(i\), the evaluation action is scored by
\begin{equation}
    \mathrm{UCB}^{\mathrm{eval}}_i
    =
    \hat{\mu}_i
    +
    c_e\sqrt{\frac{\log N}{n_i}},
    \label{eq:ucb-eval-app}
\end{equation}
where \(N\) is the number of full evaluations consumed so far. 
This action reduces uncertainty in the empirical quality of existing candidates.

For an expandable node \(i\), the generation action is scored by
\begin{equation}
    \mathrm{UCB}^{\mathrm{gen}}_i
    =
    \hat{\mu}_i
    +
    \hat{\Delta}_i
    +
    c_g\sqrt{\frac{\log N}{\beta+K_i}}.
    \label{eq:ucb-gen-app}
\end{equation}
This score combines the current quality of the parent, its estimated local improvement potential, and an exploration bonus for under-expanded parents.

The set of expandable nodes is
\begin{equation}
    \mathcal{E}
    =
    \{0\}
    \cup
    \left\{
    i\in\mathcal{T}\setminus\{0\}
    :
    |\mathrm{Children}(\mathrm{Parent}(i))|\ge B
    \right\}.
    \label{eq:expandable-set-app}
\end{equation}
Thus, the root is always expandable, while a non-root node becomes expandable only after its parent has generated at least \(B\) children. 
This minimum-width constraint encourages sibling diversity before the search grows deeper.

At each round, the selected action is
\begin{equation}
    (a_t,i_t)
    =
    \arg\max_{(a,i,u)\in\mathcal{A}_t} u ,
    \label{eq:action-select-app}
\end{equation}
where \(\mathcal{A}_t\) contains all evaluation actions and all eligible generation actions.

\subsection{Robust Final Selection}

After the search budget is exhausted, \methodname selects the final memo program using a lower confidence bound rather than the raw empirical score. 
For each node \(i\), define
\begin{equation}
    \mathrm{LCB}^{\mathrm{eval}}_i
    =
    \hat{\mu}_i
    -
    c_e\sqrt{\frac{\log N}{n_i}}.
    \label{eq:lcb-eval-app}
\end{equation}
The returned program is
\begin{equation}
    m^*
    =
    \arg\max_{i\in\mathcal{T}}
    \mathrm{LCB}^{\mathrm{eval}}_i .
    \label{eq:final-select-app}
\end{equation}
This criterion reduces the chance of selecting a candidate whose empirical score is high only because it has been evaluated too few times.

\section{Dataset Details}
\label{app:dataset_details}

For benchmarks without an official training split, we construct train/test splits with an approximately 1:4 ratio while preserving the distribution of task categories, domains, or websites within each benchmark. 
For Mind2Web, whose original scale is substantially larger than the other benchmarks, we sample websites to match the overall evaluation budget. Specifically, we select cross-task examples from the same websites as the sampled training split, and additionally sample websites from the cross-website and cross-domain splits. 
Across the four benchmarks, the resulting splits contain 304 training tasks and 1,231 held-out test tasks in total.

For all benchmarks, the training split is used for both memory construction and memory-design search, while the held-out test split is used only for final offline evaluation. During memory-design search, the training split is further divided into an update batch \(\mathcal{D}_u\) and a search-validation retrieve batch \(\mathcal{D}_r\). Candidate memo programs are evaluated on \(\mathcal{D}_r\) only for search-time model selection, and no held-out test examples are used for memory construction, memory-design search, or candidate selection. All statistics reported below are computed from the final train/test split files used in our experiments. Aggregate files and scratch files are excluded from the reported statistics.

\subsection{Overall Statistics}

Table~\ref{tab:dataset_details} summarizes the resulting train/test splits.
Across the four benchmarks, our splits contain 304 training tasks and 1,231 held-out test tasks.
The number of domains or sites is computed from the union of the train and test files used for each benchmark.

\begin{table}[t]
\centering
\caption{Benchmark statistics. The train split is used for memory construction and design search, while the held-out test split is used only for final offline evaluation.}
\label{tab:dataset_details}
\begin{tabular}{lccc}
\toprule
Dataset & Train & Test & Domains/Sites \\
\midrule
AgentVista & 39 & 170 & 7 \\
MMSearch-Plus & 59 & 252 & 8 \\
Mind2Web & 108 & 393 & 22 \\
WebVoyager & 98 & 416 & 12 \\
\midrule
Total & 304 & 1,231 & -- \\
\bottomrule
\end{tabular}
\end{table}

\subsection{Split Construction}

For AgentVista, MMSearch-Plus, and WebVoyager, we group tasks by their task category, domain, or website, and sample approximately 20\% of the tasks within each group for training.
The remaining tasks are used for held-out testing.
This stratified procedure keeps the train/test distributions comparable while ensuring that final evaluation is performed on tasks that are not directly used during memory construction or design search.

Mind2Web provides three evaluation settings: cross-task, cross-website, and cross-domain.
Since its original scale is substantially larger than the other benchmarks, we sample websites to match the overall evaluation budget.
We include cross-task examples from websites that also appear in the sampled training split, and additionally sample websites from the cross-website and cross-domain settings.
This preserves the intended generalization structure of Mind2Web while keeping the final evaluation scale comparable to the other benchmarks.

\subsection{Mind2Web Evaluation Settings}

Table~\ref{tab:mind2web_split_sites} reports the sampled Mind2Web sites under the three evaluation settings.
The cross-task setting evaluates tasks from websites that are also observed in the training split, while the cross-website and cross-domain settings evaluate generalization to unseen websites or domains.

\begin{table}[t]
\centering
\caption{Mind2Web sampled sites by evaluation setting.}
\label{tab:mind2web_split_sites}
\begin{tabular}{llc}
\toprule
Setting & Site & Tasks \\
\midrule
\multirow{5}{*}{\begin{tabular}[c]{@{}l@{}}cross-task\\(26 tasks, 5 sites)\end{tabular}}
& budget & 10 \\
& newegg & 7 \\
& ticketcenter & 4 \\
& spothero & 3 \\
& resy & 2 \\
\midrule
\multirow{6}{*}{\begin{tabular}[c]{@{}l@{}}cross-website\\(109 tasks, 6 sites)\end{tabular}}
& tripadvisor & 23 \\
& recreation.gov & 20 \\
& bestbuy & 17 \\
& trip & 17 \\
& macys & 16 \\
& stubhub & 16 \\
\midrule
\multirow{11}{*}{\begin{tabular}[c]{@{}l@{}}cross-domain\\(258 tasks, 11 sites)\end{tabular}}
& reddit & 33 \\
& babycenter & 28 \\
& thumbtack & 24 \\
& drugs & 23 \\
& finance.yahoo & 23 \\
& healthline & 23 \\
& healthgrades & 22 \\
& student & 22 \\
& coinmarketcap & 21 \\
& webmd & 20 \\
& akc.org & 19 \\
\bottomrule
\end{tabular}
\end{table}

\subsection{Domain and Site Distributions}

Table~\ref{tab:domain_site_distributions} reports the main domain or website distributions after combining the train and test splits used in our experiments.
For AgentVista and MMSearch-Plus, the entries correspond to high-level domains.
For Mind2Web and WebVoyager, the entries correspond to websites, and we list the top websites by number of tasks.
These distributions are used only for split construction and dataset analysis; they are not provided as task inputs to the agent.

\begin{table}[t]
\centering
\caption{Domain and site distributions of the sampled benchmark splits.}
\label{tab:domain_site_distributions}
\resizebox{\linewidth}{!}{
\begin{tabular}{llr@{\qquad}|llr}
\toprule
Dataset & Domain/Site & Tasks & Dataset & Domain/Site & Tasks \\
\midrule
\multirow{8}{*}{AgentVista}
& commerce & 42
& \multirow{8}{*}{MMSearch-Plus}
& Geography & 64 \\
& geography & 39
& & Sports & 54 \\
& entertainment & 39
& & Academic Research & 50 \\
& technology & 34
& & Film \& TV & 40 \\
& society & 25
& & Technology & 36 \\
& academics & 15
& & Video Games & 31 \\
& culture & 15
& & Vlog & 19 \\
& -- & --
& & Music & 17 \\
\midrule
\multirow{10}{*}{Mind2Web}
& budget & 34
& \multirow{10}{*}{WebVoyager}
& Wolfram Alpha & 46 \\
& reddit & 33
& & Allrecipes & 45 \\
& babycenter & 28
& & ESPN & 44 \\
& newegg & 28
& & Apple & 43 \\
& ticketcenter & 26
& & Cambridge Dictionary & 43 \\
& thumbtack & 24
& & Huggingface & 43 \\
& spothero & 24
& & ArXiv & 43 \\
& drugs & 23
& & Coursera & 42 \\
& finance.yahoo & 23
& & BBC News & 42 \\
& healthline & 23
& & Amazon & 41 \\
\bottomrule
\end{tabular}
}
\end{table}

\subsection{Fine-Grained Annotations}

AgentVista and MMSearch-Plus additionally contain fine-grained subdomain annotations.
We use these annotations only for stratified split construction and dataset analysis.
They are not provided as task inputs to the agent or to the memory system during training or evaluation.

\section{Baseline Details}
\label{app:baseline_details}

This section provides implementation details for the baselines used in our experiments.
For fair comparison, all memory baselines are instantiated under the same update-then-retrieve
protocol as \methodname. For each benchmark, a baseline first processes the training episodes
to construct its memory, and is then evaluated on the held-out test split without further memory
updates. Unless otherwise specified, all baselines use the same execution agent, train/test split,
evaluation protocol, retrieval budget, and context budget as \methodname. In particular, retrieved
memory contents are truncated to the same maximum payload length, and the number of returned
visual references is restricted under the same image budget when the baseline supports multimodal
memory.

\paragraph{NoMemory.}
NoMemory is the base execution agent without any external memory. The agent receives only the
current task input, environment observations, and its own interaction history during execution. This
baseline measures the performance of the underlying execution agent before introducing any
experience-driven memory mechanism.

\subsection{Text-based Baselines}
\label{app:text_based_baselines}

Text-based baselines convert previous episodes into textual memories and retrieve relevant textual
contents at test time. These methods do not directly return visual references to the execution agent.
When the original method is not defined for our multimodal agent setting, we adapt it by extracting
textual information from the episode recorder, including the task instruction, textual observations,
action history, model messages, and final outcome.

\paragraph{Trajectory Retrieval.}
Trajectory Retrieval stores previous interaction trajectories as retrievable examples. During the
update phase, each training episode is converted into a textual trajectory record, including the task
instruction, important observations, executed actions, intermediate responses, and final outcome.
During retrieval, the current task is used as the query to select the most relevant trajectory records.
The retrieved records are formatted as demonstrations or reference cases and injected into the
execution agent. To ensure a fair comparison, retrieved trajectories are truncated to satisfy the same
memory payload budget used by \methodname.

\paragraph{ReasoningBank.}
ReasoningBank~\citep{ouyang2025reasoningbank} stores reusable reasoning memories distilled from
past successful and failed experiences. During the update phase, each episode is converted into
generalizable reasoning knowledge, such as task-solving strategies, failure patterns, corrective lessons,
and reusable decision rules. These memories are indexed by task description and episode metadata.
During retrieval, the current task is used to select relevant reasoning memories, which are then
provided to the execution agent as textual guidance. Compared with Trajectory Retrieval, this
baseline uses more abstract and compressed reasoning memories rather than raw trajectory records.

\paragraph{G-Memory.}
G-Memory~\citep{zhang2025g} organizes experience using a hierarchical graph memory. Following
its original design, we instantiate three types of textual memory structures corresponding to
high-level insights, task/query-level records, and fine-grained interaction traces. During the update
phase, training episodes are decomposed into these memory units and inserted into the graph
structure. During retrieval, the current task is matched against the graph memory, and relevant
high-level insights and condensed interaction records are retrieved. The retrieved graph contents are
linearized into text before being injected into the execution agent.

\subsection{Multimodal Baselines}
\label{app:multimodal_baselines}

Multimodal baselines make use of visual information in addition to textual episode contents. These
methods are adapted to our episode-recorder format, which may contain task instructions, screenshots,
visual observations, action histories, model messages, and final outcomes. When a baseline supports
visual memory, we allow it to return visual references under the same maximum-image budget as
\methodname.

\paragraph{XSkill.}
XSkill~\citep{jiang2026xskill} is a dual-stream continual learning framework that accumulates both
experiences and skills from multimodal agent trajectories. During the update phase, each training
episode is used to extract action-level experience and task-level skill memories, grounded in the visual observations. During retrieval, XSkill selects relevant experiences and skills
conditioned on the current task and visual context. The retrieved textual knowledge and visual
references are then provided to the execution agent under the same retrieval budget as \methodname.

\paragraph{\texorpdfstring{\(\mathrm{M}^2\)}{M2}.}
\(\mathrm{M}^2\)~\citep{yan2026m} is a dual-memory framework for long-horizon web agents that
combines trajectory summarization and insight retrieval. We adapt it to our offline update-then-retrieve
evaluation protocol. During the update phase, training episodes are converted into trajectory-level
summaries and higher-level insights. During retrieval, the current task and initial visual context are
used to retrieve relevant summaries and insights from the constructed memory. The retrieved textual
contents and visual references are then provided to the execution agent at the beginning of the episode.

Under this protocol, we use the long-term memory retrieval component of \(\mathrm{M}^2\), while disabling
test-time dynamic trajectory summarization. This adaptation ensures that all baselines provide memory
only through the same initial retrieval interface.

\subsection{Automatic Memory Design Baselines}
\label{app:automatic_design_baselines}

Automatic memory design baselines search for memory mechanisms rather than relying on a single
hand-written memory schema. These baselines are evaluated under the same sequential search budget
as \methodname, so that performance differences reflect the quality of the searched memory design
rather than additional search cost.

\paragraph{ALMA.}
ALMA~\citep{xiong2026learning} is an automatic memory-design baseline that uses a meta agent to
search over memory designs expressed as executable code. We adapt ALMA to our update-then-retrieve
evaluation protocol. Each candidate memory design is evaluated by first updating memory on the
training episodes and then retrieving memory for held-out tasks. To match the sequential search
setting of \methodname, ALMA is allocated the same number of search steps, and one candidate
memory design is generated and evaluated at each step. The final ALMA memory design is selected
according to validation performance and then evaluated on the test split under the same execution
and retrieval budgets.

\subsection{Summary of Baseline Implementations}
\label{app:baseline_summary}

Table~\ref{tab:baseline_details} summarizes the main differences among the baseline implementations.
All methods are evaluated using the same benchmark splits, execution agents, and evaluation protocol.

\begin{table}[t]
\centering
\caption{Summary of baseline implementations under our unified update-then-retrieve protocol. 
All methods use the same train/test splits, execution agents, evaluation protocol, and retrieval budget.}
\label{tab:baseline_details}
\resizebox{\linewidth}{!}{
\begin{tabular}{lllll}
\toprule
\textbf{Method} & \textbf{Category} & \textbf{Stored memory} & \textbf{Visual references} & \textbf{Search / retrieval protocol} \\
\midrule
NoMemory 
& -- 
& None 
& No 
& No external memory \\

Trajectory Retrieval 
& Text-based 
& Textual trajectory records 
& No 
& Initial retrieval only \\

ReasoningBank 
& Text-based 
& Reasoning strategies and failure lessons 
& No 
& Initial retrieval only \\

G-Memory 
& Text-based 
& Hierarchical textual graph memory 
& No 
& Initial retrieval only \\

XSkill 
& Multimodal 
& Experiences and skills 
& Yes 
& Initial retrieval only \\

\(\mathrm{M}^2\) 
& Multimodal 
& Offline trajectory summaries and insights 
& Yes 
& Initial retrieval only; dynamic summarization disabled \\

ALMA 
& Automatic design 
& Searched executable memory design 
& No 
& 20-step design search \\

\methodname 
& Automatic design 
& Searched multimodal memo program 
& Yes 
& 20-step design search \\

\bottomrule
\end{tabular}
}
\end{table}

\section{Implementation Details}
\label{app:implementation_details}

\subsection{Evaluation Protocol and Metrics}

We follow the offline update-then-retrieve protocol described in Section~\ref{sec:experiments}.
This appendix provides additional implementation-level details, including interaction limits,
invalid-run handling, search hyperparameters, final-evaluation settings, and system configurations.

Each episode is limited to at most 15 interaction steps. 
Task failures are counted as zero. 
Infrastructure-level invalid runs, such as browser crashes, page loading failures, API errors, or tool
execution failures, are excluded from the aggregate results and rerun when possible. 
During final evaluation, retrieved memory is injected only once before the first user turn, and no
additional memory updates are performed on the held-out test split.

For GUI/web navigation benchmarks, we report success rate. 
For visual reasoning/search benchmarks, we report judge-based accuracy. 
AVG. GUI is computed as the simple macro-average over WebVoyager and Mind2Web, while AVG.
VR is computed as the simple macro-average over AgentVista and MMSearch-Plus. 
We do not use dataset-size-weighted averaging unless otherwise specified.

All baselines are adapted to the same memory interface and worker-side injection budget used by
\methodname. Detailed baseline adaptations and retrieval settings are provided in
Appendix~\ref{app:baseline_details}.

\subsection{Memory-Design Search}

We run memory-design search for 20 steps. 
At each search step, a candidate memo program is either generated by mutating an existing node or
selected for re-evaluation according to the search algorithm described in Section~\ref{sec:method}. 
Before full evaluation, each generated memo program is checked by a quick test on 5 randomly
sampled training tasks to filter out invalid implementations and obvious runtime failures.

If the quick test fails, we regenerate or repair the candidate for up to three attempts. 
If all attempts fail, the current search step is marked as failed, and no child memo program is saved or
sent to full evaluation. 
Otherwise, the candidate is evaluated under the update-then-retrieve protocol on the training split of
the corresponding benchmark. 
After the search budget is exhausted, the final memo program is selected using the confidence-aware
selection rule described in Section~\ref{sec:method}.

The same 20-step sequential search budget is used for \methodname and ALMA. 
Baseline-specific adaptations, including the one-candidate-per-step ALMA protocol, are described in
Appendix~\ref{app:baseline_details}.

\subsection{Search Hyperparameters}

Table~\ref{tab:search_hparams} lists the hyperparameters used during memory-design search.

\begin{table}[t]
\centering
\caption{Hyperparameters used during memory-design search.}
\label{tab:search_hparams}
\begin{tabular}{lc}
\toprule
\textbf{Hyperparameter} & \textbf{Value} \\
\midrule
Search steps & 20 \\
Quick-test tasks per candidate & 5 \\
Maximum regeneration / repair attempts & 3 \\
Full evaluation tasks per valid candidate & Training retrieve set \\
Maximum interaction steps & 15 \\
Evaluation confidence coefficient $c_{\mathrm{eval}}$ & 0.2 \\
Generation confidence coefficient $c_{\mathrm{gen}}$ & 0.2 \\
Generation prior strength $\rho$ & 0.5 \\
Generation prior pseudo-count $\beta$ & 1.0 \\
Minimum width $B$ & 2 \\
Max evaluation screenshots & 5 \\
Max browser concurrency & 8 \\
Max update batch concurrency & 8 \\
Max retrieve batch concurrency & 8 \\
Max meta observation summary length & 50,000 \\
Max meta images per episode & 4 \\
Max retrieved memory characters & 50,000 \\
Max retrieved memory images & 2 \\
Max meta retrieved-memory summary length & 20,000 \\
Max meta retrieved-memory images per episode & 2 \\
Success trajectories for meta analysis & 2 \\
Failure trajectories for meta analysis & 2 \\
\bottomrule
\end{tabular}
\end{table}

\subsection{Final Evaluation Hyperparameters}

Table~\ref{tab:eval_hparams} lists the hyperparameters used during final held-out evaluation.
All methods use the same execution budget, context budget, retrieval budget, and evaluation protocol.
During final evaluation, retrieved memories are injected before the first user turn using the same
worker-side text and image budget for all methods.

\begin{table}[t]
\centering
\caption{Hyperparameters used during final held-out evaluation.}
\label{tab:eval_hparams}
\begin{tabular}{lc}
\toprule
\textbf{Hyperparameter} & \textbf{Value} \\
\midrule
Judge model & GPT-5.4-mini \\
Maximum interaction steps & 15 \\
Max evaluation screenshots & 5 \\
Max browser concurrency & 16 \\
Max update batch concurrency & 16 \\
Max retrieve batch concurrency & 16 \\
Max images in agent context & 8 \\
Retrieved memory character budget & 50,000 \\
Retrieved memory image budget & 2 \\
Independent evaluation runs & 3 \\
\bottomrule
\end{tabular}
\end{table}

\subsection{Model and System Details}

We instantiate the execution agent with Qwen3-VL-32B~\citep{bai2025qwen3},
GPT-5.4-nano~\citep{singh2025openai}, and Qwen3.5-Plus~\citep{bai2025qwen3}. 
Unless otherwise specified, the meta agent uses GPT-5~\citep{singh2025openai}, and the LLM judge
uses GPT-5.4-mini~\citep{singh2025openai}. 
Closed-source models are accessed through their official API endpoints, while Qwen3-VL-32B is
served locally.

All experiments are conducted using PyTorch~\citep{paszke2019pytorch} on 8 NVIDIA Tesla A100
GPUs, with open-source models launched via vLLM~\citep{kwon2023efficient} to enable efficient
inference. 
During memory-design search, we set the maximum browser concurrency to 8 and the maximum
concurrent update and retrieval batches to 8. 
During final evaluation, we increase these values to 16 to improve throughput. 
These concurrency settings affect only wall-clock efficiency and are kept fixed across methods within
the same evaluation stage.

\begin{table*}[t]
\centering
\caption{
Main results on GUI-agent and visual-reasoning benchmarks across different execution models. 
Numbers in parentheses indicate absolute performance changes over the corresponding NoMemory baseline, and bold denotes the best result within each execution-model block.
}
\label{tab:main_result_qwen35}
\begin{adjustbox}{width=\columnwidth}
\begin{tabular}{lcccccc}
\toprule
\multirow{2.5}{*}{\textbf{Method}}
& \multicolumn{2}{c}{\textbf{GUI Agent}}
& \multirow{2.5}{*}{\textbf{AVG. GUI}}
& \multicolumn{2}{c}{\textbf{Visual Reasoning}}
& \multirow{2.5}{*}{\textbf{AVG. VR}} \\

\cmidrule(lr){2-3} \cmidrule(lr){5-6}

& WebVoyager & Mind2Web &  & AgentVista & MMSearch-Plus \\ 
\midrule

\multicolumn{7}{c}{\textit{Qwen3.5-plus}} \\
\midrule

NoMemory & \makebox[2.8em][r]{58.23}\makebox[2.2em][l]{\hspace{0.2em}} & \makebox[2.8em][r]{30.11}\makebox[2.2em][l]{\hspace{0.2em}} & \makebox[2.8em][r]{44.17}\makebox[2.2em][l]{\hspace{0.2em}} & \makebox[2.8em][r]{17.87}\makebox[2.2em][l]{\hspace{0.2em}} & \makebox[2.8em][r]{24.14}\makebox[2.2em][l]{\hspace{0.2em}} & \makebox[2.8em][r]{21.01}\makebox[2.2em][l]{\hspace{0.2em}} \\
\rowcolor{gray!8} \multicolumn{7}{c}{\textbf{Text-based}} \\
ReasoningBank & \makebox[2.8em][r]{59.90}\makebox[2.2em][l]{\hspace{0.2em}\scriptsize\textcolor{poscolor}{(+1.67)}} & \makebox[2.8em][r]{26.25}\makebox[2.2em][l]{\hspace{0.2em}\scriptsize\textcolor{negcolor}{(-3.86)}} & \makebox[2.8em][r]{43.08}\makebox[2.2em][l]{\hspace{0.2em}\scriptsize\textcolor{negcolor}{(-1.09)}} & \makebox[2.8em][r]{11.70}\makebox[2.2em][l]{\hspace{0.2em}\scriptsize\textcolor{negcolor}{(-6.17)}} & \makebox[2.8em][r]{23.08}\makebox[2.2em][l]{\hspace{0.2em}\scriptsize\textcolor{negcolor}{(-1.06)}} & \makebox[2.8em][r]{17.39}\makebox[2.2em][l]{\hspace{0.2em}\scriptsize\textcolor{negcolor}{(-3.62)}} \\
TrajectoryRetrieval & \makebox[2.8em][r]{59.08}\makebox[2.2em][l]{\hspace{0.2em}\scriptsize\textcolor{poscolor}{(+0.85)}} & \makebox[2.8em][r]{31.83}\makebox[2.2em][l]{\hspace{0.2em}\scriptsize\textcolor{poscolor}{(+1.72)}} & \makebox[2.8em][r]{45.45}\makebox[2.2em][l]{\hspace{0.2em}\scriptsize\textcolor{poscolor}{(+1.28)}} & \makebox[2.8em][r]{20.03}\makebox[2.2em][l]{\hspace{0.2em}\scriptsize\textcolor{poscolor}{(+2.16)}} & \makebox[2.8em][r]{24.99}\makebox[2.2em][l]{\hspace{0.2em}\scriptsize\textcolor{poscolor}{(+0.85)}} & \makebox[2.8em][r]{22.51}\makebox[2.2em][l]{\hspace{0.2em}\scriptsize\textcolor{poscolor}{(+1.50)}} \\
G-Memory & \makebox[2.8em][r]{59.68}\makebox[2.2em][l]{\hspace{0.2em}\scriptsize\textcolor{poscolor}{(+1.45)}} & \makebox[2.8em][r]{32.80}\makebox[2.2em][l]{\hspace{0.2em}\scriptsize\textcolor{poscolor}{(+2.69)}} & \makebox[2.8em][r]{46.24}\makebox[2.2em][l]{\hspace{0.2em}\scriptsize\textcolor{poscolor}{(+2.07)}} & \makebox[2.8em][r]{17.33}\makebox[2.2em][l]{\hspace{0.2em}\scriptsize\textcolor{negcolor}{(-0.54)}} & \makebox[2.8em][r]{23.00}\makebox[2.2em][l]{\hspace{0.2em}\scriptsize\textcolor{negcolor}{(-1.14)}} & \makebox[2.8em][r]{20.16}\makebox[2.2em][l]{\hspace{0.2em}\scriptsize\textcolor{negcolor}{(-0.84)}} \\
\rowcolor{gray!8} \multicolumn{7}{c}{\textbf{Multimodal-based}} \\
XSkill & \makebox[2.8em][r]{52.61}\makebox[2.2em][l]{\hspace{0.2em}\scriptsize\textcolor{negcolor}{(-5.62)}} & \makebox[2.8em][r]{25.08}\makebox[2.2em][l]{\hspace{0.2em}\scriptsize\textcolor{negcolor}{(-5.03)}} & \makebox[2.8em][r]{38.84}\makebox[2.2em][l]{\hspace{0.2em}\scriptsize\textcolor{negcolor}{(-5.33)}} & \makebox[2.8em][r]{21.75}\makebox[2.2em][l]{\hspace{0.2em}\scriptsize\textcolor{poscolor}{(+3.88)}} & \makebox[2.8em][r]{25.21}\makebox[2.2em][l]{\hspace{0.2em}\scriptsize\textcolor{poscolor}{(+1.07)}} & \makebox[2.8em][r]{23.48}\makebox[2.2em][l]{\hspace{0.2em}\scriptsize\textcolor{poscolor}{(+2.47)}} \\
M$^2$ & \makebox[2.8em][r]{60.73}\makebox[2.2em][l]{\hspace{0.2em}\scriptsize\textcolor{poscolor}{(+2.50)}} & \makebox[2.8em][r]{39.25}\makebox[2.2em][l]{\hspace{0.2em}\scriptsize\textcolor{poscolor}{(+9.14)}} & \makebox[2.8em][r]{49.99}\makebox[2.2em][l]{\hspace{0.2em}\scriptsize\textcolor{poscolor}{(+5.82)}} & \makebox[2.8em][r]{10.52}\makebox[2.2em][l]{\hspace{0.2em}\scriptsize\textcolor{negcolor}{(-7.35)}} & \makebox[2.8em][r]{18.38}\makebox[2.2em][l]{\hspace{0.2em}\scriptsize\textcolor{negcolor}{(-5.76)}} & \makebox[2.8em][r]{14.45}\makebox[2.2em][l]{\hspace{0.2em}\scriptsize\textcolor{negcolor}{(-6.56)}} \\
\rowcolor{gray!8} \multicolumn{7}{c}{\textbf{Automatic Design}} \\
ALMA & \makebox[2.8em][r]{60.06}\makebox[2.2em][l]{\hspace{0.2em}\scriptsize\textcolor{poscolor}{(+1.83)}} & \makebox[2.8em][r]{36.30}\makebox[2.2em][l]{\hspace{0.2em}\scriptsize\textcolor{poscolor}{(+6.19)}} & \makebox[2.8em][r]{48.18}\makebox[2.2em][l]{\hspace{0.2em}\scriptsize\textcolor{poscolor}{(+4.01)}} & \makebox[2.8em][r]{23.44}\makebox[2.2em][l]{\hspace{0.2em}\scriptsize\textcolor{poscolor}{(+5.57)}} & \makebox[2.8em][r]{25.73}\makebox[2.2em][l]{\hspace{0.2em}\scriptsize\textcolor{poscolor}{(+1.59)}} & \makebox[2.8em][r]{24.59}\makebox[2.2em][l]{\hspace{0.2em}\scriptsize\textcolor{poscolor}{(+3.58)}} \\
\textbf{\methodname} & \makebox[2.8em][r]{\textbf{65.04}}\makebox[2.2em][l]{\hspace{0.2em}\scriptsize\textcolor{poscolor}{(+6.81)}} & \makebox[2.8em][r]{\textbf{45.68}}\makebox[2.2em][l]{\hspace{0.2em}\scriptsize\textcolor{poscolor}{(+15.57)}} & \makebox[2.8em][r]{\textbf{55.36}}\makebox[2.2em][l]{\hspace{0.2em}\scriptsize\textcolor{poscolor}{(+11.19)}} & \makebox[2.8em][r]{\textbf{28.81}}\makebox[2.2em][l]{\hspace{0.2em}\scriptsize\textcolor{poscolor}{(+10.94)}} & \makebox[2.8em][r]{\textbf{32.61}}\makebox[2.2em][l]{\hspace{0.2em}\scriptsize\textcolor{poscolor}{(+8.47)}} & \makebox[2.8em][r]{\textbf{30.71}}\makebox[2.2em][l]{\hspace{0.2em}\scriptsize\textcolor{poscolor}{(+9.70)}} \\

\bottomrule
\end{tabular}
\end{adjustbox}
\end{table*}

\section{Additional Experimental Results}
\label{app:additional_experiments}
\subsection{Results on Qwen3.5-Plus}
\label{app:qwen35_results}

Table~\ref{tab:main_result_qwen35} also reports the results on Qwen3.5-Plus. 
The observations are consistent with the main results. 
\methodname achieves the best performance on all four benchmarks and obtains the highest average scores for both GUI navigation and visual reasoning. 
Compared with the NoMemory baseline, \methodname improves AVG. GUI from 44.17 to 55.36, yielding an absolute gain of 11.19 points, and improves AVG. VR from 21.01 to 30.71, yielding a gain of 9.70 points.

The improvement is especially notable on Mind2Web and AgentVista. 
On Mind2Web, \methodname improves from 30.11 to 45.68, substantially outperforming both text-based and multimodal-based memory baselines. 
On AgentVista, \methodname improves from 17.87 to 28.81, showing that the learned memory mechanism is also effective for visual reasoning. 
These results further confirm that \methodname generalizes beyond a specific execution model and provides consistent benefits across heterogeneous multimodal tasks.

\subsection{Interaction Budget}
\label{app:interaction_budget}

On the full scored WebVoyager set, we run each method once per task with a maximum of 60 interaction steps and evaluate the trajectory prefixes at 15, 30, 45, and 60 steps, using Qwen3-VL-32B and the main experiment's aggregation protocol. Tasks not completed by a cutoff count as failures. The memory remains fixed, built from the original 15-step update episodes. Table~\ref{tab:interaction_budget} shows that longer interaction budgets improve both methods, while the matched gain of \methodname remains 15.17--17.47 percentage points.

\begin{table}[htbp]
\centering
\caption{WebVoyager success rates (\%) at different interaction cutoffs, with fixed memory and Qwen3-VL-32B as the execution model.}
\label{tab:interaction_budget}
\small
\setlength{\tabcolsep}{4pt}
\begin{tabular}{lrrrr}
\toprule
Method & Success@15 & Success@30 & Success@45 & Success@60 \\
\midrule
NoMemory & 36.32\% & 45.28\% & 49.75\% & 50.47\% \\
\methodname & 51.84\% & 62.75\% & 65.64\% & 65.64\% \\
\bottomrule
\end{tabular}
\end{table}

\subsection{Memory Ablations}
\label{app:memory_ablations}

\paragraph{Memory update checkpoints.}
We evaluate a fixed memo program at 0/25/50/75/100\% of the same update stream on WebVoyager, using Qwen3-VL-32B, one rollout per task, and the main experiment's Success@15 protocol. Update episodes and scored query tasks remain disjoint. In Table~\ref{tab:memory_update_checkpoints}, state denotes the serialized runtime memory size; the 52 bytes at 0\% contain only empty-state metadata. Mean characters measure the retrieved context added beyond the fixed 0\% initialization, and coverage is the fraction of queries returning nonempty update-derived memory. Success rises from 36.32\% to 51.84\% as update experience accumulates.

\begin{table}[htbp]
\centering
\caption{WebVoyager memory update checkpoints with a fixed memo program. State is measured in bytes; characters and coverage refer to update-derived retrieved context.}
\label{tab:memory_update_checkpoints}
\small
\setlength{\tabcolsep}{4pt}
\begin{tabular}{lrrrr}
\toprule
Update progress & Success@15 & State (bytes) & Mean chars & Coverage \\
\midrule
0\% & 36.32\% & 52 & 0.00 & 0.0\% \\
25\% & 44.65\% & 2,337 & 211.36 & 58.0\% \\
50\% & 47.70\% & 3,041 & 273.91 & 74.5\% \\
75\% & 49.81\% & 3,387 & 340.82 & 91.5\% \\
100\% & 51.84\% & 3,831 & 359.17 & 94.3\% \\
\bottomrule
\end{tabular}
\end{table}

\paragraph{Shuffled retrieval.}
Under the same WebVoyager protocol, we apply a one-to-one permutation of complete retrieval payloads across tasks and websites (seed 42), preserving the payload multiset with no self-mapping or same-website donor. Success@15 falls from 51.84\% with correctly matched retrieval to 40.57\% with shuffled retrieval, compared with 36.32\% for NoMemory.

\paragraph{Trajectory-conditioned reflection.}
On AgentVista, removing reflection while retaining mutation, quick examination, code repair, full evaluation, and UCB/LCB search reduces the test score from 14.36\% to 12.76\%. Both settings use a 20-round search budget and the same test split and evaluation protocol as the main experiment.

\subsection{Human--Judge Agreement}
\label{app:human_judge}

For each combination of WebVoyager or MMSearch-Plus and \methodname or NoMemory, we randomly sample 50 trajectories stratified by the original LLM verdict: 20 successes and 30 failures. Three annotators independently label all 200 trajectories using the benchmark-specific binary rubric, with method identity, memory content, and the original verdict hidden. Majority vote defines the human label. Human scores, agreement, and Cohen's $\kappa$ are weighted by the verdict distribution in each full evaluation; $\kappa$ compares the majority-human label with the LLM verdict. Table~\ref{tab:human_judge_agreement} reports human-rescored gains of 15.53 and 5.51 percentage points on WebVoyager and MMSearch-Plus, respectively, close to the corresponding LLM-judge gains of 15.52 and 5.69 points.

\begin{table}[htbp]
\centering
\caption{Human audit of the two judge families. Scores and agreement are percentages, weighted by each full evaluation's judge-verdict distribution.}
\label{tab:human_judge_agreement}
\small
\setlength{\tabcolsep}{4pt}
\begin{tabular}{llrrrr}
\toprule
Benchmark & Method & LLM & Human & Agreement & $\kappa$ \\
\midrule
WebVoyager & \methodname & 51.84\% & 52.46\% & 94.20\% & 0.884 \\
WebVoyager & NoMemory & 36.32\% & 36.93\% & 92.12\% & 0.830 \\
MMSearch-Plus & \methodname & 11.46\% & 13.84\% & 96.48\% & 0.841 \\
MMSearch-Plus & NoMemory & 5.77\% & 8.33\% & 96.28\% & 0.717 \\
\bottomrule
\end{tabular}
\end{table}

\section{Additional Cost Analysis}
\label{app:additional_cost}
\subsection{Search-Time Cost}
\label{app:search_cost}

In the main text, we analyze the inference-time cost of the final learned memo program.
Here we report the offline search-time overhead incurred during memory-design optimization.
All numbers exclude the initial root evaluation, which is shared by the search procedure and is not counted as part of the 20-step search budget.

Table~\ref{tab:search_process_cost} summarizes the search process of \methodname under Qwen3-VL-32B.
Across the four benchmarks, \methodname dynamically allocates the fixed 20-step search budget between generating new memo programs and re-evaluating existing nodes.
Only 43 out of 80 search steps are generation actions, while the remaining 37 steps re-evaluate existing programs to reduce score uncertainty.
This behavior improves search efficiency because re-evaluation steps do not invoke the meta agent for reflection, mutation, or repair.
Most generated candidates pass quick examination, with only one failed generation on WebVoyager, indicating that the quick examination and repair stage effectively prevents invalid programs from consuming full-evaluation budget.

Table~\ref{tab:search_token_cost} compares the search-time token and wall-clock overhead of \methodname and ALMA.
Meta tokens include reflection, mutation, and repair calls for \methodname, and candidate memory-design generation calls for ALMA under our adapted sequential search protocol.
Evaluation tokens include execution-agent and judge-model tokens consumed by full evaluations during search.
Across all four benchmarks, \methodname uses fewer meta tokens, fewer evaluation tokens, fewer total search-time tokens, and less wall-clock time than ALMA.
Aggregated over the four benchmarks, \methodname reduces meta-search tokens by 45.7\%, evaluation tokens by 22.3\%, total search-time tokens by 22.6\%, and wall-clock time by 26.6\%.
These results show that the performance gains of \methodname are not obtained by using a larger offline design budget.

\begin{table}[t]
\centering
\caption{Search-time process statistics under Qwen3-VL-32B.
We report the number of generation actions, re-evaluation actions, valid generated programs, candidates rejected by quick examination, and full evaluations consumed during the 20-step search.
All numbers exclude the initial root evaluation.}
\label{tab:search_process_cost}
\resizebox{\columnwidth}{!}{
\begin{tabular}{lcccccc}
\toprule
Benchmark & Search Steps & Gen. Actions & Re-eval. Actions & Valid Gen. & Failed Gen. & Full Eval. \\
\midrule
WebVoyager & 20 & 15 & 5  & 14 & 1 & 19 \\
Mind2Web & 20 & 7  & 13 & 7  & 0 & 20 \\
AgentVista & 20 & 10 & 10 & 10 & 0 & 20 \\
MMSearch-Plus & 20 & 11 & 9  & 11 & 0 & 20 \\
\bottomrule
\end{tabular}
}
\end{table}

\begin{table}[t]
\centering
\caption{
Search-time token and wall-clock overhead under Qwen3-VL-32B.
Both methods use the same 20-step sequential search budget, and all numbers exclude the initial root evaluation.
Meta tokens include reflection, mutation, and repair calls for \methodname, and candidate memory-design generation calls for ALMA.
Evaluation tokens include execution-agent and judge-model tokens consumed by search-time full evaluations.
Numbers in parentheses denote relative reductions over ALMA.
}
\label{tab:search_token_cost}
\resizebox{\columnwidth}{!}{
\begin{tabular}{llcccc}
\toprule
Benchmark & Method & Meta Tok. (M) & Eval. Tok. (M) & Total Tok. (M) & Wall-clock (h) \\
\midrule
\multirow{2}{*}{WebVoyager} & ALMA & 1.706 & 172.127 & 173.833 & 28.05 \\
& \methodname & 1.254 {\scriptsize(-26.5\%)}
& 146.242 {\scriptsize(-15.0\%)}
& 147.496 {\scriptsize(-15.2\%)}
& 23.98 {\scriptsize(-14.5\%)} \\
\midrule
\multirow{2}{*}{Mind2Web} & ALMA & 1.489 & 308.702 & 310.191 & 26.85 \\
& \methodname & 0.537 {\scriptsize(-63.9\%)}
& 243.541 {\scriptsize(-21.1\%)}
& 244.078 {\scriptsize(-21.3\%)}
& 23.25 {\scriptsize(-13.4\%)} \\
\midrule
\multirow{2}{*}{AgentVista} & ALMA & 1.194 & 20.390 & 21.584 & 22.03 \\
& \methodname & 0.574 {\scriptsize(-51.9\%)}
& 14.047 {\scriptsize(-31.1\%)}
& 14.621 {\scriptsize(-32.3\%)}
& 16.04 {\scriptsize(-27.2\%)} \\
\midrule
\multirow{2}{*}{MMSearch-Plus} & ALMA & 1.448 & 62.062 & 63.510 & 63.59 \\
& \methodname & 0.804 {\scriptsize(-44.4\%)}
& 33.577 {\scriptsize(-45.9\%)}
& 34.381 {\scriptsize(-45.9\%)}
& 39.90 {\scriptsize(-37.3\%)} \\
\bottomrule
\end{tabular}
}
\end{table}

\clearpage
\suppressfloats[t]
\subsection{Memory-Operation Latency}
\label{app:memory_latency}

Table~\ref{tab:memory_operation_latency} reports deployment-time memory-operation latency under the same replay protocol across methods. Each cell contains amortized update cost and mean retrieval latency, including internal LLM or embedding calls; deferred build and finalization costs are amortized over updates. These measurements cover memory operations rather than end-to-end agent execution. Retrieval with \methodname ranges from 0.06\,ms on WebVoyager to 2.03\,s on Mind2Web.

\begin{table}[htbp]
\centering
\caption{Deployment-time memory-operation latency. Cells report amortized update cost / mean retrieval latency, including internal model calls and deferred finalization costs. Units are shown explicitly.}
\label{tab:memory_operation_latency}
\footnotesize
\setlength{\tabcolsep}{4pt}
\begin{tabular}{lllll}
\toprule
Method & WebVoyager & Mind2Web & AgentVista & MMSearch-Plus \\
\midrule
\methodname & 0.14 ms / 0.06 ms & 0.23 ms / 2.03 s & 0.38 ms / 0.37 ms & 1.67 s / 10.9 ms \\
NoMemory & $\leq$0.01 ms / $\leq$0.01 ms & $\leq$0.01 ms / $\leq$0.01 ms & $\leq$0.01 ms / $\leq$0.01 ms & $\leq$0.01 ms / $\leq$0.01 ms \\
Trajectory Retrieval & 0.02 ms / 1.03 s & $\leq$0.01 ms / 0.92 s & $\leq$0.01 ms / 0.92 s & $\leq$0.01 ms / 1.12 s \\
ReasoningBank & 3.00 min / 0.94 s & 8.07 s / 1.01 s & 7.45 s / 0.98 s & 38.0 s / 1.00 s \\
M$^2$ & 1.50 min / 0.24 s & 3.94 s / 0.24 s & 2.42 s / 0.24 s & 1.08 s / 0.23 s \\
G-Memory & 1.33 min / 13.8 s & 2.06 min / 14.3 s & 25.2 s / 10.3 s & 1.82 min / 57.4 s \\
XSkill & 13.4 s / 25.9 s & 2.20 min / 1.86 min & 15.0 s / 35.2 s & 53.7 s / 4.45 min \\
ALMA & 0.17 ms / 3.65 s & 4.26 s / 15.3 s & 9.26 s / 0.73 s & 0.46 ms / 42.98 s \\
\bottomrule
\end{tabular}
\end{table}

\section{Detailed Memory Evolution}
\label{app:memory_evolution}

\begin{figure*}[t]
    \centering
    \includegraphics[width=\textwidth]{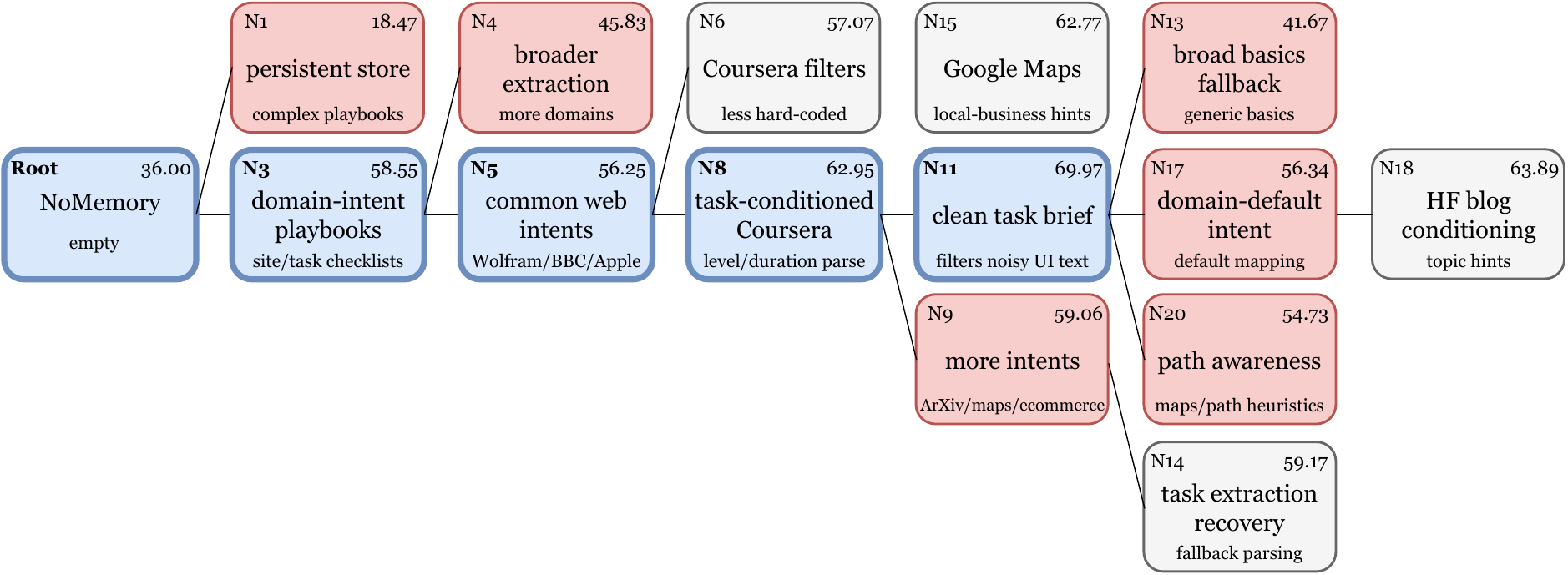}
\caption{
Visualization of the memory-design search tree for the \textsc{WebVoyager} search run with \textsc{Qwen3-VL-32B}.
Each node denotes a candidate memory design, where the score reports its validation performance.
The root node is the NoMemory baseline, and each child node summarizes the main modification made to its parent design.
The final selected memory program is N11, chosen by the LCB rule after search.
The search gradually evolves from simple domain-intent memories to more task-conditioned and recovery-aware designs, revealing how stronger memory mechanisms emerge through iterative exploration.
}
    \label{fig:memory_tree_appendix}
\end{figure*}

This appendix provides a node-level analysis of the memory evolution process discussed in Section~\ref{sec:experiments}. 
While Figure~\ref{fig:memory_evolution} shows the overall search trajectory, Figure~\ref{fig:memory_tree_appendix} visualizes how individual memory candidates are generated, refined, and evaluated during the search process.
Specifically, the figure corresponds to the \textsc{WebVoyager} search run under \textsc{Qwen3-VL-32B} as the execution agent.
Each node corresponds to a candidate memory design, annotated with its validation score and a short description of the main design change.
The scores in this figure are used for search-time model selection and are not directly comparable to the held-out test results in the main table.
The final selected memory program is node N11, which is chosen by the lower-confidence-bound (LCB) rule described in Section~3.4.

\paragraph{Evolution along the high-performing path.}
Starting from the NoMemory root, \methodname quickly discovers that adding structured task experience is beneficial. 
An early candidate introduces domain-intent playbooks, which substantially improves over the root by organizing experience around website domains and task intents. 
Later candidates further refine this idea by adding common web intents, task-conditioned filters, and cleaner task briefs. 
The strongest node in the tree emphasizes filtering noisy UI text and preserving task-relevant information, suggesting that effective memory is not simply a larger context store, but a mechanism for selecting and presenting useful experience to the agent.

\paragraph{Why broader memory is not always better.}
Several generated candidates underperform despite adding more memory content or broader fallback rules. 
For example, candidates that rely on persistent storage of complex playbooks or generic fallback rules obtain much lower validation scores than more task-conditioned candidates. 
This indicates that excessive or weakly targeted memory can introduce irrelevant context, distract the execution agent, and reduce the effectiveness of retrieval. 
The search process therefore needs to optimize not only what to remember, but also how memory is abstracted, indexed, filtered, and injected.

\paragraph{Search reveals reusable but task-sensitive design principles.}
The evolution tree also shows that some alternative branches achieve competitive scores by specializing memory toward particular domains or interaction patterns, such as local-business hints, topic-conditioned retrieval, or task extraction recovery. 
These branches suggest that useful memory designs often share reusable principles, including intent abstraction, domain-aware organization, and observation filtering. 
However, their uneven performance also confirms that memory mechanisms must be adapted to the target task distribution rather than manually fixed in advance.

\section{Learned Memo Programs Across Benchmarks}
\label{app:learned_memo_programs}

\begin{figure*}[t]
    \centering
    \includegraphics[width=\textwidth]{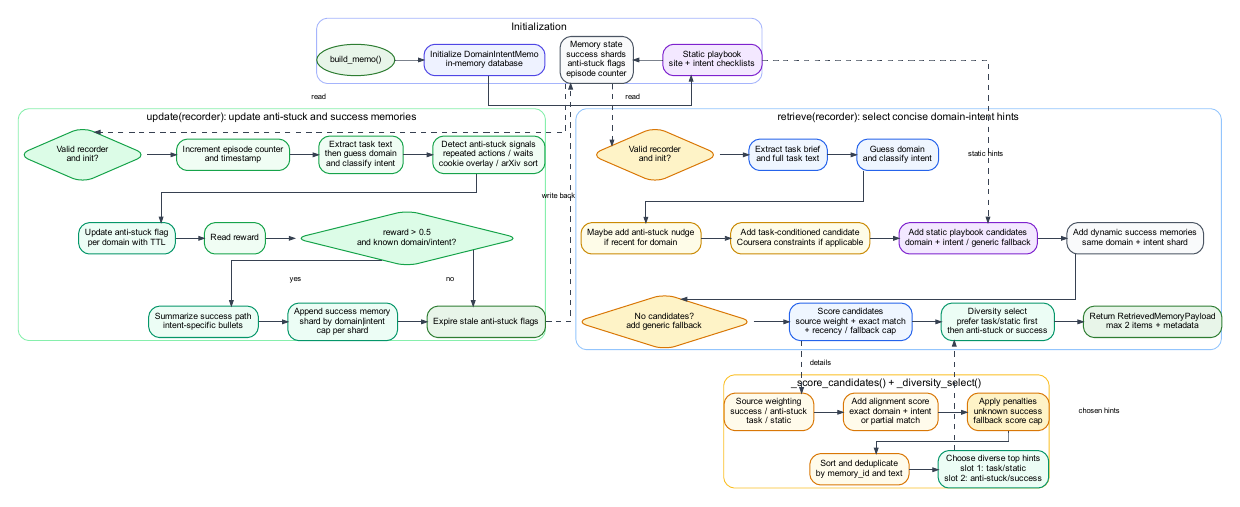}
    \caption{
    Learned memo program for WebVoyager under Qwen3-VL-32B.
    The selected program uses a domain-intent memory design.
    During initialization, it builds static site-intent playbooks and an in-memory state containing success shards, anti-stuck flags, and an episode counter.
    During \texttt{retrieve}, it extracts a clean task brief, predicts the domain and intent, adds task-conditioned and static playbook candidates, optionally includes anti-stuck nudges or dynamic success memories, and returns at most two concise hints.
    During \texttt{update}, it detects repeated actions, waiting, cookie overlays, or other stuck patterns, updates anti-stuck flags, and stores successful trajectories into domain-intent shards.
    }
    \label{fig:memo_webvoyager}
\end{figure*}

\begin{figure*}[t]
    \centering
    \includegraphics[width=\textwidth]{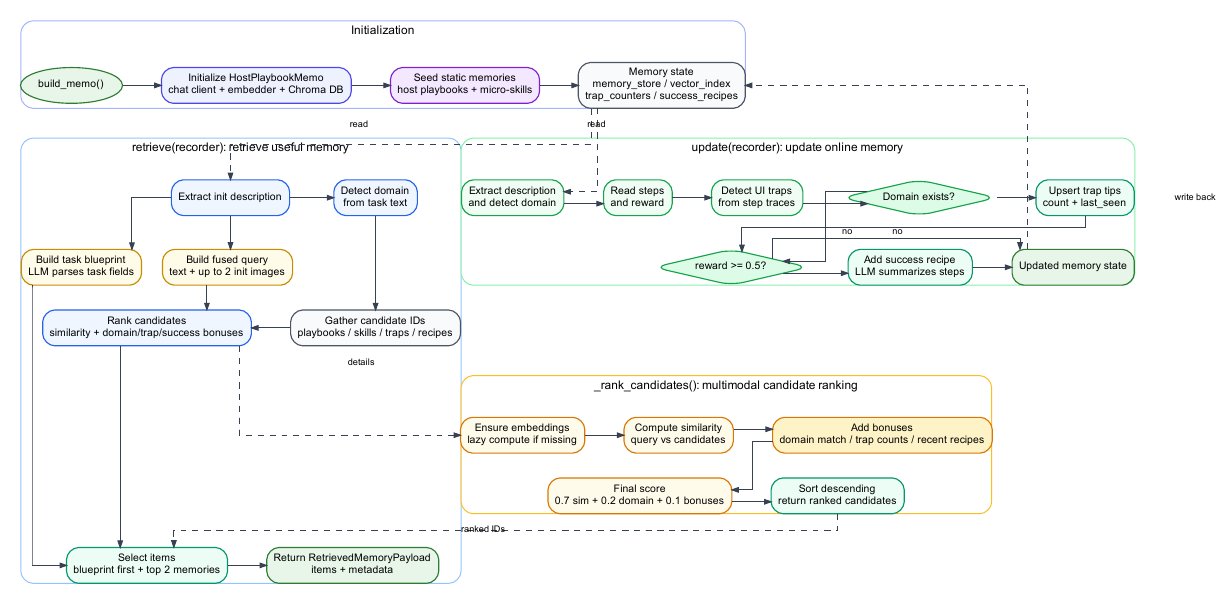}
    \caption{
    Learned memo program for Mind2Web under Qwen3-VL-32B.
    The selected program uses a host-aware playbook memory.
    It initializes static host playbooks and micro-skills, while maintaining dynamic trap counters and success recipes.
    During \texttt{retrieve}, it detects the target domain, constructs a task blueprint from the initial description, builds a fused text-image query using the task and up to two initial images, ranks candidate memories by multimodal similarity, domain match, trap signals, and success bonuses, and returns the task blueprint together with the top ranked memories.
    During \texttt{update}, it detects UI traps from failed trajectories and summarizes successful episodes into reusable recipes.
    }
    \label{fig:memo_mind2web}
\end{figure*}

\begin{figure*}[t]
    \centering
    \includegraphics[width=\textwidth]{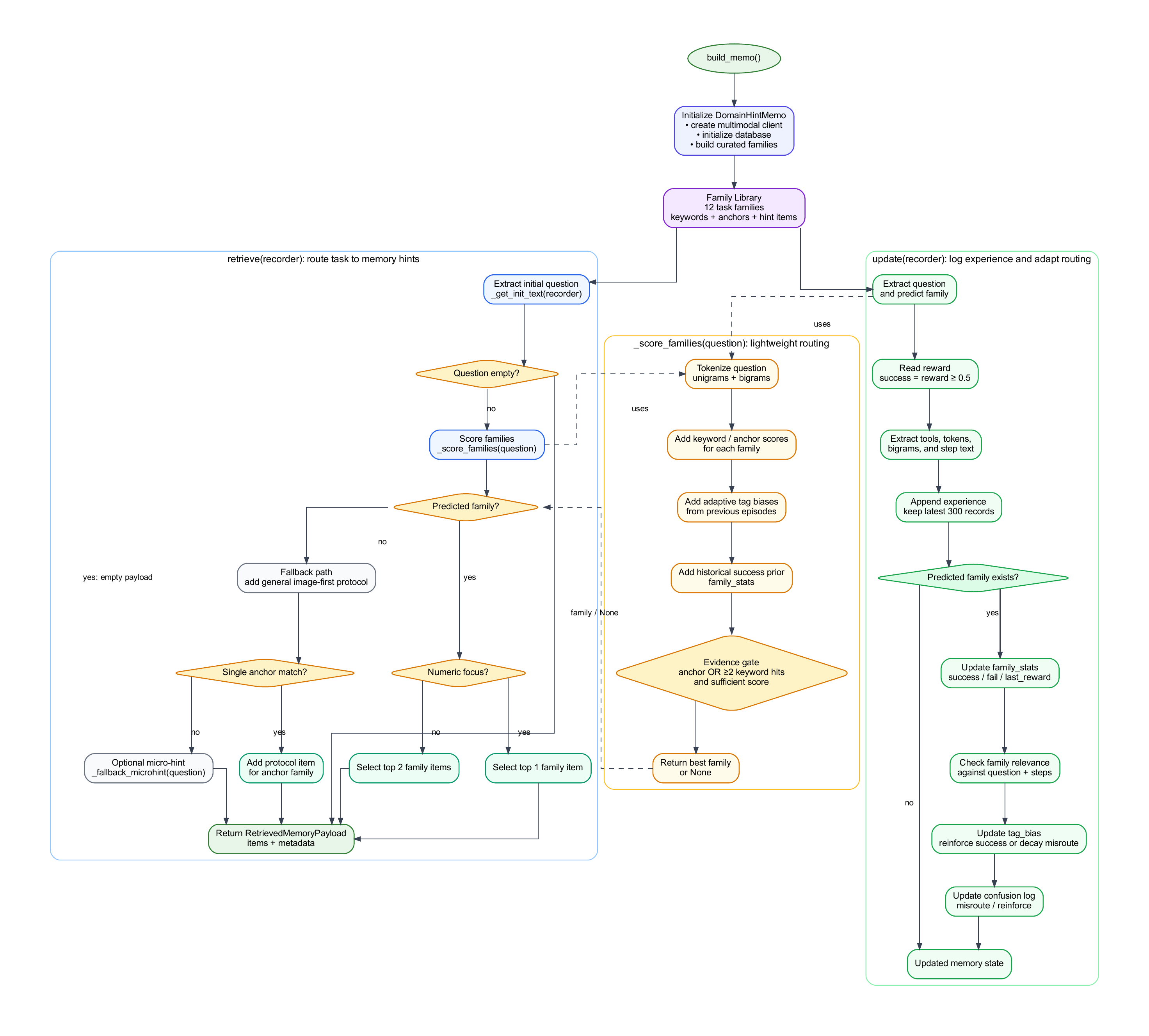}
    \caption{
    Learned memo program for AgentVista under Qwen3-VL-32B.
    The selected program uses a task-family procedural memory.
    It initializes a family library containing curated visual-reasoning task families, keywords, anchors, and concise hint items.
    During \texttt{retrieve}, it extracts the initial question, scores task families using tokens, ngrams, anchors, adaptive tag biases, and historical success priors, and returns one or two procedure-level hints for the predicted family.
    If no confident family is found, it falls back to a general image-first protocol.
    During \texttt{update}, it logs the episode outcome, updates family-level success/failure statistics, reinforces useful routing tags, and decays misleading associations.
    }
    \label{fig:memo_agentvista}
\end{figure*}

\begin{figure*}[t]
    \centering
    \includegraphics[width=\textwidth]{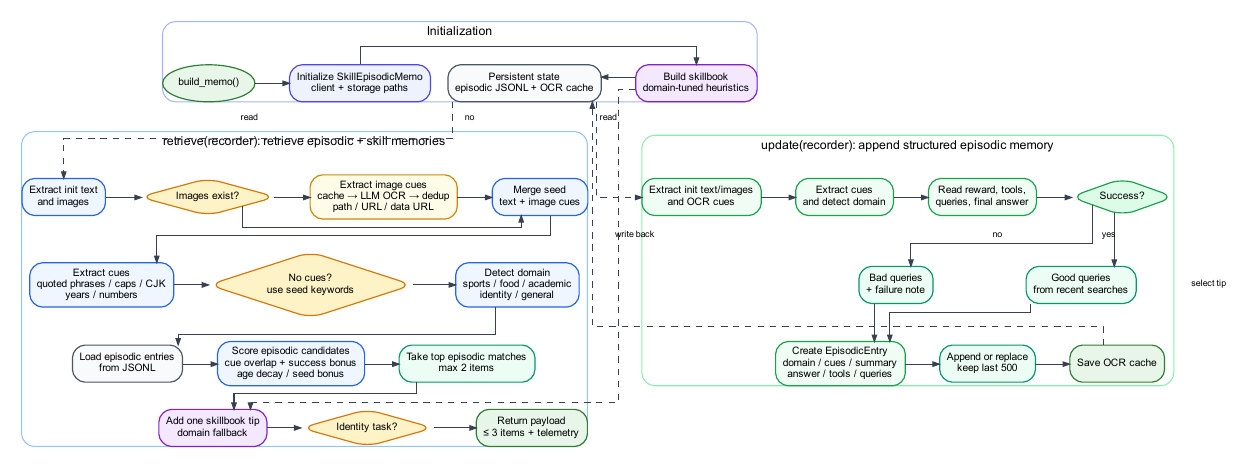}
    \caption{
    Learned memo program for MMSearch-Plus under Qwen3-VL-32B.
    The selected program combines a domain-tuned skillbook with cue-indexed episodic memory.
    During \texttt{retrieve}, it extracts text and image cues from the initial input, optionally uses OCR with caching, detects a coarse domain, scores episodic candidates by cue overlap, success bonus, age decay, and seed matching, and returns up to two episodic matches together with one skillbook tip.
    During \texttt{update}, it extracts cues, tools, search queries, final answers, and failure notes, then appends a structured episodic entry to persistent JSONL storage.
    }
    \label{fig:memo_mmsearch_plus}
\end{figure*}

To better illustrate what \methodname discovers, we visualize the final memo programs selected after search on each benchmark under Qwen3-VL-32B.
While Appendix~\ref{app:memory_evolution} analyzes how memo programs evolve during search, this section focuses on the internal structure of the final selected programs.
Figures~\ref{fig:memo_webvoyager}--\ref{fig:memo_mmsearch_plus} show implementation-level schematics of the learned memo programs for WebVoyager, Mind2Web, AgentVista, and MMSearch-Plus, respectively.
All four programs follow the same external \texttt{update}/\texttt{retrieve} interface, but their internal memory states, indexing rules, retrieval strategies, and returned payloads differ substantially across benchmarks.

\paragraph{WebVoyager.}
Figure~\ref{fig:memo_webvoyager} shows the final memo program selected for WebVoyager.
The learned design uses a domain-intent memory structure that combines static site-intent playbooks, dynamic success shards, anti-stuck flags, and a lightweight episode counter.
During retrieval, the program extracts a clean task brief, predicts the target domain and intent, collects candidates from static playbooks and dynamic success memories, and optionally adds anti-stuck nudges when recent failures suggest repeated actions, waiting, or interface blockage.
The retrieved payload is intentionally small, returning at most two concise hints with metadata.
This design reflects the website-specific nature of WebVoyager, where reusable navigation checklists and recovery hints are often more useful than long trajectory recall.

\paragraph{Mind2Web.}
Figure~\ref{fig:memo_mind2web} shows the selected memo program for Mind2Web.
Unlike the WebVoyager design, this program is organized around host-aware playbooks and task blueprints.
It maintains static host playbooks and micro-skills for recurring website widgets, while also updating trap counters and success recipes from interaction outcomes.
At retrieval time, it builds a fused text-image query from the task description and initial screenshots, ranks memories using multimodal similarity, domain match, trap signals, and success bonuses, and returns the task blueprint together with the top ranked memory items.
This structure matches Mind2Web, where tasks frequently require handling website-specific forms, dropdowns, date pickers, and recurring interaction traps.

\paragraph{AgentVista.}
Figure~\ref{fig:memo_agentvista} illustrates a more compact procedural memory learned for AgentVista.
The program initializes a library of visual-reasoning task families, where each family contains keywords, anchors, and concise hint items.
Retrieval routes the current question to a task family using token and ngram matches, anchor shortcuts, adaptive tag biases, and historical success priors.
If no family is confidently matched, the program falls back to a general image-first protocol that encourages the agent to inspect visible labels, numbers, legends, and on-screen text before using external tools.
During update, the program logs outcomes and adjusts routing biases based on success or failure.
This suggests that AgentVista benefits more from task-family-level visual reasoning procedures than from large episodic stores.

\paragraph{MMSearch-Plus.}
Figure~\ref{fig:memo_mmsearch_plus} shows the learned memo program for MMSearch-Plus.
The selected program combines a domain-tuned skillbook with cue-indexed episodic memory.
It extracts high-signal cues from task text and images, including OCR-derived text, quoted phrases, capitalized terms, numbers, years, and multilingual tokens.
These cues are used to detect a coarse domain and retrieve relevant structured episodes from persistent JSONL storage.
The returned payload contains at most a few episodic matches together with one domain-specific skillbook tip.
During update, the program records successful and failed search behavior, including tools, queries, final answers, and failure notes.
This design is well suited to MMSearch-Plus, where effective search often depends on identifying distinctive visual or textual evidence and converting it into targeted search queries.

\paragraph{Comparison across learned programs.}
Figures~\ref{fig:memo_webvoyager}--\ref{fig:memo_mmsearch_plus} show that \methodname does not simply discover a single universal memory template.
Instead, it adapts the internal organization of memory to the structure of each benchmark.
WebVoyager emphasizes domain-intent navigation playbooks and anti-stuck recovery; Mind2Web emphasizes host-aware task blueprints, widget skills, traps, and success recipes; AgentVista emphasizes task-family procedural routing; and MMSearch-Plus emphasizes cue extraction, episodic search memory, and domain-specific search skills.
Despite these differences, the learned programs share a common principle: they compress raw multimodal trajectories into compact, task-facing memory payloads rather than injecting full histories.
This supports the main observation that effective memory design requires selecting, abstracting, and presenting experience in a form matched to the target task distribution.

\section{Prompt Details}
\label{app:prompts}

This appendix reports the prompt templates used in our experiments. 
We group them according to their roles in the system: meta-agent prompts, execution-agent prompts, evaluation prompts, and baseline-specific prompts. 
When a prompt template is shared across multiple benchmarks or methods, we report it once and instantiate it with the corresponding task description, trajectory records, retrieved memories, or benchmark-specific metadata.

\subsection{Meta-Agent Prompts}
\label{app:meta_prompts}

The meta agent uses four prompt templates during memory-design search. 
First, the analysis prompt in Prompt~\ref{prompt:meta-agent-analysis} diagnoses the current memo program by inspecting its source code, sampled successful and failed trajectories, retrieved memories, benchmark scores, and previous improvement examples. 
It outputs structured feedback, including memory relevance labels, content-quality issues, structural weaknesses, and prioritized code-level suggestions. 
Second, the code-generation prompt in Prompt~\ref{prompt:meta-agent-generate-code} rewrites an existing memo program based on the analysis results and the observed performance. 
Third, at the beginning of search or when no previous source is available, the initialization prompt in Prompt~\ref{prompt:meta-agent-generate-code-no-source} generates a new memo program directly from the task description and the memo-program interface. 
Finally, when a generated program fails static checks or runtime execution, the repair prompt in Prompt~\ref{prompt:meta-agent-reflection-repair} asks the meta agent to preserve the intended memory design while fixing implementation errors.

These prompts jointly define the mutation process in our design-search loop. 
They expose the same memo-program contract to the meta agent: each candidate must implement \texttt{retrieve} and \texttt{update}, return a valid retrieved-memory payload, and respect the text and image budgets used by the execution agent. 
The prompts also specify the available implementation utilities, including multimodal chat calls, embedding clients, vector search, and graph operations.

\subsection{Execution-Agent and Memory-Injection Prompts}
\label{app:execution_prompts}

The execution agent uses benchmark-family-specific system prompts. 
For GUI/web navigation tasks, we use the WebVoyager-style execution prompt in Prompt~\ref{prompt:webvoyager-execution}. 
For multimodal visual reasoning and tool-use tasks, we use the AgentVista-style execution prompt in Prompt~\ref{prompt:agentvista-execution}. 
These execution prompts define the agent's role, action format, observation usage, and termination behavior. 
All methods and baselines use the same execution prompt under the same benchmark and execution-model setting, so performance differences come from the memory mechanism rather than from prompt tuning.

Retrieved memories are injected through a shared prompt fragment shown in Prompt~\ref{prompt:retrieved-memory-injection}. 
This fragment serializes the \texttt{RetrievedMemoryPayload} returned by \texttt{retrieve}, including textual guidance, optional image references, and structured metadata. 
When visual observations are included in the retrieved memory, they are attached subject to the same image-budget constraints used throughout the experiments. 
For AgentVista-style tasks that require summarizing information returned by the visit tool, we use the tool-summary prompt in Prompt~\ref{prompt:agentvista-visit-tool-summary}.

\subsection{Evaluation Prompts}
\label{app:eval_prompts}

We use LLM-as-judge evaluation with benchmark-family-specific prompts, following the evaluation protocols of WebVoyager~\citep{he2024webvoyager} for GUI/web navigation and AgentVista~\citep{su2026agentvista} for visual reasoning.
The GUI/web navigation evaluator uses the prompt in Prompt~\ref{prompt:webvoyager-auto-eval}, which judges whether the final browser state or answer satisfies the user task. 
The visual reasoning evaluator uses the prompt in Prompt~\ref{prompt:agentvista-evaluation}, which checks answer correctness and consistency with the available evidence. 
Both evaluators receive the task instruction, trajectory information, final answer or final state, and a bounded number of screenshots. 
The evaluator outputs a binary correctness label, which is used as success rate for GUI/web navigation tasks and judge-based accuracy for visual reasoning tasks.

\promptcaption{Meta Agent Analysis Prompt}{prompt:meta-agent-analysis}
\begin{promptbox}
\ 
\textbf{Meta Agent Analysis Prompt} \\

\textbf{[SYSTEM]}

You are a **Senior Agent Construction Engineer** responsible for provide suggestions for a memory structure written by a entry level engineer, to make the memory structure better for downstream agent to finish tasks. \\
\#\#\# Memo Information Overview \\
1. **source\_code** \\
    - A `MemoStructure` implementation coordinates memory; use private helpers or small inner abstractions for clarity. \\
    - **retrieve** / **update** are the only hooks the runtime calls. \\
    - code usage: Your memory structure will be used in the agent workflow: \\
        - `retrieve(recorder)`: used **before** executing the task. It returns a ``RetrievedMemoryPayload`` (see schema below). The worker injects a JSON text summary of that payload plus any referenced images into the execution agent's first user message. \\
        - `update(recorder)`: used **after** task is finished, to update the trajectory, reward, or other information. \\
\  \\
\#\#\# RetrievedMemoryPayload (return type of ``MemoStructure.retrieve``) \\
``retrieve`` must return a **JSON-serializable** ``RetrievedMemoryPayload`` (import from ``meta\_self\_evolve.contracts.types``), i.e. a dict with: \\
\  \\
- ``items``: a **flat** list of memory dicts; each dict may include: \\
  - ``text`` (``str``, optional): guidance for the execution agent. \\
  - ``images`` (optional): list of image refs with the same JSON shape as trajectory ``ImageRef``: ``\{"kind": "path"|"url", "value": "<relative path or URL>", "mime": "<optional>"\}``. When ``kind`` is ``"path"``, ``value`` is resolved relative to the episode **artifact root**. \\
  - ``metadata`` (``dict``, optional): any structured fields (e.g. ``memory\_id``, ``score``, ``source``, layer names). \\
- ``metadata`` (``dict``): episode-level retrieval notes (use ``\{\}`` if none). \\
\  \\
**Empty retrieval:** return ``\{"items": [], "metadata": \{\}\}`` (or use ``empty\_retrieved\_memory\_payload()`` from ``meta\_self\_evolve.common.retrieved\_memory``). \\
\  \\
**Image / text budgets (truncation is expected):** \\
- Put **only a small number** of images in ``items[*].images`` (typically 0–2). Extra images are **dropped**; do not rely on the model seeing every path you emit. \\
- The JSON text of the payload is also **truncated** to ``RolloutLimits.max\_retrieved\_memory\_chars`` in the execution agent's first user turn. \\
- **Execution agent:** at most ``RolloutLimits.max\_retrieved\_memory\_images`` images are loaded from the payload (in ``items`` order) on the first user turn; this is **separate** from the per-step screenshot budget ``RolloutLimits.max\_images\_per\_step`` for environment observations. \\
- **Meta analysis:** after the truncated JSON block, at most ``MetaTrajectoryLimits.meta\_retrieved\_memory\_images\_per\_episode`` payload images are attached per sampled episode (again in order); this is **separate** from ``MetaTrajectoryLimits.max\_meta\_images\_per\_episode`` for trajectory screenshots. \\
\  \\
**Runtime:** the worker injects this into the first user turn as (1) a text block (header + pretty-printed JSON of the payload), then (2) ``image\_url`` parts for each referenced image, subject to the limits above. \\
\  \\
2. **examples** \\
    - **examples**: sampled retrieve trajectories, split into **FAILED TRAJECTORIES** and **SUCCESSFUL TRAJECTORIES**. \\
    - In each trajectory section below, every episode may include a text block starting with `retrieved\_memory (JSON from retrieve before this episode):` — that is the serialized ``RetrievedMemoryPayload`` from ``retrieve`` (truncated in logs); use it together with trajectory screenshots to judge whether retrieval helped. \\
3. **benchmark\_eval\_score** \\
    -  performance(success rate) of current memory stucture + general agent system. Need to use the score to analyze the performance and bottleneck of current memory structure. \\
\  \\
\#\#\# Your Task: \\
You will analyze past suggestion examples(including past source code, suggestions, and the improve score it led to) and the current retrieved trajectories and memory source code, then produce concrete, prioritized suggestions to improve the memory structure. \\
Follow the numbered procedure below and produce the requested structured outputs. \\
Step 1 — Learn from past suggestions \& the improve score \\
    1. Look at the provided improve\_score (positive → improvement, negative → degradation) and the single suggestion\_example that produced that score. \\
    2. Explain why that suggestion led to improvement or degradation: \\
        - What pattern in the change made it succeed or fail? \\
        - Which behaviors, assumptions, or shortcuts in that suggestion were helpful? Which were harmful? \\
        - From these concrete cases, extract 2–5 general principles to adopt and 2–5 pitfalls to avoid when creating future suggestions. \\
\  \\
Step 2 — Inspect sampled trajectories and benchmark performance and decide which memories are useful \\
    1. Review the sampled FAILED TRAJECTORIES first, then the SUCCESSFUL TRAJECTORIES. For each episode, use the `retrieved\_memory` JSON block when present (it is omitted when empty). \\
    2. Compare the two groups: identify which retrieved memories or retrieval patterns appear in successful runs, and which missing / noisy / misleading memories correlate with failed runs. \\
    3. For each retrieved memory item (or memory group) returned for the trajectory, label it as one of: \\
        - Useful \& Relevant — clearly applies to the current situation and can guide action; \\
        - Potentially Useful — has value but needs reformatting, summarization, or indexing to be helpful; \\
        - Irrelevant / Confusing — not related to this trajectory or misleading; \\
        - Empty / Badly Formatted — blank, placeholder, or not parseable.  \\
    4. For each memory you mark Useful/Potentially Useful, say how it would help (e.g., provides a repeated subgoal, highlights a trap, identifies key object interactions). \\
    5. For Irrelevant/Empty items, explain why they failed retrieval combine with the memory source code (e.g., wrong keying, over-specific content, missing summarization). \\
\  \\
Step 3 — Inspect memory source and produce concrete suggestions \\
    1. Review the memory source code (retrieval keys, indexing, storage format, layers). Using Step 1 principles and Step 2 labels, propose specific changes to the memory system that address the observed issues. \\
    2. Combined the memory source code with your analysis in step 2, giving suggestions. For each suggested change, include: \\
        - What to change (code-level or pipeline change, e.g., add summarization layer, change indexing key, normalize objects to noun-phrases). \\
        - Why it will help (link back to a principle or a concrete failing you observed). \\
    3. Prioritize suggestions: label them High / Medium / Low priority and give an implementation order. \\
    4. Link Analysis to Benchmark Performance \\
        - Use benchmark\_eval\_score to identify which structural weaknesses correlate with poor performance. \\
\  \\
Extra checks (quality \& coherence) \\
    1. Flag obvious content issues: duplicates, empty entries, raw dumps, mis-typed fields, or numeric types that break JSON serialization. \\
    2. Check layer interaction: do layers pass structured outputs to each other, or only dump free-form text? \\
    3. If retrieval returns empty lists or dicts, emphasize structural fixes (keying, ensure type consistency, avoid over-relying on try/except fallbacks). \\
\  \\
Goal: Combine reflection on past improvement signals with current system diagnosis to produce actionable, high-level suggestions that strengthen memory structure quality. \\
\  \\
\#\#\# Benchmark Information: \\
\fstring{task\_description} \\
\  \\
\#\#\# Required Output: \\
Return a JSON object followed below json schema: \\
\{ \\
  "learned\_from\_suggestion\_example": \{ \\
    "type": "string", \\
    "description": "Findings derived from the provided suggestion\_example and improve\_score. Bullet list of concrete factors (patterns) that made the suggestion succeed or fail. And principles to adopt when making future suggestions." \\
  \}, \\
  "trajectory\_score\_assessment": \{ \\
    "type": "array", \\
    "description": "analysis each retrieved module information based on current trajectories sampled and the benchmark scores.", \\
    "items": \{ \\
      "type": "object", \\
      "properties": \{ \\
        "label": \{ \\
          "type": "string", \\
          "enum": [ \\
            "Useful", \\
            "Potentially Useful", \\
            "Irrelevant", \\
            "Empty/BadFormat" \\
          ], \\
          "description": "Categorization of the memory item's relevance based on whether the retrieved content actually helps the agent." \\
        \}, \\
        "how\_it\_can\_help": \{ \\
          "type": "string", \\
          "description": "If Useful/Potentially Useful: short note how it could guide actions (subgoal, trap, object use...). If Irrelevant/Empty: reason (e.g., wrong keying, over-specific, missing summary, formatting)." \\
        \} \\
      \}, \\
      "required": [ \\
        "label", \\
        "how\_it\_can\_help" \\
      ] \\
    \} \\
  \}, \\
  "content\_quality\_issues": \{ \\
    "type": "string", \\
    "description": "Detected content-level problems (duplicates, empty entries, serialization issues...). Why those harms retrieval or downstream planning." \\
  \}, \\
  "structure\_and\_coherence": \{ \\
    "type": "string", \\
    "description": "Analysis of layer interactions, keying, and task-awareness. Which parts generalize, which are overfitted." \\
  \}, \\
  "suggested\_changes": \{ \\
    "type": "array", \\
    "description": "Based on all your analysis above, provide concrete change that can be applied on provided current memory structure code.", \\
    "items": \{ \\
      "type": "object", \\
      "properties": \{ \\
        "priority": \{ \\
          "type": "string", \\
          "enum": [ \\
            "High", \\
            "Medium", \\
            "Low" \\
          ], \\
          "description": "How urgent/impactful this is." \\
        \}, \\
        "what": \{ \\
          "type": "string", \\
          "description": "Precise description of what to change (code/pipeline/config)." \\
        \}, \\
        "why": \{ \\
          "type": "string", \\
          "description": "Link to observations/principles: why this addresses the problem." \\
        \} \\
      \}, \\
      "required": [ \\
        "priority", \\
        "what", \\
        "why" \\
      ] \\
    \} \\
  \} \\
\} \\

\textbf{[USER]}

<Suggestion Example> \\
    Here is a previous suggestion, along with the code it looked at. 4.  \\
    The example include a memory structure and it's modification attempt, annotated with an improvement score (positive = improved, negative = degraded).  \\
    Infer the underlying patterns that differentiate effective modifications from harmful ones, and apply this reasoning to suggest an improved modification for the current memory structure. \\
    "\fstring{improve\_example}" \\
    </Suggestion Example> \\
     \\
<CURRENT SOURCE CODE> \\
\fstring{source\_code} \\
</CURRENT SOURCE CODE> \\
<CURRENT TRAJECTORY EXAMPLES> \\
\fstring{trajectory\_examples} \\
</CURRENT TRAJECTORY EXAMPLES> \\
<CURRENT BENCHMARK SCORE> \\
\fstring{benchmark\_overall\_eval\_score} \\
</CURRENT BENCHMARK SCORE> \\

\end{promptbox}

\promptcaption{Meta Agent Generate New Code Prompt}{prompt:meta-agent-generate-code}
\begin{promptbox}
\ 
\textbf{Meta Agent Generate New Code Prompt} \\

\textbf{[SYSTEM]}

You are a senior AI software engineer. Your task is to build a clear, maintainable memory system for a downstream task agent. The agent will be used in \fstring{env\_id}. Your memory structure should supply relevant experience and reference beyond raw trajectory text. \\
     \\
\fstring{task\_description} \\
\  \\
You are given the following backbone (``MemoStructure`` with ``retrieve`` / ``update``): \\
<BACKBONE\_CODE> \\
"""Abstract memo backbone inlined into codegen prompts. \\
\  \\
Only :class:`MemoStructure` is required. Implement a clear, maintainable \\
``retrieve`` / ``update`` pair; use private helpers or inner classes as needed. \\
""" \\
\  \\
from \_\_future\_\_ import annotations \\
\  \\
from abc import ABC, abstractmethod \\
from typing import Any, Optional \\
\  \\
from meta\_self\_evolve.contracts.types import EpisodeRecorder, RetrievedMemoryPayload \\
\  \\
\  \\
class MemoStructure(ABC): \\
    def \_\_init\_\_(self) -> None: \\
        self.database: Optional[Any] = None \\
\  \\
    @abstractmethod \\
    async def retrieve(self, recorder: EpisodeRecorder) -> RetrievedMemoryPayload: \\
        """Return structured memory for the next episode (see ``RetrievedMemoryPayload``).""" \\
        ... \\
\  \\
    @abstractmethod \\
    async def update(self, recorder: EpisodeRecorder) -> None: \\
        """Incorporate a finished trajectory (init, steps, reward, etc.).""" \\
        ... \\
\  \\
</BACKBONE\_CODE> \\
Subclass ``MemoStructure`` and import the types you need, for example: \\
```python \\
from meta\_self\_evolve.common.retrieved\_memory import empty\_retrieved\_memory\_payload \\
from meta\_self\_evolve.contracts.memo\_backbone import MemoStructure \\
from meta\_self\_evolve.contracts.types import EpisodeRecorder, RetrievedMemoryPayload \\
from meta\_self\_evolve.llm.client import MultimodalChatClient \\
from meta\_self\_evolve.llm.embedding import ( \\
    EmbeddingClient, \\
    embed\_item\_image\_path, \\
    embed\_item\_image\_url, \\
    embed\_item\_text, \\
) \\
\# plus your choice of Chroma, networkx, etc. \\
``` \\
\  \\
<CODE\_INPUT> \\
\  \\
    Your `retrieve` and `update` will take `EpisodeRecorder` as input, which has following attributes: \\
    \{ \\
  "init": \{ \\
    "description": "Initial task text and observation images.", \\
    "type": "InitRecord" \\
  \}, \\
  "steps": \{ \\
    "description": "Per-step action and observation history.", \\
    "type": "list[StepRecord]" \\
  \}, \\
  "memory\_retrieved": \{ \\
    "description": "Structured output from retrieve() before the episode.", \\
    "type": "RetrievedMemoryPayload" \\
  \}, \\
  "reward": \{ \\
    "description": "Scalar episode reward after finish.", \\
    "type": "float" \\
  \}, \\
  "messages": \{ \\
    "description": "Full LLM conversation history for this episode.", \\
    "type": "list[dict]" \\
  \} \\
\} \\
    - For `retrieve`, only leverage `.init` attribute. \\
    - For `update`, leverage `.init`, `.steps`, `reward` attribute. \\
    - Each element in the above dict is a attribute name as key, and description, type, and a example for the exact possible value the attribute could have. \\
    - please note that all provided current trajectory can already been seen by down stream agents(in history), your memory structure should focus on provide extra advice and reference for agents. \\
     \\
</CODE\_INPUT> \\
\  \\
<CODE\_USAGE> \\
Your memory structure will be used in the agent workflow: \\
    - `retrieve(recorder)`: used **before** executing the task; must return ``RetrievedMemoryPayload`` (see schema below). Populate ``items`` with concise ``text`` and optional ``images`` (``ImageRef`` JSON) so the execution agent receives useful guidance. \\
    - `update(recorder)`: used **after** task is finished, to update the trajectory, reward, or other information. \\
\  \\
\#\#\# RetrievedMemoryPayload (return type of ``MemoStructure.retrieve``) \\
``retrieve`` must return a **JSON-serializable** ``RetrievedMemoryPayload`` (import from ``meta\_self\_evolve.contracts.types``), i.e. a dict with: \\
\  \\
- ``items``: a **flat** list of memory dicts; each dict may include: \\
  - ``text`` (``str``, optional): guidance for the execution agent. \\
  - ``images`` (optional): list of image refs with the same JSON shape as trajectory ``ImageRef``: ``\{"kind": "path"|"url", "value": "<relative path or URL>", "mime": "<optional>"\}``. When ``kind`` is ``"path"``, ``value`` is resolved relative to the episode **artifact root**. \\
  - ``metadata`` (``dict``, optional): any structured fields (e.g. ``memory\_id``, ``score``, ``source``, layer names). \\
- ``metadata`` (``dict``): episode-level retrieval notes (use ``\{\}`` if none). \\
\  \\
**Empty retrieval:** return ``\{"items": [], "metadata": \{\}\}`` (or use ``empty\_retrieved\_memory\_payload()`` from ``meta\_self\_evolve.common.retrieved\_memory``). \\
\  \\
**Image / text budgets (truncation is expected):** \\
- Put **only a small number** of images in ``items[*].images`` (typically 0–2). Extra images are **dropped**; do not rely on the model seeing every path you emit. \\
- The JSON text of the payload is also **truncated** to ``RolloutLimits.max\_retrieved\_memory\_chars`` in the execution agent's first user turn. \\
- **Execution agent:** at most ``RolloutLimits.max\_retrieved\_memory\_images`` images are loaded from the payload (in ``items`` order) on the first user turn; this is **separate** from the per-step screenshot budget ``RolloutLimits.max\_images\_per\_step`` for environment observations. \\
- **Meta analysis:** after the truncated JSON block, at most ``MetaTrajectoryLimits.meta\_retrieved\_memory\_images\_per\_episode`` payload images are attached per sampled episode (again in order); this is **separate** from ``MetaTrajectoryLimits.max\_meta\_images\_per\_episode`` for trajectory screenshots. \\
\  \\
**Runtime:** the worker injects this into the first user turn as (1) a text block (header + pretty-printed JSON of the payload), then (2) ``image\_url`` parts for each referenced image, subject to the limits above. \\
\  \\
</CODE\_USAGE> \\
\  \\
Here is the basic tools provided: \\
<GRAPH\_DATABASE\_INTERACTION> \\
\  \\
NETWORKX GRAPH CHEATSHEET \\
\  \\
Context:  \\
    import networkx as nx \\
    G = nx.Graph() \\
\  \\
1. NODE OPERATIONS \\
- G.add\_node(node, **attrs): Add a single node with optional attributes. \\
- G.add\_nodes\_from([n1, n2], **common\_attrs): Add multiple nodes at once (shared attributes apply to all). \\
- G.remove\_node(node): Remove a node and all edges connected to it. \\
- G.remove\_nodes\_from([n1, n2]): Remove multiple nodes. \\
- node in G: Check if a node exists. \\
- G.nodes: Get all nodes (NodeView). \\
- G.nodes[node]: Access node attributes as a dict. \\
- nx.set\_node\_attributes(G, \{node: \{"attr": value\}\}): Set attributes for nodes. \\
\  \\
2. EDGE OPERATIONS \\
- G.add\_edge(u, v, **attrs): Add an edge between two nodes. \\
- G.add\_edges\_from([(u, v), (x, y)], **attrs): Add multiple edges at once (shared attributes apply to all). \\
- G.remove\_edge(u, v): Remove a single edge. \\
- G.remove\_edges\_from([(u, v), (x, y)]): Remove multiple edges. \\
- G.has\_edge(u, v): Check if an edge exists. \\
- G.edges: Get all edges (EdgeView). \\
- G.edges[(u, v)]: Access edge attributes as a dict. \\
- nx.set\_edge\_attributes(G, \{(u, v): \{"weight": 1.0\}\}): Set attributes for edges. \\
\  \\
3. TRAVERSAL / NEIGHBORHOOD \\
- G.neighbors(node): Get neighbors of a node. \\
- G.adj[node]: Get dict of neighbors with edge data. \\
- nx.shortest\_path(G, source, target): Find one shortest path between nodes. \\
- nx.shortest\_path\_length(G, source, target): Get shortest path length. \\
- nx.all\_simple\_paths(G, source, target, cutoff): Generate all simple paths up to a cutoff length. \\
- nx.connected\_components(G): Get connected components as node sets. \\
- G.subgraph([n1, n2, n3]): Extract a subgraph induced by given nodes. \\
\  \\
4. ANALYSIS / CENTRALITY \\
- G.degree(node): Get degree (number of edges) for a single node. \\
- G.degree(): Get degree for all nodes (DegreeView). \\
- nx.degree\_centrality(G): Compute degree centrality (dict of node -> score). \\
- nx.betweenness\_centrality(G): Compute betweenness centrality. \\
- nx.pagerank(G): Compute PageRank scores for nodes. \\
- nx.clustering(G): Compute local clustering coefficient. \\
- nx.is\_connected(G): Check if graph is connected. \\
- nx.number\_connected\_components(G): Count connected components. \\
\  \\
5. UTILITIES \\
- G.copy(): Make a copy of the graph. \\
- G.clear(): Remove all nodes and edges. \\
- nx.to\_dict\_of\_dicts(G): Convert graph to adjacency dict. \\
- nx.to\_numpy\_array(G): Get adjacency matrix as a NumPy array. \\
\  \\
</GRAPH\_DATABASE\_INTERACTION> \\
\  \\
<CHROMA\_DATABASE\_INTERACTION> \\
\#\# Initialize Chroma DB \\
\  \\
Import: `from langchain\_chroma import Chroma` \\
\  \\
Use `embedder = EmbeddingClient()` and `db = Chroma(embedding\_function=embedder)` to create the database. DO NOT use persist\_dir. \\
\  \\
The default LangChain helpers (`add\_texts`, `similarity\_search(query: str)`) are **text-only** on the query side: `similarity\_search` takes a string query, which `EmbeddingClient` embeds as a **single fused text vector** (DashScope multimodal embedding in text mode). \\
\  \\
For **screenshots, multiple images, or text+image fusion**, do **not** assume `similarity\_search` sees pixels. Build fused vectors explicitly with `await embedder.get\_fused\_embedding([...])` (see Embedding Client in TOOLS). One `get\_fused\_embedding` call returns **one** fused vector; need multiple vectors → call **multiple times**. \\
\  \\
Recommended multimodal storage / retrieval pattern: \\
- Keep the **human-readable memory content** (summary, advice, metadata) in Chroma / graph / your normal memory structure. \\
- Store each multimodal fused vector in a **sidecar index** keyed by a stable `memory\_id` (for example a dict or list storing `\{memory\_id, embedding, metadata\}`). \\
- At retrieve time, build **one fused query vector**, compare it against stored multimodal vectors (for example with `EmbeddingClient.compute\_one\_to\_group\_similarity(...)`), rank by similarity, then map the top hits back to `memory\_id` and finally return the linked readable memory content. \\
- For recorder screenshots, local image paths like `recorder.init.images[i].value` are relative to the current episode artifact root. In normal worker execution, `EmbeddingClient` can resolve those relative paths automatically from runtime context. If you work with images outside that runtime, pass `base\_dir=` explicitly. \\
\  \\
\#\#\# Add Memory: Adds new text entries to the database and returns their unique IDs.  \\
\  \\
db.add\_texts( \\
    texts: List[str], \\
    metadatas: Optional[Union[str, int, float, bool, None]] = None, \\
    ids: Optional[List[str]] = None \\
) -> List[str] \\
\  \\
- metadatas must be **flat list**: each value must be a single primitive type (str, int, float, bool, or None). \\
- You cannot pass lists, nested dicts, or other complex objects. \\
- If you need to store structured data, serialize it to a JSON string: \\
\  \\
\#\#\# Retrieve Memory \\
\  \\
db.similarity\_search( \\
    query: str, \\
    k: int = 4 \\
) -> List[Document] \\
\  \\
`query` is plain text only (text embedding path). For multimodal queries, use `get\_fused\_embedding(...)`, run your own similarity step, then map results back to stored memory items. \\
\  \\
return List[Document]: [ \\
  Document( \\
    page\_content="the agent found a key", \\
    metadata=\{"type": "item"\} \\
  ) \\
] \\
\  \\
\#\#\# Get by ID \\
db.get( \\
    ids: Optional[List[str]] = None \\
) -> Dict[str, List] \\
\  \\
\#\#\# Delete Memory \\
db.delete( \\
    ids: Optional[List[str]] = None \\
) -> None \\
     \\
</CHROMA\_DATABASE\_INTERACTION> \\
\  \\
<OTHER\_TOOLS> \\
\  \\
TOOLS AVAILABLE: \\
\  \\
1. Multimodal Chat Client \\
Class: MultimodalChatClient \\
\  \\
- Purpose: Asynchronous wrapper around OpenAI-compatible Chat Completions for text or multimodal user content. Use this when you need summarisation, synthesis, planning, or structured JSON output inside the memory code. \\
- Initialization (memory structure code — **must** match the benchmark execution model): \\
    from meta\_self\_evolve.llm.client import MultimodalChatClient \\
    client = MultimodalChatClient() \\
- Do **not** pass `model=...` inside generated memory code. \\
- Key Methods: \\
    - await client.complete(messages: List[Dict[str, Any]]) -> str: \\
        Send a standard chat message list and return plain text. \\
    - await client.complete\_with\_system(*, system\_prompt: str, user\_input: str | list, history: Optional[List[Dict]] = None, ...) -> str: \\
        Convenience wrapper for one system prompt + one user input (+ optional history). **All parameters after `client` are keyword-only** (the real signature uses `*` before `system\_prompt=`). \\
    - await client.complete\_json\_object(messages: List[Dict[str, Any]], *, temperature: float | None = None) -> Dict[str, Any]: \\
        Request a JSON object response. Pass `messages` positionally; `temperature=` is keyword-only if needed. \\
    - await client.complete\_json\_with\_system(*, system\_prompt: str, user\_input: str | list, history: Optional[List[Dict]] = None, temperature: float | None = None) -> Dict[str, Any]: \\
        Convenience wrapper for structured JSON output with a system prompt. **All parameters after `client` are keyword-only** (the real signature uses `*` before `system\_prompt=`). \\
- Usage Example: \\
    reply = await client.complete\_with\_system( \\
        system\_prompt="You summarize browser trajectories.", \\
        user\_input="Summarize the key failure pattern in these steps.", \\
    ) \\
    structured = await client.complete\_json\_with\_system( \\
        system\_prompt="Return ONLY a JSON object matching the schema described below.", \\
        user\_input="...", \\
    ) \\
\  \\
- IMPORTANT: \\
    - **Never** call `complete\_with\_system(...)` or `complete\_json\_with\_system(...)` with positional arguments (e.g. `complete\_json\_with\_system(system\_prompt, user\_input)` will raise `TypeError`). Always use `system\_prompt=...`, `user\_input=...`, and optionally `history=...`, `temperature=...` as keywords. \\
    - If you need structured output, use `complete\_json\_object(...)` or `complete\_json\_with\_system(...)`. \\
    - JSON output schemas should still be described clearly inside your system prompt. Format example: \\
\{ \\
    "location": \{ \\
        "type": "string", \\
        "description": "The location to get the weather for" \\
    \}, \\
    "unit": \{ \\
        "type": ["string", "null"], \\
        "description": "The unit to return the temperature in", \\
        "enum": ["F", "C"] \\
    \} \\
\} \\
\  \\
    - Maintain history explicitly by passing prior messages in `history=` or by constructing the full `messages` list yourself. \\
\  \\
2. Embedding Client \\
Class: EmbeddingClient \\
\  \\
- Purpose: DashScope **qwen3-vl-embedding** (fixed model and fixed vector dimension inside the class). Supports **fused** vectors for text-only, image(s), or text+image mixed inputs. Optional cosine similarity helpers. \\
- Initialization (do **not** pass an embedding model name): \\
    from meta\_self\_evolve.llm.embedding import EmbeddingClient, embed\_item\_text, embed\_item\_image\_path, embed\_item\_image\_url \\
    embedder = EmbeddingClient(retries=3, retry\_delay=1.0) \\
- Key Methods: \\
    - await embedder.get\_embedding(text: str) -> List[float]: \\
        Single **text** string → one fused vector (text-only path). \\
    - await embedder.get\_batch\_embeddings(texts: List[str]) -> List[List[float]]: \\
        One fused vector per string (independent calls). \\
    - await embedder.get\_fused\_embedding(items: List, base\_dir: Optional[Path | str] = None) -> List[float]: \\
        **Multimodal fusion**: `items` is an ordered list of parts (from the helpers above): \\
        - `embed\_item\_text("...")`  → `\{type:'text', text:...\}` \\
        - `embed\_item\_image\_path("relative/or/abs/path.png")`  → local image; if relative, `EmbeddingClient` resolves it against `base\_dir=` when provided, otherwise against the worker's current artifact-root runtime context. \\
        - `embed\_item\_image\_url("https://...")`  → image URL. \\
        One call uses `enable\_fusion=True` and returns **one** vector. If you need **multiple** vectors (e.g. separate vectors per document), call `get\_fused\_embedding` **multiple times**—do not pack unrelated fusion targets into one call expecting multiple vectors. \\
    - await EmbeddingClient.compute\_similarity(emb1: List[float], emb2: List[float], metric: str = "cosine") -> float: \\
        Computes similarity between two embeddings asynchronously. \\
    - await EmbeddingClient.compute\_one\_to\_group\_similarity(emb: List[float], group\_emb: List[List[float]], metric: str = "cosine") -> List[float]: \\
        Computes similarity between one embedding and a group of embeddings asynchronously. \\
    - `EmbeddingClient()` also implements `embed\_query()` / `embed\_documents()` / `\_\_call\_\_` so it can be passed into Chroma's `embedding\_function=...` for **text** document/query paths. \\
\  \\
- Notes: \\
    * Requires `DASHSCOPE\_API\_KEY` in the environment for DashScope. \\
    * For recorder screenshots in normal evaluation/runtime, relative image paths can usually be passed directly because the worker sets an artifact-root runtime context. If you use the embedder outside that runtime, pass `base\_dir=` explicitly. \\
    * Similarity helpers support cosine similarity and run in parallel for efficiency. \\
    * Do not import from `utils.hire\_agent` or `evals.utils.hire\_agent`; use `meta\_self\_evolve.llm.*` only. \\
\  \\
</OTHER\_TOOLS> \\
\  \\
\#\#\# Tooling constraints (memory structure code) \\
- Use `MultimodalChatClient()` with **no** `model=` argument so chat calls use the same model as the benchmark agent (`--execution-model`). \\
- Use `EmbeddingClient()` with **no** embedding model argument; it uses the built-in DashScope multimodal embedding with a fixed dimension. \\
- For text+image or multi-image **fusion** vectors, use `await embedder.get\_fused\_embedding([...], base\_dir=...)` with `embed\_item\_text` / `embed\_item\_image\_path` / `embed\_item\_image\_url`. In normal worker execution, recorder screenshot paths can usually be passed directly without `base\_dir=` because the worker sets artifact-root runtime context. **One** call returns **one** fused vector; for multiple independent vectors, call **multiple times**. \\
- For multimodal memory, keep a sidecar vector index keyed by `memory\_id`, then retrieve by fused query vector → similarity ranking → map top hits back to readable memory content. \\
- Do not hard-code OpenAI embedding model names or arbitrary chat model IDs in memory code. \\
\  \\
\#\#\# Your Task: \\
Modify the code above so that it fully satisfies the following design goals: \\
\  \\
1. **Modular, clear design:**   \\
   - Prefer a small, readable structure (private helpers, inner classes, or a single `MemoStructure` with well-named methods). Avoid unnecessary scaffolding.   \\
   - If you use several stores (e.g. Chroma, graph), give each a clear role and data shape. \\
\  \\
2. **Retrieve / update orchestration:**   \\
   - Implement `MemoStructure` with `retrieve` / `update`.   \\
   - `retrieve()` can chain internal steps when that helps.   \\
   - `update()` should persist new evidence from the finished trajectory in a consistent order. \\
\  \\
3. **Out-of-the-Box Reasoning:**   \\
   - Do not just mechanically call each layer one by one — think about the **semantic flow of information**.   \\
   - Consider cases like:   \\
     - what type of memory layer can be used according to the analysis result or task description, with the aim to better assist the agent to finish it's task?  \\
     - what order and input output should be best suitable for the analysis result or task description?  \\
     - Think about high-level stategy: what can be a good memory structure, and has good ability to transfer to other area?  \\
   - Make sure each layer plays a meaningful role in the system. \\
   - Directly perform simple plans or content writing based on if/else patterns should be avoided, since this will hurt the transfer ability. \\
   - Keep the retrieved memory clean and useful, aviod cutting off meaningful texts, repeat same patterns in the retrieved memory. \\
\  \\
4. **Integration with Utilities:**   \\
   - Feel free to use any provided utility functions (e.g., similarity calculation, interaction with databases, hire new agent) if relevant. The tools available will be listed in `TOOLS` section. \\
   - You can also create your own tools if neccessary, think out of the box. \\
\  \\
5. **Code Quality:**   \\
   - Output clean, runnable Python code following PEP8.   \\
   - Ensure `retrieve()` and `update()` accept `EpisodeRecorder` and orchestrate the pipeline end-to-end.   \\
   - Initialize any stores or clients in `MemoStructure.\_\_init\_\_` as needed. \\
   - Define a module-level factory function exactly as `def build\_memo() -> MemoStructure`, and return an instance of your top-level `MemoStructure` subclass. \\
   - Do not overuse defensive programming; raise appropriate exceptions when unexpected conditions occur to facilitate debugging. \\
\  \\
6. **Coherent Policy Logic:** \\
    - Avoid placeholders like pass or \# TODO. \\
    - Avoid hard-coded if/else branches or enumerated case handling; instead, express the logic through modular policy functions, scoring mechanisms, or composable decision rules. \\
    - Instead of enumerating case-specific rules, express generalizable principles that could apply across different families or new unseen tasks. \\
    - The logic should be adaptable and compositional, not dependent on predefined constants or string names. \\
    - Use abstractions instead of specific family identifiers. \\
\  \\
The goal is to ensure the memory policy behaves consistently across tasks and supports generalization, not to hard-code specific task behaviors. \\
\  \\
\#\#\# Important: \\
- Think creatively about data flow — outputs of one layer can feed into the next. \\
- Each layer's functionality and stored data should be clearly designed. \\
- Use project-local imports under `meta\_self\_evolve.*`; do not import from `utils.hire\_agent` or `evals.utils.hire\_agent`. \\
- Provide **only the final rewritten code**, no explanations. \\

\textbf{[USER]}

        Here is the current code that you must edit: \\
        <CURRENT\_CODE> \\
        \fstring{source\_code} \\
        </CURRENT\_CODE> \\
\  \\
        Here is the score of current code: \\
        <REWARD> \\
        \fstring{benchmark\_overall\_eval\_score} \\
        </REWARD> \\
\  \\
        Here is the analysis result (suggestions): \\
        <ANALYSIS\_RESULT> \\
        \{ \\
  "trajectory\_score\_assessment": "\fstring{trajectory\_score\_assessment}", \\
  "suggested\_changes": "\fstring{suggested\_changes}" \\
\} \\
        </ANALYSIS\_RESULT> \\
             \\

\end{promptbox}

\promptcaption{Meta Agent Generate New Code Prompt Without Existing Source}{prompt:meta-agent-generate-code-no-source}
\begin{promptbox}
\ 
\textbf{Meta Agent Generate New Code Prompt Without Existing Source} \\

\textbf{[SYSTEM]}

You are a senior AI software engineer. Your task is to build a clear, maintainable memory system for a downstream task agent. The agent will be used in \fstring{env\_id}. Your memory structure should supply relevant experience and reference beyond raw trajectory text. \\
     \\
\fstring{task\_description} \\
\  \\
You are given the following backbone (``MemoStructure`` with ``retrieve`` / ``update``): \\
<BACKBONE\_CODE> \\
"""Abstract memo backbone inlined into codegen prompts. \\
\  \\
Only :class:`MemoStructure` is required. Implement a clear, maintainable \\
``retrieve`` / ``update`` pair; use private helpers or inner classes as needed. \\
""" \\
\  \\
from \_\_future\_\_ import annotations \\
\  \\
from abc import ABC, abstractmethod \\
from typing import Any, Optional \\
\  \\
from meta\_self\_evolve.contracts.types import EpisodeRecorder, RetrievedMemoryPayload \\
\  \\
\  \\
class MemoStructure(ABC): \\
    def \_\_init\_\_(self) -> None: \\
        self.database: Optional[Any] = None \\
\  \\
    @abstractmethod \\
    async def retrieve(self, recorder: EpisodeRecorder) -> RetrievedMemoryPayload: \\
        """Return structured memory for the next episode (see ``RetrievedMemoryPayload``).""" \\
        ... \\
\  \\
    @abstractmethod \\
    async def update(self, recorder: EpisodeRecorder) -> None: \\
        """Incorporate a finished trajectory (init, steps, reward, etc.).""" \\
        ... \\
\  \\
</BACKBONE\_CODE> \\
Subclass ``MemoStructure`` and import the types you need, for example: \\
```python \\
from meta\_self\_evolve.common.retrieved\_memory import empty\_retrieved\_memory\_payload \\
from meta\_self\_evolve.contracts.memo\_backbone import MemoStructure \\
from meta\_self\_evolve.contracts.types import EpisodeRecorder, RetrievedMemoryPayload \\
from meta\_self\_evolve.llm.client import MultimodalChatClient \\
from meta\_self\_evolve.llm.embedding import ( \\
    EmbeddingClient, \\
    embed\_item\_image\_path, \\
    embed\_item\_image\_url, \\
    embed\_item\_text, \\
) \\
\# plus your choice of Chroma, networkx, etc. \\
``` \\
\  \\
<CODE\_INPUT> \\
\  \\
    Your `retrieve` and `update` will take `EpisodeRecorder` as input, which has following attributes: \\
    \{ \\
  "init": \{ \\
    "description": "Initial task text and observation images.", \\
    "type": "InitRecord" \\
  \}, \\
  "steps": \{ \\
    "description": "Per-step action and observation history.", \\
    "type": "list[StepRecord]" \\
  \}, \\
  "memory\_retrieved": \{ \\
    "description": "Structured output from retrieve() before the episode.", \\
    "type": "RetrievedMemoryPayload" \\
  \}, \\
  "reward": \{ \\
    "description": "Scalar episode reward after finish.", \\
    "type": "float" \\
  \}, \\
  "messages": \{ \\
    "description": "Full LLM conversation history for this episode.", \\
    "type": "list[dict]" \\
  \} \\
\} \\
    - For `retrieve`, only leverage `.init` attribute. \\
    - For `update`, leverage `.init`, `.steps`, `reward` attribute. \\
    - Each element in the above dict is a attribute name as key, and description, type, and a example for the exact possible value the attribute could have. \\
    - please note that all provided current trajectory can already been seen by down stream agents(in history), your memory structure should focus on provide extra advice and reference for agents. \\
     \\
</CODE\_INPUT> \\
\  \\
<CODE\_USAGE> \\
Your memory structure will be used in the agent workflow: \\
    - `retrieve(recorder)`: used **before** executing the task; must return ``RetrievedMemoryPayload`` (see schema below). Populate ``items`` with concise ``text`` and optional ``images`` (``ImageRef`` JSON) so the execution agent receives useful guidance. \\
    - `update(recorder)`: used **after** task is finished, to update the trajectory, reward, or other information. \\
\  \\
\#\#\# RetrievedMemoryPayload (return type of ``MemoStructure.retrieve``) \\
``retrieve`` must return a **JSON-serializable** ``RetrievedMemoryPayload`` (import from ``meta\_self\_evolve.contracts.types``), i.e. a dict with: \\
\  \\
- ``items``: a **flat** list of memory dicts; each dict may include: \\
  - ``text`` (``str``, optional): guidance for the execution agent. \\
  - ``images`` (optional): list of image refs with the same JSON shape as trajectory ``ImageRef``: ``\{"kind": "path"|"url", "value": "<relative path or URL>", "mime": "<optional>"\}``. When ``kind`` is ``"path"``, ``value`` is resolved relative to the episode **artifact root**. \\
  - ``metadata`` (``dict``, optional): any structured fields (e.g. ``memory\_id``, ``score``, ``source``, layer names). \\
- ``metadata`` (``dict``): episode-level retrieval notes (use ``\{\}`` if none). \\
\  \\
**Empty retrieval:** return ``\{"items": [], "metadata": \{\}\}`` (or use ``empty\_retrieved\_memory\_payload()`` from ``meta\_self\_evolve.common.retrieved\_memory``). \\
\  \\
**Image / text budgets (truncation is expected):** \\
- Put **only a small number** of images in ``items[*].images`` (typically 0–2). Extra images are **dropped**; do not rely on the model seeing every path you emit. \\
- The JSON text of the payload is also **truncated** to ``RolloutLimits.max\_retrieved\_memory\_chars`` in the execution agent's first user turn. \\
- **Execution agent:** at most ``RolloutLimits.max\_retrieved\_memory\_images`` images are loaded from the payload (in ``items`` order) on the first user turn; this is **separate** from the per-step screenshot budget ``RolloutLimits.max\_images\_per\_step`` for environment observations. \\
- **Meta analysis:** after the truncated JSON block, at most ``MetaTrajectoryLimits.meta\_retrieved\_memory\_images\_per\_episode`` payload images are attached per sampled episode (again in order); this is **separate** from ``MetaTrajectoryLimits.max\_meta\_images\_per\_episode`` for trajectory screenshots. \\
\  \\
**Runtime:** the worker injects this into the first user turn as (1) a text block (header + pretty-printed JSON of the payload), then (2) ``image\_url`` parts for each referenced image, subject to the limits above. \\
\  \\
</CODE\_USAGE> \\
\  \\
Here is the basic tools provided: \\
<GRAPH\_DATABASE\_INTERACTION> \\
\  \\
NETWORKX GRAPH CHEATSHEET \\
\  \\
Context:  \\
    import networkx as nx \\
    G = nx.Graph() \\
\  \\
1. NODE OPERATIONS \\
- G.add\_node(node, **attrs): Add a single node with optional attributes. \\
- G.add\_nodes\_from([n1, n2], **common\_attrs): Add multiple nodes at once (shared attributes apply to all). \\
- G.remove\_node(node): Remove a node and all edges connected to it. \\
- G.remove\_nodes\_from([n1, n2]): Remove multiple nodes. \\
- node in G: Check if a node exists. \\
- G.nodes: Get all nodes (NodeView). \\
- G.nodes[node]: Access node attributes as a dict. \\
- nx.set\_node\_attributes(G, \{node: \{"attr": value\}\}): Set attributes for nodes. \\
\  \\
2. EDGE OPERATIONS \\
- G.add\_edge(u, v, **attrs): Add an edge between two nodes. \\
- G.add\_edges\_from([(u, v), (x, y)], **attrs): Add multiple edges at once (shared attributes apply to all). \\
- G.remove\_edge(u, v): Remove a single edge. \\
- G.remove\_edges\_from([(u, v), (x, y)]): Remove multiple edges. \\
- G.has\_edge(u, v): Check if an edge exists. \\
- G.edges: Get all edges (EdgeView). \\
- G.edges[(u, v)]: Access edge attributes as a dict. \\
- nx.set\_edge\_attributes(G, \{(u, v): \{"weight": 1.0\}\}): Set attributes for edges. \\
\  \\
3. TRAVERSAL / NEIGHBORHOOD \\
- G.neighbors(node): Get neighbors of a node. \\
- G.adj[node]: Get dict of neighbors with edge data. \\
- nx.shortest\_path(G, source, target): Find one shortest path between nodes. \\
- nx.shortest\_path\_length(G, source, target): Get shortest path length. \\
- nx.all\_simple\_paths(G, source, target, cutoff): Generate all simple paths up to a cutoff length. \\
- nx.connected\_components(G): Get connected components as node sets. \\
- G.subgraph([n1, n2, n3]): Extract a subgraph induced by given nodes. \\
\  \\
4. ANALYSIS / CENTRALITY \\
- G.degree(node): Get degree (number of edges) for a single node. \\
- G.degree(): Get degree for all nodes (DegreeView). \\
- nx.degree\_centrality(G): Compute degree centrality (dict of node -> score). \\
- nx.betweenness\_centrality(G): Compute betweenness centrality. \\
- nx.pagerank(G): Compute PageRank scores for nodes. \\
- nx.clustering(G): Compute local clustering coefficient. \\
- nx.is\_connected(G): Check if graph is connected. \\
- nx.number\_connected\_components(G): Count connected components. \\
\  \\
5. UTILITIES \\
- G.copy(): Make a copy of the graph. \\
- G.clear(): Remove all nodes and edges. \\
- nx.to\_dict\_of\_dicts(G): Convert graph to adjacency dict. \\
- nx.to\_numpy\_array(G): Get adjacency matrix as a NumPy array. \\
\  \\
</GRAPH\_DATABASE\_INTERACTION> \\
\  \\
<CHROMA\_DATABASE\_INTERACTION> \\
\#\# Initialize Chroma DB \\
\  \\
Import: `from langchain\_chroma import Chroma` \\
\  \\
Use `embedder = EmbeddingClient()` and `db = Chroma(embedding\_function=embedder)` to create the database. DO NOT use persist\_dir. \\
\  \\
The default LangChain helpers (`add\_texts`, `similarity\_search(query: str)`) are **text-only** on the query side: `similarity\_search` takes a string query, which `EmbeddingClient` embeds as a **single fused text vector** (DashScope multimodal embedding in text mode). \\
\  \\
For **screenshots, multiple images, or text+image fusion**, do **not** assume `similarity\_search` sees pixels. Build fused vectors explicitly with `await embedder.get\_fused\_embedding([...])` (see Embedding Client in TOOLS). One `get\_fused\_embedding` call returns **one** fused vector; need multiple vectors → call **multiple times**. \\
\  \\
Recommended multimodal storage / retrieval pattern: \\
- Keep the **human-readable memory content** (summary, advice, metadata) in Chroma / graph / your normal memory structure. \\
- Store each multimodal fused vector in a **sidecar index** keyed by a stable `memory\_id` (for example a dict or list storing `\{memory\_id, embedding, metadata\}`). \\
- At retrieve time, build **one fused query vector**, compare it against stored multimodal vectors (for example with `EmbeddingClient.compute\_one\_to\_group\_similarity(...)`), rank by similarity, then map the top hits back to `memory\_id` and finally return the linked readable memory content. \\
- For recorder screenshots, local image paths like `recorder.init.images[i].value` are relative to the current episode artifact root. In normal worker execution, `EmbeddingClient` can resolve those relative paths automatically from runtime context. If you work with images outside that runtime, pass `base\_dir=` explicitly. \\
\  \\
\#\#\# Add Memory: Adds new text entries to the database and returns their unique IDs.  \\
\  \\
db.add\_texts( \\
    texts: List[str], \\
    metadatas: Optional[Union[str, int, float, bool, None]] = None, \\
    ids: Optional[List[str]] = None \\
) -> List[str] \\
\  \\
- metadatas must be **flat list**: each value must be a single primitive type (str, int, float, bool, or None). \\
- You cannot pass lists, nested dicts, or other complex objects. \\
- If you need to store structured data, serialize it to a JSON string: \\
\  \\
\#\#\# Retrieve Memory \\
\  \\
db.similarity\_search( \\
    query: str, \\
    k: int = 4 \\
) -> List[Document] \\
\  \\
`query` is plain text only (text embedding path). For multimodal queries, use `get\_fused\_embedding(...)`, run your own similarity step, then map results back to stored memory items. \\
\  \\
return List[Document]: [ \\
  Document( \\
    page\_content="the agent found a key", \\
    metadata=\{"type": "item"\} \\
  ) \\
] \\
\  \\
\#\#\# Get by ID \\
db.get( \\
    ids: Optional[List[str]] = None \\
) -> Dict[str, List] \\
\  \\
\#\#\# Delete Memory \\
db.delete( \\
    ids: Optional[List[str]] = None \\
) -> None \\
     \\
</CHROMA\_DATABASE\_INTERACTION> \\
\  \\
<OTHER\_TOOLS> \\
\  \\
TOOLS AVAILABLE: \\
\  \\
1. Multimodal Chat Client \\
Class: MultimodalChatClient \\
\  \\
- Purpose: Asynchronous wrapper around OpenAI-compatible Chat Completions for text or multimodal user content. Use this when you need summarisation, synthesis, planning, or structured JSON output inside the memory code. \\
- Initialization (memory structure code — **must** match the benchmark execution model): \\
    from meta\_self\_evolve.llm.client import MultimodalChatClient \\
    client = MultimodalChatClient() \\
- Do **not** pass `model=...` inside generated memory code. \\
- Key Methods: \\
    - await client.complete(messages: List[Dict[str, Any]]) -> str: \\
        Send a standard chat message list and return plain text. \\
    - await client.complete\_with\_system(*, system\_prompt: str, user\_input: str | list, history: Optional[List[Dict]] = None, ...) -> str: \\
        Convenience wrapper for one system prompt + one user input (+ optional history). **All parameters after `client` are keyword-only** (the real signature uses `*` before `system\_prompt=`). \\
    - await client.complete\_json\_object(messages: List[Dict[str, Any]], *, temperature: float | None = None) -> Dict[str, Any]: \\
        Request a JSON object response. Pass `messages` positionally; `temperature=` is keyword-only if needed. \\
    - await client.complete\_json\_with\_system(*, system\_prompt: str, user\_input: str | list, history: Optional[List[Dict]] = None, temperature: float | None = None) -> Dict[str, Any]: \\
        Convenience wrapper for structured JSON output with a system prompt. **All parameters after `client` are keyword-only** (the real signature uses `*` before `system\_prompt=`). \\
- Usage Example: \\
    reply = await client.complete\_with\_system( \\
        system\_prompt="You summarize browser trajectories.", \\
        user\_input="Summarize the key failure pattern in these steps.", \\
    ) \\
    structured = await client.complete\_json\_with\_system( \\
        system\_prompt="Return ONLY a JSON object matching the schema described below.", \\
        user\_input="...", \\
    ) \\
\  \\
- IMPORTANT: \\
    - **Never** call `complete\_with\_system(...)` or `complete\_json\_with\_system(...)` with positional arguments (e.g. `complete\_json\_with\_system(system\_prompt, user\_input)` will raise `TypeError`). Always use `system\_prompt=...`, `user\_input=...`, and optionally `history=...`, `temperature=...` as keywords. \\
    - If you need structured output, use `complete\_json\_object(...)` or `complete\_json\_with\_system(...)`. \\
    - JSON output schemas should still be described clearly inside your system prompt. Format example: \\
\{ \\
    "location": \{ \\
        "type": "string", \\
        "description": "The location to get the weather for" \\
    \}, \\
    "unit": \{ \\
        "type": ["string", "null"], \\
        "description": "The unit to return the temperature in", \\
        "enum": ["F", "C"] \\
    \} \\
\} \\
\  \\
    - Maintain history explicitly by passing prior messages in `history=` or by constructing the full `messages` list yourself. \\
\  \\
2. Embedding Client \\
Class: EmbeddingClient \\
\  \\
- Purpose: DashScope **qwen3-vl-embedding** (fixed model and fixed vector dimension inside the class). Supports **fused** vectors for text-only, image(s), or text+image mixed inputs. Optional cosine similarity helpers. \\
- Initialization (do **not** pass an embedding model name): \\
    from meta\_self\_evolve.llm.embedding import EmbeddingClient, embed\_item\_text, embed\_item\_image\_path, embed\_item\_image\_url \\
    embedder = EmbeddingClient(retries=3, retry\_delay=1.0) \\
- Key Methods: \\
    - await embedder.get\_embedding(text: str) -> List[float]: \\
        Single **text** string → one fused vector (text-only path). \\
    - await embedder.get\_batch\_embeddings(texts: List[str]) -> List[List[float]]: \\
        One fused vector per string (independent calls). \\
    - await embedder.get\_fused\_embedding(items: List, base\_dir: Optional[Path | str] = None) -> List[float]: \\
        **Multimodal fusion**: `items` is an ordered list of parts (from the helpers above): \\
        - `embed\_item\_text("...")`  → `\{type:'text', text:...\}` \\
        - `embed\_item\_image\_path("relative/or/abs/path.png")`  → local image; if relative, `EmbeddingClient` resolves it against `base\_dir=` when provided, otherwise against the worker's current artifact-root runtime context. \\
        - `embed\_item\_image\_url("https://...")`  → image URL. \\
        One call uses `enable\_fusion=True` and returns **one** vector. If you need **multiple** vectors (e.g. separate vectors per document), call `get\_fused\_embedding` **multiple times**—do not pack unrelated fusion targets into one call expecting multiple vectors. \\
    - await EmbeddingClient.compute\_similarity(emb1: List[float], emb2: List[float], metric: str = "cosine") -> float: \\
        Computes similarity between two embeddings asynchronously. \\
    - await EmbeddingClient.compute\_one\_to\_group\_similarity(emb: List[float], group\_emb: List[List[float]], metric: str = "cosine") -> List[float]: \\
        Computes similarity between one embedding and a group of embeddings asynchronously. \\
    - `EmbeddingClient()` also implements `embed\_query()` / `embed\_documents()` / `\_\_call\_\_` so it can be passed into Chroma's `embedding\_function=...` for **text** document/query paths. \\
\  \\
- Notes: \\
    * Requires `DASHSCOPE\_API\_KEY` in the environment for DashScope. \\
    * For recorder screenshots in normal evaluation/runtime, relative image paths can usually be passed directly because the worker sets an artifact-root runtime context. If you use the embedder outside that runtime, pass `base\_dir=` explicitly. \\
    * Similarity helpers support cosine similarity and run in parallel for efficiency. \\
    * Do not import from `utils.hire\_agent` or `evals.utils.hire\_agent`; use `meta\_self\_evolve.llm.*` only. \\
\  \\
</OTHER\_TOOLS> \\
\  \\
\#\#\# Tooling constraints (memory structure code) \\
- Use `MultimodalChatClient()` with **no** `model=` argument so chat calls use the same model as the benchmark agent (`--execution-model`). \\
- Use `EmbeddingClient()` with **no** embedding model argument; it uses the built-in DashScope multimodal embedding with a fixed dimension. \\
- For text+image or multi-image **fusion** vectors, use `await embedder.get\_fused\_embedding([...], base\_dir=...)` with `embed\_item\_text` / `embed\_item\_image\_path` / `embed\_item\_image\_url`. In normal worker execution, recorder screenshot paths can usually be passed directly without `base\_dir=` because the worker sets artifact-root runtime context. **One** call returns **one** fused vector; for multiple independent vectors, call **multiple times**. \\
- For multimodal memory, keep a sidecar vector index keyed by `memory\_id`, then retrieve by fused query vector → similarity ranking → map top hits back to readable memory content. \\
- Do not hard-code OpenAI embedding model names or arbitrary chat model IDs in memory code. \\
\  \\
\#\#\# Your Task: \\
Modify the code above so that it fully satisfies the following design goals: \\
\  \\
1. **Modular, clear design:**   \\
   - Prefer a small, readable structure (private helpers, inner classes, or a single `MemoStructure` with well-named methods). Avoid unnecessary scaffolding.   \\
   - If you use several stores (e.g. Chroma, graph), give each a clear role and data shape. \\
\  \\
2. **Retrieve / update orchestration:**   \\
   - Implement `MemoStructure` with `retrieve` / `update`.   \\
   - `retrieve()` can chain internal steps when that helps.   \\
   - `update()` should persist new evidence from the finished trajectory in a consistent order. \\
\  \\
3. **Out-of-the-Box Reasoning:**   \\
   - Do not just mechanically call each layer one by one — think about the **semantic flow of information**.   \\
   - Consider cases like:   \\
     - what type of memory layer can be used according to the analysis result or task description, with the aim to better assist the agent to finish it's task?  \\
     - what order and input output should be best suitable for the analysis result or task description?  \\
     - Think about high-level stategy: what can be a good memory structure, and has good ability to transfer to other area?  \\
   - Make sure each layer plays a meaningful role in the system. \\
   - Directly perform simple plans or content writing based on if/else patterns should be avoided, since this will hurt the transfer ability. \\
   - Keep the retrieved memory clean and useful, aviod cutting off meaningful texts, repeat same patterns in the retrieved memory. \\
\  \\
4. **Integration with Utilities:**   \\
   - Feel free to use any provided utility functions (e.g., similarity calculation, interaction with databases, hire new agent) if relevant. The tools available will be listed in `TOOLS` section. \\
   - You can also create your own tools if neccessary, think out of the box. \\
\  \\
5. **Code Quality:**   \\
   - Output clean, runnable Python code following PEP8.   \\
   - Ensure `retrieve()` and `update()` accept `EpisodeRecorder` and orchestrate the pipeline end-to-end.   \\
   - Initialize any stores or clients in `MemoStructure.\_\_init\_\_` as needed. \\
   - Define a module-level factory function exactly as `def build\_memo() -> MemoStructure`, and return an instance of your top-level `MemoStructure` subclass. \\
   - Do not overuse defensive programming; raise appropriate exceptions when unexpected conditions occur to facilitate debugging. \\
\  \\
6. **Coherent Policy Logic:** \\
    - Avoid placeholders like pass or \# TODO. \\
    - Avoid hard-coded if/else branches or enumerated case handling; instead, express the logic through modular policy functions, scoring mechanisms, or composable decision rules. \\
    - Instead of enumerating case-specific rules, express generalizable principles that could apply across different families or new unseen tasks. \\
    - The logic should be adaptable and compositional, not dependent on predefined constants or string names. \\
    - Use abstractions instead of specific family identifiers. \\
\  \\
The goal is to ensure the memory policy behaves consistently across tasks and supports generalization, not to hard-code specific task behaviors. \\
\  \\
\#\#\# Important: \\
- Think creatively about data flow — outputs of one layer can feed into the next. \\
- Each layer's functionality and stored data should be clearly designed. \\
- Use project-local imports under `meta\_self\_evolve.*`; do not import from `utils.hire\_agent` or `evals.utils.hire\_agent`. \\
- Provide **only the final rewritten code**, no explanations. \\

\textbf{[USER]}

Please generated new code based on your understanding about the task and requirements.  \\

\end{promptbox}

\promptcaption{Meta Agent Reflection Repair Prompt}{prompt:meta-agent-reflection-repair}
\begin{promptbox}
\ 
\textbf{Meta Agent Reflection Repair Prompt} \\

\textbf{[SYSTEM]}

You are a senior AI software engineer and code repair expert. \\
Your role is to carefully analyze the provided code and the error information, identify potential errors or design flaws, and directly rewrite or edit the code to fix those issues — while keeping the main design goals and intentions exactly the same. \\
\  \\
You are given the following context and base classes: \\
<BACKBONE\_CODE> \\
"""Abstract memo backbone inlined into codegen prompts. \\
\  \\
Only :class:`MemoStructure` is required. Implement a clear, maintainable \\
``retrieve`` / ``update`` pair; use private helpers or inner classes as needed. \\
""" \\
\  \\
from \_\_future\_\_ import annotations \\
\  \\
from abc import ABC, abstractmethod \\
from typing import Any, Optional \\
\  \\
from meta\_self\_evolve.contracts.types import EpisodeRecorder, RetrievedMemoryPayload \\
\  \\
\  \\
class MemoStructure(ABC): \\
    def \_\_init\_\_(self) -> None: \\
        self.database: Optional[Any] = None \\
\  \\
    @abstractmethod \\
    async def retrieve(self, recorder: EpisodeRecorder) -> RetrievedMemoryPayload: \\
        """Return structured memory for the next episode (see ``RetrievedMemoryPayload``).""" \\
        ... \\
\  \\
    @abstractmethod \\
    async def update(self, recorder: EpisodeRecorder) -> None: \\
        """Incorporate a finished trajectory (init, steps, reward, etc.).""" \\
        ... \\
\  \\
</BACKBONE\_CODE> \\
\  \\
<CODE\_INPUT> \\
\  \\
    Your `retrieve`, `update` will take `EpisodeRecorder` as input, which has following attributes: \\
    \{ \\
  "init": \{ \\
    "description": "Initial task text and observation images.", \\
    "type": "InitRecord" \\
  \}, \\
  "steps": \{ \\
    "description": "Per-step action and observation history.", \\
    "type": "list[StepRecord]" \\
  \}, \\
  "memory\_retrieved": \{ \\
    "description": "Structured output from retrieve() before the episode.", \\
    "type": "RetrievedMemoryPayload" \\
  \}, \\
  "reward": \{ \\
    "description": "Scalar episode reward after finish.", \\
    "type": "float" \\
  \}, \\
  "messages": \{ \\
    "description": "Full LLM conversation history for this episode.", \\
    "type": "list[dict]" \\
  \} \\
\} \\
    - For `retrieve`, only leverage `.init` attribute. \\
    - For `update`, you can leverage all listed attribute. \\
     \\
</CODE\_INPUT> \\
\  \\
<CODE\_USAGE> \\
Your memory structure will be used in the agent workflow: \\
    - `retrieve(recorder)`: used **before** executing the task; must return ``RetrievedMemoryPayload`` (see schema below). \\
    - `update(recorder)`: used **after** task is finished, to update the trajectory, reward, or other information. \\
\  \\
\#\#\# RetrievedMemoryPayload (return type of ``MemoStructure.retrieve``) \\
``retrieve`` must return a **JSON-serializable** ``RetrievedMemoryPayload`` (import from ``meta\_self\_evolve.contracts.types``), i.e. a dict with: \\
\  \\
- ``items``: a **flat** list of memory dicts; each dict may include: \\
  - ``text`` (``str``, optional): guidance for the execution agent. \\
  - ``images`` (optional): list of image refs with the same JSON shape as trajectory ``ImageRef``: ``\{"kind": "path"|"url", "value": "<relative path or URL>", "mime": "<optional>"\}``. When ``kind`` is ``"path"``, ``value`` is resolved relative to the episode **artifact root**. \\
  - ``metadata`` (``dict``, optional): any structured fields (e.g. ``memory\_id``, ``score``, ``source``, layer names). \\
- ``metadata`` (``dict``): episode-level retrieval notes (use ``\{\}`` if none). \\
\  \\
**Empty retrieval:** return ``\{"items": [], "metadata": \{\}\}`` (or use ``empty\_retrieved\_memory\_payload()`` from ``meta\_self\_evolve.common.retrieved\_memory``). \\
\  \\
**Image / text budgets (truncation is expected):** \\
- Put **only a small number** of images in ``items[*].images`` (typically 0–2). Extra images are **dropped**; do not rely on the model seeing every path you emit. \\
- The JSON text of the payload is also **truncated** to ``RolloutLimits.max\_retrieved\_memory\_chars`` in the execution agent's first user turn. \\
- **Execution agent:** at most ``RolloutLimits.max\_retrieved\_memory\_images`` images are loaded from the payload (in ``items`` order) on the first user turn; this is **separate** from the per-step screenshot budget ``RolloutLimits.max\_images\_per\_step`` for environment observations. \\
- **Meta analysis:** after the truncated JSON block, at most ``MetaTrajectoryLimits.meta\_retrieved\_memory\_images\_per\_episode`` payload images are attached per sampled episode (again in order); this is **separate** from ``MetaTrajectoryLimits.max\_meta\_images\_per\_episode`` for trajectory screenshots. \\
\  \\
**Runtime:** the worker injects this into the first user turn as (1) a text block (header + pretty-printed JSON of the payload), then (2) ``image\_url`` parts for each referenced image, subject to the limits above. \\
\  \\
</CODE\_USAGE> \\
\  \\
\#\#\# Your Task: \\
Carefully inspect the code, and error information, detect the root cause of the errors, structural issues, or missing implementations, and **only fix the root cause code**. Here are some cheetsheet that could be useful: \\
\  \\
<GRAPH\_DATABASE\_INTERACTION> \\
\  \\
NETWORKX GRAPH CHEATSHEET \\
\  \\
Context:  \\
    import networkx as nx \\
    G = nx.Graph() \\
\  \\
1. NODE OPERATIONS \\
- G.add\_node(node, **attrs): Add a single node with optional attributes. \\
- G.add\_nodes\_from([n1, n2], **common\_attrs): Add multiple nodes at once (shared attributes apply to all). \\
- G.remove\_node(node): Remove a node and all edges connected to it. \\
- G.remove\_nodes\_from([n1, n2]): Remove multiple nodes. \\
- node in G: Check if a node exists. \\
- G.nodes: Get all nodes (NodeView). \\
- G.nodes[node]: Access node attributes as a dict. \\
- nx.set\_node\_attributes(G, \{node: \{"attr": value\}\}): Set attributes for nodes. \\
\  \\
2. EDGE OPERATIONS \\
- G.add\_edge(u, v, **attrs): Add an edge between two nodes. \\
- G.add\_edges\_from([(u, v), (x, y)], **attrs): Add multiple edges at once (shared attributes apply to all). \\
- G.remove\_edge(u, v): Remove a single edge. \\
- G.remove\_edges\_from([(u, v), (x, y)]): Remove multiple edges. \\
- G.has\_edge(u, v): Check if an edge exists. \\
- G.edges: Get all edges (EdgeView). \\
- G.edges[(u, v)]: Access edge attributes as a dict. \\
- nx.set\_edge\_attributes(G, \{(u, v): \{"weight": 1.0\}\}): Set attributes for edges. \\
\  \\
3. TRAVERSAL / NEIGHBORHOOD \\
- G.neighbors(node): Get neighbors of a node. \\
- G.adj[node]: Get dict of neighbors with edge data. \\
- nx.shortest\_path(G, source, target): Find one shortest path between nodes. \\
- nx.shortest\_path\_length(G, source, target): Get shortest path length. \\
- nx.all\_simple\_paths(G, source, target, cutoff): Generate all simple paths up to a cutoff length. \\
- nx.connected\_components(G): Get connected components as node sets. \\
- G.subgraph([n1, n2, n3]): Extract a subgraph induced by given nodes. \\
\  \\
4. ANALYSIS / CENTRALITY \\
- G.degree(node): Get degree (number of edges) for a single node. \\
- G.degree(): Get degree for all nodes (DegreeView). \\
- nx.degree\_centrality(G): Compute degree centrality (dict of node -> score). \\
- nx.betweenness\_centrality(G): Compute betweenness centrality. \\
- nx.pagerank(G): Compute PageRank scores for nodes. \\
- nx.clustering(G): Compute local clustering coefficient. \\
- nx.is\_connected(G): Check if graph is connected. \\
- nx.number\_connected\_components(G): Count connected components. \\
\  \\
5. UTILITIES \\
- G.copy(): Make a copy of the graph. \\
- G.clear(): Remove all nodes and edges. \\
- nx.to\_dict\_of\_dicts(G): Convert graph to adjacency dict. \\
- nx.to\_numpy\_array(G): Get adjacency matrix as a NumPy array. \\
\  \\
</GRAPH\_DATABASE\_INTERACTION> \\
\  \\
<CHROMA\_DATABASE\_INTERACTION> \\
\#\# Initialize Chroma DB \\
\  \\
Import: `from langchain\_chroma import Chroma` \\
\  \\
Use `embedder = EmbeddingClient()` and `db = Chroma(embedding\_function=embedder)` to create the database. DO NOT use persist\_dir. \\
\  \\
The default LangChain helpers (`add\_texts`, `similarity\_search(query: str)`) are **text-only** on the query side: `similarity\_search` takes a string query, which `EmbeddingClient` embeds as a **single fused text vector** (DashScope multimodal embedding in text mode). \\
\  \\
For **screenshots, multiple images, or text+image fusion**, do **not** assume `similarity\_search` sees pixels. Build fused vectors explicitly with `await embedder.get\_fused\_embedding([...])` (see Embedding Client in TOOLS). One `get\_fused\_embedding` call returns **one** fused vector; need multiple vectors → call **multiple times**. \\
\  \\
Recommended multimodal storage / retrieval pattern: \\
- Keep the **human-readable memory content** (summary, advice, metadata) in Chroma / graph / your normal memory structure. \\
- Store each multimodal fused vector in a **sidecar index** keyed by a stable `memory\_id` (for example a dict or list storing `\{memory\_id, embedding, metadata\}`). \\
- At retrieve time, build **one fused query vector**, compare it against stored multimodal vectors (for example with `EmbeddingClient.compute\_one\_to\_group\_similarity(...)`), rank by similarity, then map the top hits back to `memory\_id` and finally return the linked readable memory content. \\
- For recorder screenshots, local image paths like `recorder.init.images[i].value` are relative to the current episode artifact root. In normal worker execution, `EmbeddingClient` can resolve those relative paths automatically from runtime context. If you work with images outside that runtime, pass `base\_dir=` explicitly. \\
\  \\
\#\#\# Add Memory: Adds new text entries to the database and returns their unique IDs.  \\
\  \\
db.add\_texts( \\
    texts: List[str], \\
    metadatas: Optional[Union[str, int, float, bool, None]] = None, \\
    ids: Optional[List[str]] = None \\
) -> List[str] \\
\  \\
- metadatas must be **flat list**: each value must be a single primitive type (str, int, float, bool, or None). \\
- You cannot pass lists, nested dicts, or other complex objects. \\
- If you need to store structured data, serialize it to a JSON string: \\
\  \\
\#\#\# Retrieve Memory \\
\  \\
db.similarity\_search( \\
    query: str, \\
    k: int = 4 \\
) -> List[Document] \\
\  \\
`query` is plain text only (text embedding path). For multimodal queries, use `get\_fused\_embedding(...)`, run your own similarity step, then map results back to stored memory items. \\
\  \\
return List[Document]: [ \\
  Document( \\
    page\_content="the agent found a key", \\
    metadata=\{"type": "item"\} \\
  ) \\
] \\
\  \\
\#\#\# Get by ID \\
db.get( \\
    ids: Optional[List[str]] = None \\
) -> Dict[str, List] \\
\  \\
\#\#\# Delete Memory \\
db.delete( \\
    ids: Optional[List[str]] = None \\
) -> None \\
     \\
</CHROMA\_DATABASE\_INTERACTION> \\
\  \\
<OTHER\_TOOLS> \\
\  \\
TOOLS AVAILABLE: \\
\  \\
1. Multimodal Chat Client \\
Class: MultimodalChatClient \\
\  \\
- Purpose: Asynchronous wrapper around OpenAI-compatible Chat Completions for text or multimodal user content. Use this when you need summarisation, synthesis, planning, or structured JSON output inside the memory code. \\
- Initialization (memory structure code — **must** match the benchmark execution model): \\
    from meta\_self\_evolve.llm.client import MultimodalChatClient \\
    client = MultimodalChatClient() \\
- Do **not** pass `model=...` inside generated memory code. \\
- Key Methods: \\
    - await client.complete(messages: List[Dict[str, Any]]) -> str: \\
        Send a standard chat message list and return plain text. \\
    - await client.complete\_with\_system(*, system\_prompt: str, user\_input: str | list, history: Optional[List[Dict]] = None, ...) -> str: \\
        Convenience wrapper for one system prompt + one user input (+ optional history). **All parameters after `client` are keyword-only** (the real signature uses `*` before `system\_prompt=`). \\
    - await client.complete\_json\_object(messages: List[Dict[str, Any]], *, temperature: float | None = None) -> Dict[str, Any]: \\
        Request a JSON object response. Pass `messages` positionally; `temperature=` is keyword-only if needed. \\
    - await client.complete\_json\_with\_system(*, system\_prompt: str, user\_input: str | list, history: Optional[List[Dict]] = None, temperature: float | None = None) -> Dict[str, Any]: \\
        Convenience wrapper for structured JSON output with a system prompt. **All parameters after `client` are keyword-only** (the real signature uses `*` before `system\_prompt=`). \\
- Usage Example: \\
    reply = await client.complete\_with\_system( \\
        system\_prompt="You summarize browser trajectories.", \\
        user\_input="Summarize the key failure pattern in these steps.", \\
    ) \\
    structured = await client.complete\_json\_with\_system( \\
        system\_prompt="Return ONLY a JSON object matching the schema described below.", \\
        user\_input="...", \\
    ) \\
\  \\
- IMPORTANT: \\
    - **Never** call `complete\_with\_system(...)` or `complete\_json\_with\_system(...)` with positional arguments (e.g. `complete\_json\_with\_system(system\_prompt, user\_input)` will raise `TypeError`). Always use `system\_prompt=...`, `user\_input=...`, and optionally `history=...`, `temperature=...` as keywords. \\
    - If you need structured output, use `complete\_json\_object(...)` or `complete\_json\_with\_system(...)`. \\
    - JSON output schemas should still be described clearly inside your system prompt. Format example: \\
\{ \\
    "location": \{ \\
        "type": "string", \\
        "description": "The location to get the weather for" \\
    \}, \\
    "unit": \{ \\
        "type": ["string", "null"], \\
        "description": "The unit to return the temperature in", \\
        "enum": ["F", "C"] \\
    \} \\
\} \\
\  \\
    - Maintain history explicitly by passing prior messages in `history=` or by constructing the full `messages` list yourself. \\
\  \\
2. Embedding Client \\
Class: EmbeddingClient \\
\  \\
- Purpose: DashScope **qwen3-vl-embedding** (fixed model and fixed vector dimension inside the class). Supports **fused** vectors for text-only, image(s), or text+image mixed inputs. Optional cosine similarity helpers. \\
- Initialization (do **not** pass an embedding model name): \\
    from meta\_self\_evolve.llm.embedding import EmbeddingClient, embed\_item\_text, embed\_item\_image\_path, embed\_item\_image\_url \\
    embedder = EmbeddingClient(retries=3, retry\_delay=1.0) \\
- Key Methods: \\
    - await embedder.get\_embedding(text: str) -> List[float]: \\
        Single **text** string → one fused vector (text-only path). \\
    - await embedder.get\_batch\_embeddings(texts: List[str]) -> List[List[float]]: \\
        One fused vector per string (independent calls). \\
    - await embedder.get\_fused\_embedding(items: List, base\_dir: Optional[Path | str] = None) -> List[float]: \\
        **Multimodal fusion**: `items` is an ordered list of parts (from the helpers above): \\
        - `embed\_item\_text("...")`  → `\{type:'text', text:...\}` \\
        - `embed\_item\_image\_path("relative/or/abs/path.png")`  → local image; if relative, `EmbeddingClient` resolves it against `base\_dir=` when provided, otherwise against the worker's current artifact-root runtime context. \\
        - `embed\_item\_image\_url("https://...")`  → image URL. \\
        One call uses `enable\_fusion=True` and returns **one** vector. If you need **multiple** vectors (e.g. separate vectors per document), call `get\_fused\_embedding` **multiple times**—do not pack unrelated fusion targets into one call expecting multiple vectors. \\
    - await EmbeddingClient.compute\_similarity(emb1: List[float], emb2: List[float], metric: str = "cosine") -> float: \\
        Computes similarity between two embeddings asynchronously. \\
    - await EmbeddingClient.compute\_one\_to\_group\_similarity(emb: List[float], group\_emb: List[List[float]], metric: str = "cosine") -> List[float]: \\
        Computes similarity between one embedding and a group of embeddings asynchronously. \\
    - `EmbeddingClient()` also implements `embed\_query()` / `embed\_documents()` / `\_\_call\_\_` so it can be passed into Chroma's `embedding\_function=...` for **text** document/query paths. \\
\  \\
- Notes: \\
    * Requires `DASHSCOPE\_API\_KEY` in the environment for DashScope. \\
    * For recorder screenshots in normal evaluation/runtime, relative image paths can usually be passed directly because the worker sets an artifact-root runtime context. If you use the embedder outside that runtime, pass `base\_dir=` explicitly. \\
    * Similarity helpers support cosine similarity and run in parallel for efficiency. \\
    * Do not import from `utils.hire\_agent` or `evals.utils.hire\_agent`; use `meta\_self\_evolve.llm.*` only. \\
\  \\
</OTHER\_TOOLS> \\
\  \\
\#\#\# Output: \\
Return **only the final corrected Python code**, no explanations or commentary. \\

\textbf{[USER]}

    Here's the code with potential error: \\
    <CODE\_FOR\_MODIFY> \\
    ```python \\
    \fstring{code\_str} \\
    ``` \\
    </CODE\_FOR\_MODIFY> \\
    And here's corresponding error message: \\
    \fstring{error\_msg} \\
    Find the root cause first and then modify only the corresponding code to avoid the error. \\
     \\

\end{promptbox}

\promptcaption{WebVoyager Execution Prompt}{prompt:webvoyager-execution}
\begin{promptbox}
\ 
\textbf{WebVoyager Execution Prompt} \\

\textbf{[SYSTEM]}

Imagine you are a robot browsing the web, just like humans. Now you need to complete a task. In each iteration, you will receive an Observation that includes a screenshot of a webpage and some texts. This screenshot will feature Numerical Labels placed in the TOP LEFT corner of each Web Element. \\
Carefully analyze the visual information to identify the Numerical Label corresponding to the Web Element that requires interaction, then follow the guidelines and choose one of the following actions: \\
1. Click a Web Element. \\
2. Delete existing content in a textbox and then type content.  \\
3. Scroll up or down. Multiple scrolls are allowed to browse the webpage. Pay attention!! The default scroll is the whole window. If the scroll widget is located in a certain area of the webpage, then you have to specify a Web Element in that area. I would hover the mouse there and then scroll. \\
4. Wait. Typically used to wait for unfinished webpage processes, with a duration of 5 seconds. \\
5. Go back, returning to the previous webpage. \\
6. Google, directly jump to the Google search page. When you can't find information in some websites, try starting over with Google. \\
7. Answer. This action should only be chosen when all questions in the task have been solved. \\
\  \\
Correspondingly, Action should STRICTLY follow the format: \\
- Click [Numerical\_Label] \\
- Type [Numerical\_Label]; [Content] \\
- Scroll [Numerical\_Label or WINDOW]; [up or down] \\
- Wait \\
- GoBack \\
- Google \\
- ANSWER; [content] \\
\  \\
Key Guidelines You MUST follow: \\
* Action guidelines * \\
1) To input text, NO need to click textbox first, directly type content. After typing, the system automatically hits `ENTER` key. Sometimes you should click the search button to apply search filters. Try to use simple language when searching.   \\
2) You must Distinguish between textbox and search button, don't type content into the button! If no textbox is found, you may need to click the search button first before the textbox is displayed.  \\
3) Execute only one action per iteration.  \\
4) STRICTLY Avoid repeating the same action if the webpage remains unchanged. You may have selected the wrong web element or numerical label. Continuous use of the Wait is also NOT allowed. \\
5) When a complex Task involves multiple questions or steps, select "ANSWER" only at the very end, after addressing all of these questions (steps). Flexibly combine your own abilities with the information in the web page. Double check the formatting requirements in the task when ANSWER.  \\
* Web Browsing Guidelines * \\
1) Don't interact with useless web elements like Login, Sign-in, donation that appear in Webpages. Pay attention to Key Web Elements like search textbox and menu. \\
2) Vsit video websites like YouTube is allowed BUT you can't play videos. Clicking to download PDF is allowed and will be analyzed by the Assistant API. \\
3) Focus on the numerical labels in the TOP LEFT corner of each rectangle (element). Ensure you don't mix them up with other numbers (e.g. Calendar) on the page. \\
4) Focus on the date in task, you must look for results that match the date. It may be necessary to find the correct year, month and day at calendar. \\
5) Pay attention to the filter and sort functions on the page, which, combined with scroll, can help you solve conditions like 'highest', 'cheapest', 'lowest', 'earliest', etc. Try your best to find the answer that best fits the task. \\
\  \\
Your reply should strictly follow the format: \\
Thought: \{Your brief thoughts (briefly summarize the info that will help ANSWER)\} \\
Action: \{One Action format you choose\} \\
\  \\
Then the User will provide: \\
Observation: \{A labeled screenshot Given by User\} \\

\textbf{[USER]}

Observation: \fstring{labeled\_screenshot\_and\_text} \\

\end{promptbox}

\promptcaption{WebVoyager Auto Evaluation Prompt}{prompt:webvoyager-auto-eval}
\begin{promptbox}
\ 
\textbf{WebVoyager Auto Evaluation Prompt} \\

\textbf{[SYSTEM]}

As an evaluator, you will be presented with three primary components to assist you in your role: \\
\  \\
1. Web Task Instruction: This is a clear and specific directive provided in natural language, detailing the online activity to be carried out. These requirements may include conducting searches, verifying information, comparing prices, checking availability, or any other action relevant to the specified web service (such as Amazon, Apple, ArXiv, BBC News, Booking etc). \\
\  \\
2. Result Screenshots: This is a visual representation of the screen showing the result or intermediate state of performing a web task. It serves as visual proof of the actions taken in response to the instruction. \\
\  \\
3. Result Response: This is a textual response obtained after the execution of the web task. It serves as textual result in response to the instruction. \\
\  \\
-- You DO NOT NEED to interact with web pages or perform actions such as booking flights or conducting searches on websites. \\
-- You SHOULD NOT make assumptions based on information not presented in the screenshot when comparing it to the instructions. \\
-- Your primary responsibility is to conduct a thorough assessment of the web task instruction against the outcome depicted in the screenshot and in the response, evaluating whether the actions taken align with the given instructions. \\
-- NOTE that the instruction may involve more than one task, for example, locating the garage and summarizing the review. Failing to complete either task, such as not providing a summary, should be considered unsuccessful. \\
-- NOTE that the screenshot is authentic, but the response provided by LLM is generated at the end of web browsing, and there may be discrepancies between the text and the screenshots. \\
-- Note the difference: 1) Result response may contradict the screenshot, then the content of the screenshot prevails, 2) The content in the Result response is not mentioned on the screenshot, choose to believe the content. \\
\  \\
You should elaborate on how you arrived at your final evaluation and then provide a definitive verdict on whether the task has been successfully accomplished, either as 'SUCCESS' or 'NOT SUCCESS'. \\

\textbf{[USER]}

TASK: <task> \\
Result Response: <answer> \\
<num> screenshots at the end:  \\
\  \\
\fstring{trailing\_screenshot\_image\_parts} \\
\  \\
Your verdict: \\

\end{promptbox}

\promptcaption{AgentVista Execution Prompt}{prompt:agentvista-execution}
\begin{promptbox}
\ 
\textbf{AgentVista Execution Prompt} \\

\textbf{[SYSTEM]}

You are a visual reasoning agent for the AgentVista benchmark. \\
\  \\
Each turn you must output: \\
Thought: <your reasoning> \\
Action: <exactly one action> \\
\  \\
Allowed actions (JSON payloads must be valid JSON): \\
- Action: WEB\_SEARCH[\{"query": "...", "max\_results": 10\}] \\
- Action: IMAGE\_SEARCH[\{"query": "...", "max\_results": 10\}] \\
- Action: VISIT[\{"url": "https://...", "goal": "what to extract"\}] \\
- Action: CODE[\{"code": "..."\}] \\
- Action: ANSWER[your final answer] \\
\  \\
When you are done, use Action: ANSWER[...]. If the task expects a tagged final answer, include <answer>...</answer> inside the ANSWER payload. \\
Only one Action line per reply. \\

\textbf{[USER]}

Solve the task using tools if needed. Images are attached in order (original\_image, original\_image\_1, ...). \\
\  \\
Question: \\
\fstring{question\_display} \\

\end{promptbox}

\promptcaption{AgentVista Evaluation Prompt}{prompt:agentvista-evaluation}
\begin{promptbox}
\ 
\textbf{AgentVista Evaluation Prompt} \\

\textbf{[SYSTEM]}

You are an intelligent chatbot designed for evaluating the correctness of generative outputs for question-answer pairs. Compare the predicted answer with the correct answer and determine if they match meaningfully. Consider synonyms or paraphrases as valid matches. \\

\textbf{[USER]}

1. **Question**: \fstring{question} \\
2. **Ground Truth Answer**: \fstring{ground\_truth} \\
3. **Model Predicted Answer**: \fstring{prediction} \\
\  \\
Evaluate the model's prediction against the ground truth. Output an integer score: 1 for correct, 0 for incorrect. \\
Respond using exactly: Score: 1 or Score: 0 \\
Explanation: <your explanation> \\

\end{promptbox}

\promptcaption{AgentVista Visit Tool Summary Prompt}{prompt:agentvista-visit-tool-summary}
\begin{promptbox}
\ 
\textbf{AgentVista Visit Tool Summary Prompt} \\

\textbf{[SYSTEM]}

You are a helpful assistant that summarizes webpage content based on user goals. \\

\textbf{[USER]}

Please process the following webpage content and user goal to extract relevant information: \\
\  \\
\#\# **Webpage Content** \\
\fstring{content} \\
\  \\
\#\# **User Goal** \\
\fstring{goal} \\
\  \\
\#\# **Task Guidelines** \\
1. **Content Scanning**: Locate the **specific sections/data** directly related to the user's goal within the webpage content \\
2. **Key Extraction**: Identify and extract the **most relevant information** from the content. Never miss any important information. Output the **full original context** as far as possible (can be more than three paragraphs) \\
3. **Summary Output**: Organize into a concise paragraph with logical flow, prioritizing clarity and judging the contribution of the information to the goal \\
\  \\
\#\# **Output Format** \\
Please respond in JSON format with the following fields: \\
\{ \\
  "evidence": "Key quotes or facts from the page that are directly relevant to the goal", \\
  "summary": "A concise summary of how the webpage content answers or relates to the user's goal" \\
\} \\

\end{promptbox}

\promptcaption{Retrieved Memory Injection Prompt Fragment}{prompt:retrieved-memory-injection}
\begin{promptbox}
\ 
\textbf{Retrieved Memory Injection Prompt Fragment} \\

\textbf{[USER]}

Retrieved memory (from retrieve; use as guidance, not as task text): \\
\fstring{retrieved\_memory\_payload\_json} \\

\end{promptbox}






\newpage
\section*{NeurIPS Paper Checklist}

\begin{enumerate}

\item {\bf Claims}
    \item[] Question: Do the main claims made in the abstract and introduction accurately reflect the paper's contributions and scope?
    \item[] Answer: \answerYes{} 
    \item[] Justification: The paper’s contributions and scope can be found in the section of Abstract and Section 1.
    \item[] Guidelines:
    \begin{itemize}
        \item The answer \answerNA{} means that the abstract and introduction do not include the claims made in the paper.
        \item The abstract and/or introduction should clearly state the claims made, including the contributions made in the paper and important assumptions and limitations. A \answerNo{} or \answerNA{} answer to this question will not be perceived well by the reviewers. 
        \item The claims made should match theoretical and experimental results, and reflect how much the results can be expected to generalize to other settings. 
        \item It is fine to include aspirational goals as motivation as long as it is clear that these goals are not attained by the paper. 
    \end{itemize}

\item {\bf Limitations}
    \item[] Question: Does the paper discuss the limitations of the work performed by the authors?
    \item[] Answer: \answerYes{} 
    \item[] Justification: The limitations of the work are described in Appendix A.
    \item[] Guidelines:
    \begin{itemize}
        \item The answer \answerNA{} means that the paper has no limitation while the answer \answerNo{} means that the paper has limitations, but those are not discussed in the paper. 
        \item The authors are encouraged to create a separate ``Limitations'' section in their paper.
        \item The paper should point out any strong assumptions and how robust the results are to violations of these assumptions (e.g., independence assumptions, noiseless settings, model well-specification, asymptotic approximations only holding locally). The authors should reflect on how these assumptions might be violated in practice and what the implications would be.
        \item The authors should reflect on the scope of the claims made, e.g., if the approach was only tested on a few datasets or with a few runs. In general, empirical results often depend on implicit assumptions, which should be articulated.
        \item The authors should reflect on the factors that influence the performance of the approach. For example, a facial recognition algorithm may perform poorly when image resolution is low or images are taken in low lighting. Or a speech-to-text system might not be used reliably to provide closed captions for online lectures because it fails to handle technical jargon.
        \item The authors should discuss the computational efficiency of the proposed algorithms and how they scale with dataset size.
        \item If applicable, the authors should discuss possible limitations of their approach to address problems of privacy and fairness.
        \item While the authors might fear that complete honesty about limitations might be used by reviewers as grounds for rejection, a worse outcome might be that reviewers discover limitations that aren't acknowledged in the paper. The authors should use their best judgment and recognize that individual actions in favor of transparency play an important role in developing norms that preserve the integrity of the community. Reviewers will be specifically instructed to not penalize honesty concerning limitations.
    \end{itemize}

\item {\bf Theory assumptions and proofs}
    \item[] Question: For each theoretical result, does the paper provide the full set of assumptions and a complete (and correct) proof?
    \item[] Answer: \answerNA{} 
    \item[] Justification: The paper is empirical study and does not include theoretical results.
    \item[] Guidelines:
    \begin{itemize}
        \item The answer \answerNA{} means that the paper does not include theoretical results. 
        \item All the theorems, formulas, and proofs in the paper should be numbered and cross-referenced.
        \item All assumptions should be clearly stated or referenced in the statement of any theorems.
        \item The proofs can either appear in the main paper or the supplemental material, but if they appear in the supplemental material, the authors are encouraged to provide a short proof sketch to provide intuition. 
        \item Inversely, any informal proof provided in the core of the paper should be complemented by formal proofs provided in appendix or supplemental material.
        \item Theorems and Lemmas that the proof relies upon should be properly referenced. 
    \end{itemize}

    \item {\bf Experimental result reproducibility}
    \item[] Question: Does the paper fully disclose all the information needed to reproduce the main experimental results of the paper to the extent that it affects the main claims and/or conclusions of the paper (regardless of whether the code and data are provided or not)?
    \item[] Answer: \answerYes{} 
    \item[] Justification: Implementation Details can be found in Section 3, Section 4, Appendix D, Appendix E and Appendix F.
    \item[] Guidelines:
    \begin{itemize}
        \item The answer \answerNA{} means that the paper does not include experiments.
        \item If the paper includes experiments, a \answerNo{} answer to this question will not be perceived well by the reviewers: Making the paper reproducible is important, regardless of whether the code and data are provided or not.
        \item If the contribution is a dataset and\slash or model, the authors should describe the steps taken to make their results reproducible or verifiable. 
        \item Depending on the contribution, reproducibility can be accomplished in various ways. For example, if the contribution is a novel architecture, describing the architecture fully might suffice, or if the contribution is a specific model and empirical evaluation, it may be necessary to either make it possible for others to replicate the model with the same dataset, or provide access to the model. In general. releasing code and data is often one good way to accomplish this, but reproducibility can also be provided via detailed instructions for how to replicate the results, access to a hosted model (e.g., in the case of a large language model), releasing of a model checkpoint, or other means that are appropriate to the research performed.
        \item While NeurIPS does not require releasing code, the conference does require all submissions to provide some reasonable avenue for reproducibility, which may depend on the nature of the contribution. For example
        \begin{enumerate}
            \item If the contribution is primarily a new algorithm, the paper should make it clear how to reproduce that algorithm.
            \item If the contribution is primarily a new model architecture, the paper should describe the architecture clearly and fully.
            \item If the contribution is a new model (e.g., a large language model), then there should either be a way to access this model for reproducing the results or a way to reproduce the model (e.g., with an open-source dataset or instructions for how to construct the dataset).
            \item We recognize that reproducibility may be tricky in some cases, in which case authors are welcome to describe the particular way they provide for reproducibility. In the case of closed-source models, it may be that access to the model is limited in some way (e.g., to registered users), but it should be possible for other researchers to have some path to reproducing or verifying the results.
        \end{enumerate}
    \end{itemize}

\item {\bf Open access to data and code}
    \item[] Question: Does the paper provide open access to the data and code, with sufficient instructions to faithfully reproduce the main experimental results, as described in supplemental material?
    \item[] Answer: \answerYes{} 
    \item[] Justification: Our data and codes could be found in supplementary files.
    \item[] Guidelines:
    \begin{itemize}
        \item The answer \answerNA{} means that paper does not include experiments requiring code.
        \item Please see the NeurIPS code and data submission guidelines (\url{https://neurips.cc/public/guides/CodeSubmissionPolicy}) for more details.
        \item While we encourage the release of code and data, we understand that this might not be possible, so \answerNo{} is an acceptable answer. Papers cannot be rejected simply for not including code, unless this is central to the contribution (e.g., for a new open-source benchmark).
        \item The instructions should contain the exact command and environment needed to run to reproduce the results. See the NeurIPS code and data submission guidelines (\url{https://neurips.cc/public/guides/CodeSubmissionPolicy}) for more details.
        \item The authors should provide instructions on data access and preparation, including how to access the raw data, preprocessed data, intermediate data, and generated data, etc.
        \item The authors should provide scripts to reproduce all experimental results for the new proposed method and baselines. If only a subset of experiments are reproducible, they should state which ones are omitted from the script and why.
        \item At submission time, to preserve anonymity, the authors should release anonymized versions (if applicable).
        \item Providing as much information as possible in supplemental material (appended to the paper) is recommended, but including URLs to data and code is permitted.
    \end{itemize}

\item {\bf Experimental setting/details}
    \item[] Question: Does the paper specify all the training and test details (e.g., data splits, hyperparameters, how they were chosen, type of optimizer) necessary to understand the results?
    \item[] Answer: \answerYes{} 
    \item[] Justification: Implementation Details can be found in Section 3, Section 4, Appendix D, Appendix E and Appendix F.
    \item[] Guidelines:
    \begin{itemize}
        \item The answer \answerNA{} means that the paper does not include experiments.
        \item The experimental setting should be presented in the core of the paper to a level of detail that is necessary to appreciate the results and make sense of them.
        \item The full details can be provided either with the code, in appendix, or as supplemental material.
    \end{itemize}

\item {\bf Experiment statistical significance}
    \item[] Question: Does the paper report error bars suitably and correctly defined or other appropriate information about the statistical significance of the experiments?
    \item[] Answer: \answerNo{} 
    \item[] Justification: The paper reports mean performance over three independent runs under the same held-out evaluation sets and evaluation protocols. This repeated-run evaluation is intended to reduce the effect of single-run randomness from search and model sampling.
    \item[] Guidelines:
    \begin{itemize}
        \item The answer \answerNA{} means that the paper does not include experiments.
        \item The authors should answer \answerYes{} if the results are accompanied by error bars, confidence intervals, or statistical significance tests, at least for the experiments that support the main claims of the paper.
        \item The factors of variability that the error bars are capturing should be clearly stated (for example, train/test split, initialization, random drawing of some parameter, or overall run with given experimental conditions).
        \item The method for calculating the error bars should be explained (closed form formula, call to a library function, bootstrap, etc.)
        \item The assumptions made should be given (e.g., Normally distributed errors).
        \item It should be clear whether the error bar is the standard deviation or the standard error of the mean.
        \item It is OK to report 1-sigma error bars, but one should state it. The authors should preferably report a 2-sigma error bar than state that they have a 96\% CI, if the hypothesis of Normality of errors is not verified.
        \item For asymmetric distributions, the authors should be careful not to show in tables or figures symmetric error bars that would yield results that are out of range (e.g., negative error rates).
        \item If error bars are reported in tables or plots, the authors should explain in the text how they were calculated and reference the corresponding figures or tables in the text.
    \end{itemize}

\item {\bf Experiments compute resources}
    \item[] Question: For each experiment, does the paper provide sufficient information on the computer resources (type of compute workers, memory, time of execution) needed to reproduce the experiments?
    \item[] Answer: \answerYes{} 
    \item[] Justification: Experiments compute resources can be found in Section 4, Appendix F and Appendix H.
    \item[] Guidelines:
    \begin{itemize}
        \item The answer \answerNA{} means that the paper does not include experiments.
        \item The paper should indicate the type of compute workers CPU or GPU, internal cluster, or cloud provider, including relevant memory and storage.
        \item The paper should provide the amount of compute required for each of the individual experimental runs as well as estimate the total compute. 
        \item The paper should disclose whether the full research project required more compute than the experiments reported in the paper (e.g., preliminary or failed experiments that didn't make it into the paper). 
    \end{itemize}
    
\item {\bf Code of ethics}
    \item[] Question: Does the research conducted in the paper conform, in every respect, with the NeurIPS Code of Ethics \url{https://neurips.cc/public/EthicsGuidelines}?
    \item[] Answer: \answerYes{} 
    \item[] Justification: The research conducted in the paper conform, in every respect, with the NeurIPS Code of Ethics.
    \item[] Guidelines:
    \begin{itemize}
        \item The answer \answerNA{} means that the authors have not reviewed the NeurIPS Code of Ethics.
        \item If the authors answer \answerNo, they should explain the special circumstances that require a deviation from the Code of Ethics.
        \item The authors should make sure to preserve anonymity (e.g., if there is a special consideration due to laws or regulations in their jurisdiction).
    \end{itemize}

\item {\bf Broader impacts}
    \item[] Question: Does the paper discuss both potential positive societal impacts and negative societal impacts of the work performed?
    \item[] Answer: \answerYes{} 
    \item[] Justification: Broader impacts can be found in Appendix B.
    \item[] Guidelines:
    \begin{itemize}
        \item The answer \answerNA{} means that there is no societal impact of the work performed.
        \item If the authors answer \answerNA{} or \answerNo, they should explain why their work has no societal impact or why the paper does not address societal impact.
        \item Examples of negative societal impacts include potential malicious or unintended uses (e.g., disinformation, generating fake profiles, surveillance), fairness considerations (e.g., deployment of technologies that could make decisions that unfairly impact specific groups), privacy considerations, and security considerations.
        \item The conference expects that many papers will be foundational research and not tied to particular applications, let alone deployments. However, if there is a direct path to any negative applications, the authors should point it out. For example, it is legitimate to point out that an improvement in the quality of generative models could be used to generate Deepfakes for disinformation. On the other hand, it is not needed to point out that a generic algorithm for optimizing neural networks could enable people to train models that generate Deepfakes faster.
        \item The authors should consider possible harms that could arise when the technology is being used as intended and functioning correctly, harms that could arise when the technology is being used as intended but gives incorrect results, and harms following from (intentional or unintentional) misuse of the technology.
        \item If there are negative societal impacts, the authors could also discuss possible mitigation strategies (e.g., gated release of models, providing defenses in addition to attacks, mechanisms for monitoring misuse, mechanisms to monitor how a system learns from feedback over time, improving the efficiency and accessibility of ML).
    \end{itemize}
    
\item {\bf Safeguards}
    \item[] Question: Does the paper describe safeguards that have been put in place for responsible release of data or models that have a high risk for misuse (e.g., pre-trained language models, image generators, or scraped datasets)?
    \item[] Answer: \answerNA{} 
    \item[] Justification: The paper does not release a new pretrained model, image generator, scraped dataset, or other high-risk artifact for public use.
    \item[] Guidelines:
    \begin{itemize}
        \item The answer \answerNA{} means that the paper poses no such risks.
        \item Released models that have a high risk for misuse or dual-use should be released with necessary safeguards to allow for controlled use of the model, for example by requiring that users adhere to usage guidelines or restrictions to access the model or implementing safety filters. 
        \item Datasets that have been scraped from the Internet could pose safety risks. The authors should describe how they avoided releasing unsafe images.
        \item We recognize that providing effective safeguards is challenging, and many papers do not require this, but we encourage authors to take this into account and make a best faith effort.
    \end{itemize}

\item {\bf Licenses for existing assets}
    \item[] Question: Are the creators or original owners of assets (e.g., code, data, models), used in the paper, properly credited and are the license and terms of use explicitly mentioned and properly respected?
    \item[] Answer: \answerYes{} 
    \item[] Justification: The creators or original owners of assets (e.g., code, data, models), used in the paper, are properly credited.
    \item[] Guidelines:
    \begin{itemize}
        \item The answer \answerNA{} means that the paper does not use existing assets.
        \item The authors should cite the original paper that produced the code package or dataset.
        \item The authors should state which version of the asset is used and, if possible, include a URL.
        \item The name of the license (e.g., CC-BY 4.0) should be included for each asset.
        \item For scraped data from a particular source (e.g., website), the copyright and terms of service of that source should be provided.
        \item If assets are released, the license, copyright information, and terms of use in the package should be provided. For popular datasets, \url{paperswithcode.com/datasets} has curated licenses for some datasets. Their licensing guide can help determine the license of a dataset.
        \item For existing datasets that are re-packaged, both the original license and the license of the derived asset (if it has changed) should be provided.
        \item If this information is not available online, the authors are encouraged to reach out to the asset's creators.
    \end{itemize}

\item {\bf New assets}
    \item[] Question: Are new assets introduced in the paper well documented and is the documentation provided alongside the assets?
    \item[] Answer: \answerYes{} 
    \item[] Justification: This can be found in Appendix F and our supplementary files.
    \item[] Guidelines:
    \begin{itemize}
        \item The answer \answerNA{} means that the paper does not release new assets.
        \item Researchers should communicate the details of the dataset\slash code\slash model as part of their submissions via structured templates. This includes details about training, license, limitations, etc. 
        \item The paper should discuss whether and how consent was obtained from people whose asset is used.
        \item At submission time, remember to anonymize your assets (if applicable). You can either create an anonymized URL or include an anonymized zip file.
    \end{itemize}

\item {\bf Crowdsourcing and research with human subjects}
    \item[] Question: For crowdsourcing experiments and research with human subjects, does the paper include the full text of instructions given to participants and screenshots, if applicable, as well as details about compensation (if any)? 
    \item[] Answer: \answerNA{} 
    \item[] Justification: The paper does not involve crowdsourcing nor research with human subjects.
    \item[] Guidelines:
    \begin{itemize}
        \item The answer \answerNA{} means that the paper does not involve crowdsourcing nor research with human subjects.
        \item Including this information in the supplemental material is fine, but if the main contribution of the paper involves human subjects, then as much detail as possible should be included in the main paper. 
        \item According to the NeurIPS Code of Ethics, workers involved in data collection, curation, or other labor should be paid at least the minimum wage in the country of the data collector. 
    \end{itemize}

\item {\bf Institutional review board (IRB) approvals or equivalent for research with human subjects}
    \item[] Question: Does the paper describe potential risks incurred by study participants, whether such risks were disclosed to the subjects, and whether Institutional Review Board (IRB) approvals (or an equivalent approval/review based on the requirements of your country or institution) were obtained?
    \item[] Answer: \answerNA{} 
    \item[] Justification: The paper does not involve crowdsourcing nor research with human subjects.
    \item[] Guidelines:
    \begin{itemize}
        \item The answer \answerNA{} means that the paper does not involve crowdsourcing nor research with human subjects.
        \item Depending on the country in which research is conducted, IRB approval (or equivalent) may be required for any human subjects research. If you obtained IRB approval, you should clearly state this in the paper. 
        \item We recognize that the procedures for this may vary significantly between institutions and locations, and we expect authors to adhere to the NeurIPS Code of Ethics and the guidelines for their institution. 
        \item For initial submissions, do not include any information that would break anonymity (if applicable), such as the institution conducting the review.
    \end{itemize}

\item {\bf Declaration of LLM usage}
    \item[] Question: Does the paper describe the usage of LLMs if it is an important, original, or non-standard component of the core methods in this research? Note that if the LLM is used only for writing, editing, or formatting purposes and does \emph{not} impact the core methodology, scientific rigor, or originality of the research, declaration is not required.
    \item[] Answer: \answerNA{} 
    \item[] Justification: LLM is used only for writing, editing, or formatting purposes.
    \item[] Guidelines:
    \begin{itemize}
        \item The answer \answerNA{} means that the core method development in this research does not involve LLMs as any important, original, or non-standard components.
        \item Please refer to our LLM policy in the NeurIPS handbook for what should or should not be described.
    \end{itemize}

\end{enumerate}

\end{document}